\documentclass[nonacm,acmlarge]{acmart}

\AtBeginDocument{%
  }

\setcopyright{acmlicensed}
\copyrightyear{2023}
\acmYear{2023}
\acmDOI{XXXXXXX.XXXXXXX}

\usepackage[normalem]{ulem}
\usepackage{eqparbox}
\usepackage{amsmath}
\usepackage{latexsym, amsfonts, stmaryrd, graphicx, color}
\usepackage{cleveref}
\usepackage{tikz}
\usetikzlibrary{arrows,arrows.meta,shapes,snakes,automata,backgrounds,petri,
decorations.pathmorphing,positioning,fit}
\usepackage[pagewise]{lineno}
\usepackage{proof}
\usepackage{mathtools}
\usepackage{thm-restate}
\usepackage{subcaption}
\usepackage{listings}

\usepackage{xspace}
\usepackage{extarrows}
\usepackage{todonotes}

\newcommand{\zero}{\ensuremath{\mathsf{0}}}

\newcommand{\CF}[1]{\ensuremath{\mathsf{CF}(#1)}}
\newcommand{\trans}[1]{\ensuremath{\,[\/{#1}\/\rangle}\,}
\newcommand{\pre}[1]{\ensuremath{\!~^{\bullet}{#1}}}
\newcommand{\post}[1]{\ensuremath{{#1}^{\bullet}}}

\newcommand{\hist}[1]{\ensuremath{\lfloor #1 \rfloor}}
\newcommand{\histtwo}[2]{\ensuremath{\lfloor #1 \rfloor_{#2}}}

\newcommand{\nat}{\ensuremath{\mathbb{N}}}

\newcommand{\flt}[1]{\ensuremath{[\![{#1}]\!]}}

\newcommand{\pes}{\textsc{pes}}

\newcommand{\setenum}[1]{\{#1\}}
\newcommand{\setcomp}[2]{\{{#1} \mid {#2}\}}

\newcommand{\reachMark}[1]{\ensuremath{\mathcal{M}_{#1}}}

\newcommand{\firseq}[2]{\ensuremath{\mathcal{R}^{#1}_{#2}}}
\newcommand{\states}[1]{\ensuremath{\mathsf{St}(#1)}}

\newcommand{\lead}[1]{\ensuremath{\mathit{lead}(#1)}}
\newcommand{\start}[1]{\ensuremath{\mathit{start}(#1)}}
\newcommand{\fs}{\textsf{fs}}

\newcommand{\MC}[1]{\ensuremath{\sim}}

\newcommand{\pmv}[1]{\ensuremath{#1}}

\newcommand{\Conf}[2]{\ensuremath{\mathsf{Conf}_{{#2}}(#1)}}

\newcommand{\inib}[1]{\ensuremath{\!~^{\circ}{#1}}}

\newcommand{\len}[1]{\ensuremath{\mathit{len}(#1)}}

\newcommand{\cn}{\textsc{on}}
\newcommand{\ca}{\textsc{cn}}
\newcommand{\ontocn}[1]{\ensuremath{\mathcal{O}(#1)}}

\newcommand{\Comment}[1]{}
\newcommand{\un}[1]{\underline{#1}}
\newcommand{\pr}{\mathrel{\triangleright}}

\newcommand{\re}{\mathtt{r}}

\newcommand{\anR}{U}
\newcommand{\anr}{u}

\newcommand{\rcn}{r\textsc{cn}}

\newcommand{\Inib}[2]{\ensuremath{\!~^{\circ}{#1}_{#2}}}

\newcommand{\cnconf}{{\natural}}

\newcommand{\rcntorpes}[1]{{\mathcal{Q}_r(#1)}}
\newcommand{\rpestorcn}[1]{{\mathcal{A}_r(#1)}}
\newcommand{\pestocn}[1]{{\mathcal{A}(#1)}}

\newcommand{\anRCN}{\arcn{V}{\bwdset}}

\newcommand{\rpes}{{r\pes}\xspace}

\newcommand{\inet}{{\sc{ipt}}\xspace}

\newcommand{\toberemoved}[1]{}

\makeatletter
\DeclareRobustCommand{\cev}[1]{%
  \mathpalette\do@cev{#1}%
}
\newcommand{\do@cev}[2]{%
  \fix@cev{#1}{+}%
  \reflectbox{$\m@th#1\overrightarrow{\reflectbox{$\fix@cev{#1}{-}\m@th#1#2\fix@cev{#1}{+}$}}$}%
  \fix@cev{#1}{-}%
}
\newcommand{\fix@cev}[2]{%
  \ifx#1\displaystyle
    \mkern#23mu
  \else
    \ifx#1\textstyle
      \mkern#23mu
    \else
      \ifx#1\scriptstyle
        \mkern#22mu
      \else
        \mkern#22mu
      \fi
    \fi
  \fi
}

\makeatother

\newcommand{\fwdset}{{\overline{T}}}
\newcommand{\bwdset}{{\underline T}}
\newcommand{\afwd}{{t}}
\newcommand{\abwd}{{\underline t}}

\newcommand{\arcn}[2]{{#1^{#2}}}

\newcommand{\cntopes}[1]{{\mathcal{Q}(#1)}}
\newcommand{\pcntoocc}[1]{{\mathcal{Z}(#1)}}
\newcommand{\toconf}[2]{{\ \xlongrightarrow{#1}}_{#2}\ }

\lstdefinelanguage{customc}{
  belowcaptionskip=1\baselineskip,
  breaklines=true,
  xleftmargin=\parindent,
  language=C,
  showstringspaces=false,
  basicstyle=\footnotesize\ttfamily,
  keywordstyle=\bfseries\color{green!40!black},
  commentstyle=\itshape\color{purple!40!black},
  identifierstyle=\color{blue},
  stringstyle=\color{orange},
}

\lstdefinestyle{INLINE}{
}

\newcommand{\CI}{\lstinline[language=customc,style=INLINE]}

\begin{document}

\title{A Reversible Perspective on Petri Nets and Event Structures}

\author{Hern\'an Melgratti}
\orcid{0000-0003-0760-0618}
\affiliation{%
  \institution{ICC - Universidad de Buenos Aires, Conicet
}
  \city{Buenos Aires}
  \country{Argentina}
}

\author{Claudio Antares Mezzina}
\orcid{0000-0003-1556-2623}
\affiliation{%
  \institution{Dipartimento di Scienze Pure e Applicate, Universit\`a di Urbino}
  \city{Urbino}
  \country{Italy}}

\author{G. Michele Pinna}
\orcid{0000-0001-8911-1580}
\affiliation{%
  \institution{Dipartimento di Matematica e Informatica, Universit\`a di Cagliari}
  \city{Cagliari}
  \country{Italy}
}

\renewcommand{\shortauthors}{H. Melgratti, C.A. Mezzina and G.M. Pinna}

\begin{abstract}
  Event structures have emerged as a foundational model for concurrent
computation, explaining computational processes by outlining the events and
the relationships that dictate their execution. They play a pivotal role in
the study of key aspects of concurrent computation models, such as causality
and independence, and have found applications across a broad range of
languages and models, spanning realms like persistence, probabilities, and
quantum computing.
Recently, event structures have been extended to address reversibility, where
computational processes can undo previous computations. In this context,
reversible event structures provide abstract representations of processes
capable of both forward and backward steps in a computation.
Since their introduction, event structures have played a crucial role in
bridging operational models, traditionally exemplified by Petri nets and
process calculi, with denotational ones, i.e., algebraic domains. In this
context, we revisit the standard connection between Petri nets and event
structures under the lenses of reversibility. Specifically, we introduce a
subset of contextual Petri nets, dubbed {\em reversible causal nets\/}, that
precisely correspond to reversible prime event structures. The distinctive
feature of reversible causal nets lies in deriving causality from inhibitor
arcs, departing from the conventional dependence on the overlap between the
post and preset of transitions. In this way, we are able to operationally
explain the full model of reversible prime event structures.


\end{abstract}




\maketitle

\section{Introduction} \label{sec:intro}

\begin{figure}[t]
  \begin{subfigure}{.1\textwidth}
    \begin{center}
      \scalebox{0.75}{\scalebox{0.9}{\begin{tikzpicture}
\usetikzlibrary{decorations.pathmorphing}

\tikzset{snake it/.style={decorate, decoration=snake}}
\tikzstyle{cau}=[-,thick]
\tikzstyle{conf}=[snake it,thick]
\tikzstyle{transition}=[rectangle, draw=none,thick,minimum size=5mm]
\node[transition] (b) at (0,2)  {$b$}
;
\node[transition] (a) at (0,0) {$a$}
edge[cau] (b)
;
\node[transition] (c) at (2,2)  {$c$}
edge[conf] (b);
\end{tikzpicture}}}
    \end{center}
    \caption{$P$}\label{fig:exintro-es}
  \end{subfigure}
  \hspace{2cm}
  \begin{subfigure}{.25\textwidth}
    \begin{center}
      \scalebox{0.75}{\scalebox{0.9}{\begin{tikzpicture}
\usetikzlibrary{decorations.pathmorphing}

\tikzset{snake it/.style={decorate, decoration=snake}}
\tikzstyle{tr}=[->,thick]
\tikzstyle{transition}=[rectangle, draw=none,thick,minimum size=5mm]
\node[transition] (ab) at (4,2)  {$\{a,b\}$}
;
\node[transition] (ac) at (2,0)  {$\{a,c\}$}
;
\node[transition] (a) at (2,2){$\{a\}$}
edge[tr] (ac)
edge[tr] (ab)
;
\node[transition] (c) at (0,0) {$\{c\}$}
edge[tr] (ac)
;
\node[transition] (empty) at (0,2)  {$\emptyset$}
edge[tr] (a)
edge[tr] (c)
edge[tr] (ac)
;
\end{tikzpicture}}}
    \end{center}
    \caption{Transition system}\label{fig:exintro-conf}
  \end{subfigure}
  \hspace{2cm}
  \begin{subfigure}{.15\textwidth}
    \begin{center}
      \scalebox{0.75}{\scalebox{0.9}{\begin{tikzpicture}
\tikzstyle{inhibitorred}=[o-, draw=red,thick]
\tikzstyle{inhibitorblu}=[o-, draw=blue,thick]
\tikzstyle{pre}=[<-,thick]
\tikzstyle{post}=[->,thick]
\tikzstyle{readblue}=[-, draw=blue,thick]
\tikzstyle{transition}=[rectangle, draw=black,thick,minimum size=5mm]
\tikzstyle{place}=[circle, draw=black,thick,minimum size=5mm]
\node[place,tokens=1] (p1) at (0,5) [label=left:$s_1$] {};
\node[place,tokens=1] (p5) at (2,2.5) [label=right:$s_3$] {};
\node[place] (p2) at (0,2.5) [label=left:$s_2$] {};
\node[place] (p4) at (0,0) [label=left:$s_4$] {};
\node[place] (p6) at (2,0) [label=right:$s_5$] {};
\node[transition] (a) at (0,3.75)  {$a$}
edge[pre] (p1)
edge[post](p2);

\node[transition] (b) at (0,1.25) {$b$}
edge[pre] (p2)
edge[pre] (p5)
edge[post] (p4)
;
\node[transition] (c) at (2,1.25)  {$c$}
edge[pre] (p5)
edge[post] (p6);
\end{tikzpicture}}}
    \end{center}
    \caption{$N$}\label{fig:exintro-pn}
  \end{subfigure}
  \caption{A simple Prime Event Structure and its associated Petri net}
  \Description{A simple Prime Event Structure and its associated Petri net}
\end{figure}
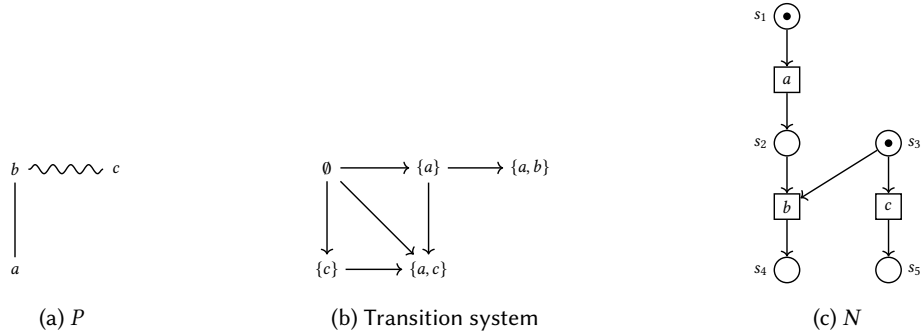

Event structures~\cite{Win:ES} are a well-established model of concurrency,
that describe computational processes through event occurrences and
constraints that regulate such occurrences (i.e., relations over events). In
their simplest form, known as \emph{prime event structures}, the relations are
\emph{causality} and \emph{conflict}. Causality dictates the order and
precedence of events in valid executions, while conflict denotes the mutual
exclusion of events -- conflicting events cannot coexist in any valid system
execution.
Consider a computational process denoted as $\mathsf{P}$, involving three
distinct events: $a$, $b$, and $c$, where $b$ causally depends on $a$, and $b$
and $c$ are in conflict. Describing the behavior of this system as a prime
event structure necessitates defining two binary relations over the set of
events: $<$ for causality and $\#$ for conflicts. Specifically, the relations
are defined as follows: ${<} = \{(a, b)\}$ and ${\#} = \{(b, c) , (c, b)\}$.
A graphical representation is shown in \Cref{fig:exintro-es}, where causality
is drawn with straight lines (to be read from bottom to top) and binary
conflicts are represented by curly lines.

The behaviour associated with such event structure can be understood in terms
of a transition system defined over {\em configurations} (i.e., sets of
events), as illustrated in \Cref{fig:exintro-conf}. For instance, the
transition $\emptyset \rightarrow \{a, c\}$ indicates that the initial state
$\emptyset$ (i.e., no event has been executed yet) may evolve to the state
$\{a,c\}$ by concurrently executing $a$ and $c$.
Moreover, the same configuration can be achieved by initially executing $a$
followed by $c$ (as outlined in the transitions
$\emptyset \rightarrow \{a\} \rightarrow \{a, c\}$).
Conversely, one can attain the identical configuration by first transitioning
through $c$ and then through $a$ (i.e.,
$\emptyset \rightarrow \{c\} \rightarrow \{a, c\}$). In other words, $a$ and
$c$ are concurrent. In contrast, the configuration $\{a, b\}$ can only be
reached by executing $a$ before $b$ since the occurrence of $b$ causally
depends on the prior execution of $a$.
Note that neither $\{b\}$ nor $\{a, b, c\}$ are configurations because, on the
one hand, $b$ cannot occur without $a$ and, on the other hand, $b$ and $c$
cannot occur in the same run of the system.

Event structures were originally proposed by Nielsen, Plotkin and
Winskel~\cite{NPW:PNES} as an intermediate abstraction in between Scott
domains (i.e., a denotational model) and Petri nets (i.e., an operational
model).
While event structures encompass a set of event occurrences and constraints
that govern the occurrence of such events, Petri nets describe the behaviour
of a system in terms of the consumption and production of data items (i.e.,
tokens) from repositories (i.e., places).
A Petri net $N$ corresponding to the prime event structure $\mathsf{P}$ above,
is depicted in \Cref{fig:exintro-pn}. It consists of five places $s_1$, $s_2$,
$s_3$, $s_4$ and $s_5$, three transitions $a$, $b$ and $c$ and two tokens
(depicted as bullets) respectively placed on $s_1$ and $s_3$.
The edges connecting places to transitions describe the consumption and
production of tokens; e.g., the firing (i.e., execution) of $b$ consumes a
token from each $s_2$ and $s_3$ and produces a token in $s_4$. The tokens
available in $N$ enable the firing of the transitions $a$ and $c$ but not that
of $b$ because there is no token in $s_2$. That missing token is produced by
the firing of $a$; hence, $b$ can be fired only after $a$ is so. For this
reason, $b$ {\em causally depends} on $a$.
However, $b$ can be fired only if $c$ is not (and vice versa) because each
transition requires a token from $s_3$, which just contains one. In this case,
$b$ and $c$ are in {\em conflict}.

The foundational work in~\cite{Win:ES} shows a tight connection, established
through a chain of correflections, between the category of safe nets and prime
event structures. Subsequent research efforts have systematically explored the
alignment of distinct incarnations of Petri nets with their respective classes
of event structures. Examples of this line of work include ~\cite{Bou:FESFN,
  Langerak:1992:BES, CaPi:PN14, GP:CSESPN}.

\begin{figure}[t]
  \begin{subfigure}{.2\textwidth}
    \begin{center}
      \scalebox{0.75}{\scalebox{0.9}
{\begin{tikzpicture}
\usetikzlibrary{decorations.pathmorphing}

\tikzset{snake it/.style={decorate, decoration=snake}}
\tikzstyle{tr}=[->,thick]
\tikzstyle{transition}=[rectangle, draw=none,thick,minimum size=5mm]
\tikzstyle{revtr}=[->,thick,draw=gray]

\node[transition] (a) at (2,2){$\{a\}$}
edge[tr] (ac)
edge[tr] (ab)
edge[revtr, bend right] (empty)

;
\node[transition] (c) at (0,0) {$\{c\}$}
edge[tr] (ac)
edge[revtr, bend right] (empty)
;
\node[transition] (empty) at (0,2)  {$\emptyset$}
edge[tr] (a)
edge[tr] (c)
edge[tr] (ac)
;
\node[transition] (ab) at (4,2)  {$\{a,b\}$}
 edge[revtr, bend right] (a)
 ;
\node[transition] (ac) at (2,0)  {$\{a,c\}$}
edge[revtr, bend right] (a)
edge[revtr, bend right] (c)
edge[revtr, bend right] (empty)
;
\end{tikzpicture}}}
    \end{center}
    \caption{Transition system}\label{fig:exintro-conf-rev}
  \end{subfigure}
  \hspace{3cm}
  \begin{subfigure}{.2\textwidth}
    \begin{center}
      \scalebox{0.75}{\scalebox{0.9}{\begin{tikzpicture}
\tikzstyle{inhibitorred}=[o-, draw=red,thick]
\tikzstyle{inhibitorblu}=[o-, draw=blue,thick]
\tikzstyle{pre}=[<-,thick]
\tikzstyle{post}=[->,thick]
\tikzstyle{readblue}=[-, draw=blue,thick]
\tikzstyle{transition}=[rectangle, draw=black,thick,minimum size=5mm]
\tikzstyle{place}=[circle, draw=black,thick,minimum size=5mm]
\tikzstyle{rev}=[rectangle, draw=gray,thick,,minimum size=5mm]
\tikzstyle{prerev}=[<-,thick,draw=gray]
\tikzstyle{postrev}=[->,thick,draw=gray]

\node[place,tokens=1] (p1) at (0,5) [label=left:$s_1$] {};
\node[place,tokens=1] (p5) at (2,2.5) [label=right:$s_3$] {};
\node[place] (p2) at (0,2.5) [label=left:$s_2$] {};
\node[place] (p4) at (0,0) [label=left:$s_4$] {};
\node[place] (p6) at (2,0) [label=right:$s_5$] {};
\node[transition] (a) at (0,3.75)  {$a$}
edge[pre] (p1)
edge[post](p2);
\node[rev] (ra) at (1,3.75)  {$\un a$}
edge[postrev, bend right] (p1)
edge[prerev, bend left](p2);

\node[transition] (b) at (0,1.25) {$b$}
edge[pre] (p2)
edge[pre] (p5)
edge[post] (p4)
;
\node[rev] (rb) at (1,1.25) {$\un b$}
edge[postrev, bend right] (p2)
edge[postrev, bend right] (p5)
edge[prerev, bend left] (p4)
;

\node[transition] (c) at (2,1.25)  {$c$}
edge[pre] (p5)
edge[post] (p6);

\node[rev] (rc) at (3,1.25)  {$\un c$}
edge[postrev, bend right] (p5)
edge[prerev, bend left] (p6);
\end{tikzpicture}}}
    \end{center}
    \caption{$\un N$}\label{fig:exintro-pn-rev}
  \end{subfigure}
  \caption{Causal-consistent reversible version of $N$}
   \Description{Causal-consistent reversible version of $N$}
\end{figure}
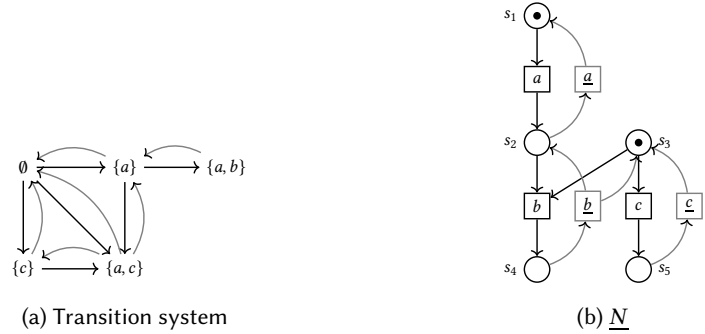

Recently, there has been a noteworthy extension of event structures to
accommodate {\em reversible} concurrent systems, namely a class of systems
that have lately received lot of attention because of their applications in
different fields~\cite{wg1,wg2}. These applications span programming
abstractions for reliable systems~\cite{DanosK05,DMedicM0Y20,LaneseLMSS13},
program analysis and debugging~\cite{LanesePV19}, bio-chemical simulation
modeling~\cite{biorev}, and quantum computing~\cite{quantumcomp}.
The distinctive feature of a reversible system is that the execution of
actions is liable to be undone.
{\em Reversible prime event structures} (\rpes)~\cite{PU:jlamp15} accommodate
the undoing of executed actions by allowing configurations to evolve by
removing events. For instance, if $c$ were an undoable event of the event
structure $\mathsf{P}$ in \Cref{fig:exintro-es}, then the associated
transition system would include the transition $\{a,c\} \rightarrow \{a\}$.
This is a disruptive feature in event structures since it breaks the
underlying assumption by which configurations evolve by adding events. In
fact, if $X$ and $Y$ are two configurations of an $\rpes$ then
$X \rightarrow Y$ does not imply $X\subseteq Y$.
As a consequence, the existing approaches to recover Petri nets out of event
structures, even the most general ones~\cite{GP:ESRC}, are not applicable.
As a matter of fact, we still lack a procedure to associate a Petri net to a
given \rpes.
Indeed, the approach presented here stands as the first to provide a
systematic procedure for associating a Petri net with a given \rpes.
Previous attempts~\cite{MMPPU:RC20} do this job just for the subclass of {\em
  cause-respecting} \rpes{}es, i.e., \rpes{}es that allow the reversing of an
event once all events it caused have been reversed.
For instance, the transition system associated with a cause-respecting
reversible version of the event structure $P$ in \Cref{fig:exintro-es} is
depicted in \Cref{fig:exintro-conf-rev}.
Note that each transition $X\rightarrow Y$ is paired with a reversing one
$Y \rightarrow X$; consequently, the configuration $\{a,b\}$ can be reversed
only by undoing first $b$ and then $a$, i.e.,
$\{a,b\} \rightarrow \{a\} \rightarrow \emptyset$.
Contrastingly, the transition $\{a,b\} \rightarrow \{b\}$ is not included
because it accounts for the reversal of $a$ before the reversal of the event
$b$, which causally depends on $a$.

As shown in~\cite{MMPPU:RC20}, the transition system of a {\em
  cause-respecting} \rpes can be implemented (concurrently / distributedly) as
a Petri net where the undoing of events is achieved via {\em reversing}
transitions, i.e., each transition $t$ (corresponding to some event of the
\rpes) is accompanied by another transition $\un t$ that undoes the effects of
the firing of $t$, i.e., $\un t$ (i) consumes the tokens produced by $t$; and
(ii) produces the tokens consumed by $t$.
The transition system in~\Cref{fig:exintro-conf-rev} is implemented by the net
$\un N$ in \Cref{fig:exintro-pn-rev}, which is essentially the extension of
$N$ (\Cref{fig:exintro-pn}) with the reversing transitions $\un a$, $\un b$
and $\un c$.

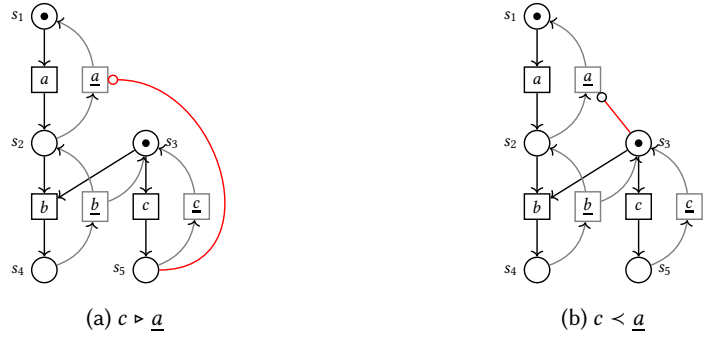
\begin{figure}[t]
  \begin{subfigure}{.2\textwidth}
    \begin{center}
      \scalebox{0.75}{\scalebox{0.9}{\begin{tikzpicture}
\tikzstyle{inhibitorred}=[o-, draw=red,thick]
\tikzstyle{inhibitorblu}=[o-, draw=blue,thick]
\tikzstyle{pre}=[<-,thick]
\tikzstyle{post}=[->,thick]
\tikzstyle{readblue}=[-, draw=blue,thick]
\tikzstyle{transition}=[rectangle, draw=black,thick,minimum size=5mm]
\tikzstyle{place}=[circle, draw=black,thick,minimum size=5mm]
\tikzstyle{rev}=[rectangle, draw=gray,thick,minimum size=5mm]
\tikzstyle{prerev}=[<-,thick,draw=gray]
\tikzstyle{postrev}=[->,thick,draw=gray]

\node[place,tokens=1] (p1) at (0,5) [label=left:$s_1$] {};
\node[place,tokens=1] (p5) at (2,2.5) [label=right:$s_3$] {};
\node[place] (p2) at (0,2.5) [label=left:$s_2$] {};
\node[place] (p4) at (0,0) [label=left:$s_4$] {};
\node[place] (p6) at (2,0) [label=left:$s_5$] {};
\node[transition] (a) at (0,3.75)  {$a$}
edge[pre] (p1)
edge[post](p2);
\node[rev] (ra) at (1,3.75)  {$\un a$}
edge[postrev, bend right] (p1)
edge[prerev, bend left](p2)
;
\draw[o-, thick,red] (1.25,3.75)  [out = 0, in = 0, looseness = 1.5] to (p6);

\node[transition] (b) at (0,1.25) {$b$}
edge[pre] (p2)
edge[pre] (p5)
edge[post] (p4)
;
\node[rev] (rb) at (1,1.25) {$\un b$}
edge[postrev, bend right] (p2)
edge[postrev, bend right] (p5)
edge[prerev, bend left] (p4)
;

\node[transition] (c) at (2,1.25)  {$c$}
edge[pre] (p5)
edge[post] (p6);

\node[rev] (rc) at (3,1.25)  {$\un c$}
edge[postrev, bend right] (p5)
edge[prerev, bend left] (p6);
\end{tikzpicture}}}
    \end{center}
    \caption{$c \pr \un a$}\label{fig:exintro-pn-rev-prev}
  \end{subfigure}
  \hspace{3cm}
  \begin{subfigure}{.2\textwidth}
    \begin{center}
      \scalebox{0.75}{\scalebox{0.9}{\begin{tikzpicture}
\tikzstyle{inhibitorred}=[o-, draw=red,thick]
\tikzstyle{inhibitorblu}=[o-, draw=blue,thick]
\tikzstyle{pre}=[<-,thick]
\tikzstyle{post}=[->,thick]
\tikzstyle{readblue}=[-, draw=blue,thick]
\tikzstyle{transition}=[rectangle, draw=black,thick,minimum size=5mm]
\tikzstyle{place}=[circle, draw=black,thick,minimum size=5mm]
\tikzstyle{rev}=[rectangle, draw=gray,thick,minimum size=5mm]
\tikzstyle{prerev}=[<-,thick,draw=gray]
\tikzstyle{postrev}=[->,thick,draw=gray]

\node[place,tokens=1] (p1) at (0,5) [label=left:$s_1$] {};
\node[place,tokens=1] (p5) at (2,2.5) [label=right:$s_3$] {};
\node[place] (p2) at (0,2.5) [label=left:$s_2$] {};
\node[place] (p4) at (0,0) [label=left:$s_4$] {};
\node[place] (p6) at (2,0) [label=right:$s_5$] {};
\node[transition] (a) at (0,3.75)  {$a$}
edge[pre] (p1)
edge[post](p2);
\node[rev] (ra) at (1,3.75)  {$\un a$}
edge[postrev, bend right] (p1)
edge[prerev, bend left](p2)
edge[inhibitorred](p5);

\node[transition] (b) at (0,1.25) {$b$}
edge[pre] (p2)
edge[pre] (p5)
edge[post] (p4)
;
\node[rev] (rb) at (1,1.25) {$\un b$}
edge[postrev, bend right] (p2)
edge[postrev, bend right] (p5)
edge[prerev, bend left] (p4)
;

\node[transition] (c) at (2,1.25)  {$c$}
edge[pre] (p5)
edge[post] (p6);

\node[rev] (rc) at (3,1.25)  {$\un c$}
edge[postrev, bend right] (p5)
edge[prerev, bend left] (p6);
\end{tikzpicture}}}
    \end{center}
    \caption{$c \prec \un a$}\label{fig:exintro-pn-rev-revcaus}
  \end{subfigure}
  \caption{Prevention and reverse causality operationally}
  \Description{Prevention and reverse causality operationally}
\end{figure}

This approach however falls short when addressing the full expressivity of
\rpes{}es, which accommodates different flavours of
reversibility~\cite{wg1,wg2}. The reversing mechanism of an \rpes{} is defined
in terms of two relations (additional to the classical causality and
conflicts): {\em prevention} and {\em reverse causality}.
For instance, an \rpes can stipulate that some event can be undone only when
some other events have not occurred (prevention). For instance, if we
stipulate that $c$ prevents the undoing of $a$ (written $c \pr \underline a$),
then $\{a,c\} \rightarrow \{c\}$ is banned from the transition system even
though $a$ is undoable and $c$ does not causally depend on $a$.
We can also specify that a particular event can be undone only when some other
events have already occurred (reverse causality). For instance, if $c$ is a
reverse cause of $a$ (written $c\prec \underline a$) then $a$ cannot be
reversed until $c$ occurs, i.e., the transition $\{a\} \rightarrow \emptyset$
is not admissible.
We note that these constraints can be translated into Petri nets in the form
of contextual arcs; in particular, inhibitor arcs~\cite{MR:CN,BBCP:rivista}
that prevent the firing of a transition if a token is present in some place of
the net.
For instance, the prevention $c \pr \un a$ can be represented in a Petri net
with an inhibitor arc in $\un a$, as shown in \Cref{fig:exintro-pn-rev-prev};
the added arc (depicted as $\multimap$) forbids the firing of $\un a$ when
$s_5$ contains a token. Note that $s_5$ contains a token only when $c$ has
been fired, hence $\un a$ cannot be fired if $c$ has occurred.
Along the same lines, the reverse causality $c \prec \un a$ can be represented
as shown in \Cref{fig:exintro-pn-rev-revcaus}: in this case, the inhibitor arc
is connected to $s_3$, which will contain a token if $c$ has not been fired.
Hence, $\un a$ will be enabled only after $c$ is fired.

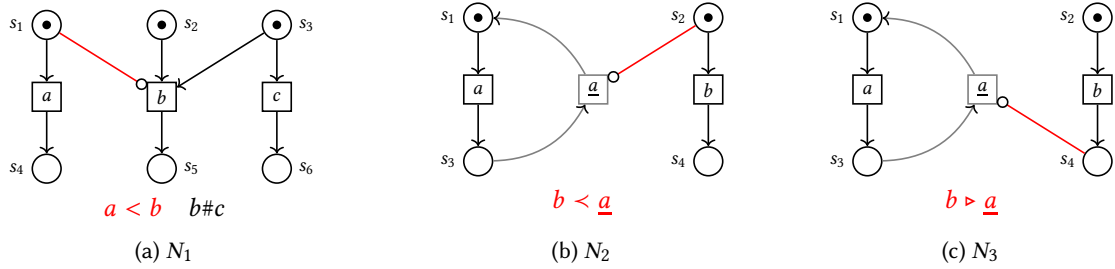
\begin{figure*}[t]
  \begin{subfigure}{.32\textwidth}
    \begin{center}
      \scalebox{0.85}{\scalebox{0.9}{\begin{tikzpicture}
\tikzstyle{inhibitorred}=[o-, draw=red,thick]
\tikzstyle{inhibitorblu}=[o-, draw=blue,thick]
\tikzstyle{pre}=[<-,thick]
\tikzstyle{post}=[->,thick]
\tikzstyle{readblue}=[-, draw=blue,thick]
\tikzstyle{transition}=[rectangle, draw=black,thick,minimum size=5mm]
\tikzstyle{place}=[circle, draw=black,thick,minimum size=5mm]
\node[place,tokens=1] (p1) at (0,2.5) [label=left:$s_1$] {};
\node[place,tokens=1] (p3) at (2,2.5) [label=right:$s_2$] {};
\node[place,tokens=1] (p5) at (4,2.5) [label=right:$s_3$] {};
\node[place] (p2) at (0,0) [label=left:$s_4$] {};
\node[place] (p4) at (2,0) [label=right:$s_5$] {};
\node[place] (p6) at (4,0) [label=right:$s_6$] {};
\node[transition] (a) at (0,1.25)  {$a$}
edge[pre] (p1)
edge[post](p2);

\node[transition] (b) at (2,1.25) {$b$}
edge[pre] (p3)
edge[pre] (p5)
edge[post] (p4)
edge[inhibitorred] (p1)
;
\node[transition] (c) at (4,1.25)  {$c$}
edge[pre] (p5)
edge[post] (p6);
\end{tikzpicture}}}\\
	       {\textcolor{red}{$a < b$}$ \quad  b \# c$}
    \end{center}
    \caption{$N_1$\label{ex:introa}}
  \end{subfigure}\hfill
  \begin{subfigure}{.32\textwidth}
    \begin{center}
      \scalebox{0.85}{\scalebox{0.9}{\begin{tikzpicture}
\tikzstyle{inhibitorred}=[o-, draw=red,thick]
\tikzstyle{inhibitorblu}=[o-, draw=blue,thick]
\tikzstyle{pre}=[<-,thick]
\tikzstyle{post}=[->,thick]
\tikzstyle{readblue}=[-, draw=blue,thick]
\tikzstyle{transition}=[rectangle, draw=black,thick,minimum size=5mm]
\tikzstyle{place}=[circle, draw=black,thick,minimum size=5mm]
\tikzstyle{rev}=[rectangle, draw=gray,thick,minimum size=5mm]
\tikzstyle{prerev}=[<-,thick,draw=gray]
\tikzstyle{postrev}=[->,thick,draw=gray]

\node[place,tokens=1] (p1) at (0,2.5) [label=left:$s_1$] {};
\node[place,tokens=1] (p2) at (4,2.5) [label=left:$s_2$] {};
\node[place] (p3) at (0,0) [label=left:$s_3$] {};
\node[place] (p4) at (4,0) [label=left:$s_4$] {};
\node[transition] (a) at (0,1.25)  {$a$}
edge[pre] (p1)
edge[post](p3);

\node[transition] (b) at (4,1.25) {$b$}
edge[pre] (p2)
edge[post] (p4)
;

\node[rev] (nota) at (2,1.25) {$\underline a$}
edge[inhibitorred] (p2)
edge[prerev, bend left] (p3)
edge[postrev, bend right] (p1)
;

\end{tikzpicture}}}\\
	       {\textcolor{red}{$b \prec \underline a$}}
    \end{center}
    \caption{$N_2$\label{ex:introb}}
  \end{subfigure}
  \begin{subfigure}{.32\textwidth}
    \begin{center}
      \scalebox{0.85}{\scalebox{0.9}{\begin{tikzpicture}
\tikzstyle{inhibitorred}=[o-, draw=red,thick]
\tikzstyle{inhibitorblu}=[o-, draw=blue,thick]
\tikzstyle{pre}=[<-,thick]
\tikzstyle{post}=[->,thick]
\tikzstyle{readblue}=[-, draw=blue,thick]
\tikzstyle{transition}=[rectangle, draw=black,thick,minimum size=5mm]
\tikzstyle{place}=[circle, draw=black,thick,minimum size=5mm]
\tikzstyle{rev}=[rectangle, draw=gray,thick,minimum size=5mm]
\tikzstyle{prerev}=[<-,thick,draw=gray]
\tikzstyle{postrev}=[->,thick,draw=gray]

\node[place,tokens=1] (p1) at (0,2.5) [label=left:$s_1$] {};
\node[place,tokens=1] (p2) at (4,2.5) [label=left:$s_2$] {};
\node[place] (p3) at (0,0) [label=left:$s_3$] {};
\node[place] (p4) at (4,0) [label=left:$s_4$] {};
\node[transition] (a) at (0,1.25)  {$a$}
edge[pre] (p1)
edge[post](p3);

\node[transition] (b) at (4,1.25) {$b$}
edge[pre] (p2)
edge[post] (p4)
;

\node[rev] (nota) at (2,1.25) {$\underline a$}
edge[inhibitorred] (p4)
edge[prerev, bend left] (p3)
edge[postrev, bend right] (p1)
;

\end{tikzpicture}}}\\
	       {\textcolor{red}{$b \pr \underline a$}}
    \end{center}
    \caption{$N_3$\label{ex:introc}}
  \end{subfigure}
  \label{fig:exintro}
  \caption{Examples on how inhibitor acts ($\multimap$ arcs) can be used to
    model causality $<$, reverse causality $\prec$ and prevention
    $\pr$ relation among events}
    \Description{Examples on how inhibitor acts ($\multimap$ arcs) can be used to
    model causality $<$, reverse causality $\prec$ and prevention
    $\pr$ relation among events}
\end{figure*}

Unfortunately, the previous observation is insufficient for capturing the full
spectrum of \rpes{}es due to the interplay among causality, prevention and
reverse causality. This becomes clear when addressing \rpes{}es enjoying
\emph{out-of-causal} order reversibility, which is typical in bio-chemical
reactions~\cite{PhillipsUY12}.
Consider again the reversible system in \Cref{fig:exintro-es}. Assume now that
$a$ can be undone also in an out-of-causal order fashion, i.e., $a$ can be
reversed independently of the events that it may have caused (which in this
case is $b$).
Hence, the transition system would be extended to include the transition
$\{a,b\} \rightarrow \{b\}$, in which $a$ is reversed even though $b$ is not,
and also $\{b\} \rightarrow \{b,a\}$, in which the minimal event $a$ is
executed.
When looking at the net $\un N$ in \Cref{fig:exintro-pn-rev}, the transition
$\{a,b\} \rightarrow \{b\}$ would require to be able to fire the reversing
transition $\un a$ also when the place $s_2$ does not contain any token
(because the firing of $b$ has consumed that token). Hence, a more involved
definition of $\un a$ would be needed for handling the undoing of $a$.
Moreover, after reversing $a$ we should be able to fire $a$ again, since the
transition system associated to the \rpes allows both
$\{a\}\rightarrow \emptyset \rightarrow \{a\}$ and
$\{a,b\}\rightarrow \{b\} \rightarrow \{a,b\}$.
It should be noted that the execution of $a$ has different effects in the two
computations above: while the configuration $\{a\}$ allows for the firing of
$b$, the configuration $\{a,b\}$ does not (because, $b$ has been already
fired).
The way in which the causality relation between $a$ and $b$ is described in
$\un N$ (\Cref{fig:exintro-pn-rev}) would be insufficient to distinguish the
cases above.

In this paper, we take a different approach by observing that inhibitor arcs
can be used to model also causality.
This simple idea is rendered by the net in \Cref{ex:introa}, which is an
operational counterpart of the event structure $\mathsf{P}$ in
\Cref{fig:exintro-es}.
The inhibitor arc in \Cref{ex:introa} is used to model causality among the
events $a$ and $b$. Indeed, $b$ can happen only after $a$ has happened, hence
$a<b$.
As previously discussed, we represent prevention and reverse causality with
inhibitor arcs, as illustrated in \Cref{ex:introb,ex:introc}.
On the one hand, the inhibitor arc in \Cref{ex:introb} models the reverse
causality $b \prec \underline a$, i.e., $a$ can be reversed only when $b$ has
been executed. On the other hand, the inhibitor arc in \Cref{ex:introc} models
prevention $b\pr \underline a$, i.e., the undoing of $a$ cannot be executed if
the event $b$ has happened.
A liberal usage of inhibitor arcs would not do the work. Therefore, we impose
some (structural) constraints on nets to achieve our purpose, namely to
identify a subclass of nets with inhibitor arcs that corresponds to reversible
prime structures.


The main contribution of our work, which is a revised and extended version of
\cite{lics}, is the definition of an operational interpretation of \rpes{}es
in terms of a proper subclass of Petri nets with inhibitor arcs, called {\em
reversible causal nets}.
The considered subclass is expressive enough to recover the classical
operational interpretation of \pes{es} in term of Petri nets, meaning that
this is a conservative extension of the classic notion of occurrence net.
The novel way to model dependencies using inhibitor arcs can be fruitfully
used to model non standard dependencies among events, e.g., asymmetric
conflicts.


\subsection*{Structure of the paper}
This paper is structured as follows: after setting some notions that will be used in the paper,
we start by recalling the basics
of prime event structures and reversible prime event structures
(\Cref{sec:es}), and those of nets with inhibitor arcs (\Cref{sec:nets}).
Then, we introduce \emph{causal nets}, which are subclass of nets with
inhibitor arcs, and show that they are a suitable counterpart of prime
event structures (\Cref{ssec:causal-nets}).
In \Cref{sec:rpesandnets}, we introduce a \emph{reversible} notion of
causal nets and prove that they fully correspond to reversible prime
event structures, thus giving a proper operational model of them.
In \Cref{sec:discussion} we compare and discuss the relationship
between a more classical operational model for prime event structures,
namely occurrence nets, with our proposal, and illustrates the reasons
that undermine the possibility of generalising occurrence nets to cope
with richer models of reversibility as, e.g., out-of-causal order
reversibility. 
\Cref{sec:app} illustrates some real use of our approach and the last section, after 
recapping our achievements, provides a brief comparison with the results we 
extend.

\section{Preliminaries} \label{sec:preliminaries}

We begin by revisiting some key concepts that will be employed throughout this
paper.
The symbol $\nat$ signifies the set of natural numbers.
A \emph{multiset} over a set $A$ is a function $m: A \rightarrow \nat$.
Multisets are assumed to be equipped with the standard operations of union
($+$) and difference ($-$). We write $m \subseteq m'$ if $m(a) \leq m'(a)$ for
all $a \in A$.
The multiset $\flt{m}$ is defined such that $\flt{m}(a) = 1$ if $m(a) > 0$ and
$\flt{m}(a) = 0$ otherwise.
We often confuse a multiset $m$ with the set $\setcomp{a\in A}{m(a) \neq 0}$
when $m = \flt{m}$. In such cases, $a\in m$ denotes $m(a) \neq 0$, and
$m\subseteq A$ signifies that $m(a) = 1$ implies $a\in A$ for all $a$.
The underlying set of a multiset $m$, namely the one formed by the elements
$a$ with $m(a)$, is precisely $\flt{m}$.
Additionally, we will employ standard set operations like $\cap$, $\cup$, or
$\setminus$.
The set of all multisets over $A$ is denoted as $\mu A$; the symbol $\zero$
stands for the unique multiset defined such that $\flt{\zero} = \emptyset$.

Given a function $f: A \to B$, the domain of $f$ is defined as
$\mathit{dom}(f) = \setcomp{a\in A}{\exists b\in B.\ f(a) = b}$.
A sequence of elements of $A$ is a mapping $\rho: \nat\rightarrow A$ such that
if $n\in \mathit{dom}(\rho)$, then $\forall n' < n.\ n'\in\mathit{dom}(\rho)$.
A sequence $\rho$ is often represented as $a_1a_2\cdots$ where $a_i =
\rho(i)$.
The length of a sequence $\rho$, denoted by $\len{\rho}$, is the cardinality
of its domain, i.e., $\len{\rho} = |\mathit{dom}(\rho)|$. A sequence is
considered finite when its length is finite.
Requiring that a sequence $\rho$ has distinct elements implies asserting that
$\rho$ is injective on $\mathit{dom}(\rho)$.

A binary relation $<\ \subseteq A\times A$ is an {\em irreflexive partial
order} when it is irreflexive and transitive, and we use $\leq$ to denote its
reflexive closure.
We write $\histtwo{a}{<}$ for the set $\setcomp{a' \in A}{a'<
a}\cup\setenum{a}$ and shall omit the subscript $<$ when it is clear from the
context.

\section{Event structures} \label{sec:es}

In this section, we provide an overview of the fundamentals of \emph{prime}
event structures. Subsequently, we delve into the \emph{reversible} variant of
prime event structures, by following the presentation in~\cite{PU:jlamp15}.


\subsection{pre-{\pes} and \pes}
\label{ssec:prepes}

Pre-\emph{prime event structures} represent a relaxed form of prime event
structures where conflict heredity may not necessarily hold. They play a key
role in the definition of reversible prime event structures.

\begin{definition} \label{de:pre-pes}
  A \emph{pre-prime event structure} (p\pes) is a triple $\mathsf{P} = (E, <,
  \#)$, where

  \begin{enumerate}
  \item
    $E$ is a countable set of \emph{events};
  \item
    $\#\ \subseteq E\times E$ is an irreflexive and symmetric relation,
    called the \emph{conflict relation}; and
  \item
    $<\ \subseteq E\times E$ is an irreflexive partial order, called the
    \emph{causality relation}, defined such that $\forall e\in E$.
    $\histtwo{e}{<}$ is finite and $\forall e', e''\in \histtwo{e}{<}$.
    $\neg (e'\ \#\ e'')$.
  \end{enumerate}

  We say a p\pes\ $\mathsf{P}$ is a \emph{prime event structure} (\pes) when
  $\#$ is \emph{hereditary} with respect to $<$, i.e., if $e\ \#\ e' <
  e''$ then $e\ \#\ e''$ for all $e,e',e'' \in E$.
\end{definition}

The third condition prevents events from depending on an infinite number of
causes, rendering their execution unfeasible. It also prohibits conflicts
between two causes of the same event, which would likewise make the occurrence
of the event unfeasible. Moreover, it ensures that an event is not in
conflict with any of its causes.

Conflict inheritance is intuitively clear in systems whose state evolves
monotonically. If an event $e$ is in conflict with another event $e'$ that
causes a third event $e''$, then $e$ would necessarily be in conflict with
$e''$. This is because $e''$ cannot occur without $e'$, and the conflict
between $e$ and $e'$ prevents the occurrence of $e$. However, this
straightforward relationship breaks down in situations where certain events
can be undone, as will be discussed further in \Cref{sec:rpes-rpes}.

\begin{example}\label{ex:ppes}
  Let ${P} = (E,<,\#)$ be defined such that
  \[
    \begin{array}{l@{\hspace{2cm}}l@{\hspace{2cm}}l}
      E = \{a, b, c, d\}
      &
        {\#} =\{(a, b), (b, a)\}
      &
        {<} = \{(b, c)\}
    \end{array}
  \]

  It is immediate to check that $\#$ is irreflexive and symmetric, and $<$ is
  an irreflexive partial order.
  Furthermore, if $e \in E$ and $e \neq c$, then $\histtwo{e}{<} = {e}$---a
  finite and conflict-free set.
  Additionally, $\histtwo{c}{<} = \{b, c\}$ is also finite and conflict-free.
  Hence, $\mathsf{P}$ is a p\pes.
  However, $\mathsf{P}$ is not a \pes\ because conflicts are not inherited
  along $<$. Indeed, while $a\ \#\ b < c$ holds, the relation $a\ \# \ c$ does
  not.
  $\mathsf{P}$ would be a \pes\ if $\#$ were defined as
  $\{(a, b), (b, a), (a, c), (c, a)\}$.
\end{example}

A set of events $X\subseteq E$ is \emph{conflict-free} if for every pair of
events $e$ and $e'$ in $X$, it holds that $\neg(e\ \#\ e')$. We write $\CF{X}$
if $X$ is conflict-free.
Note that for any subset $Y$ of $X$, if $X$ is conflict-free ($\CF{X}$), then
$Y$ is also conflict-free (i.e., $\CF{Y}$).

A configuration is a conflict-free set of events, encompassing, for each event
$e$, the set of its causes, namely $\histtwo{e}{<}$---as formally defined
below.

\begin{definition}\label{de:ppes-conf}
  Let $\mathsf{P} = (E, <, \#)$ be a p\pes\ and $X\subseteq E$ be a subset of
  events. We say that $X$ is a \emph{configuration} if
  \begin{enumerate}
  \item
    $\CF{X}$, and
  \item
    $\forall e\in X$. $\histtwo{e}{<}\subseteq X$.
  \end{enumerate}

  We denote the set of all configurations of $\mathsf{P}$ as
  $\Conf{P}{p\pes}$. When $\mathsf{P}$ is a \pes, we write $\Conf{P}{\pes}$
  instead of $\Conf{P}{p\pes}$.
\end{definition}

The subsequent definition introduces the concept of (labeled) transitions
between sets of events in a p\pes.

\begin{definition}\label{de:ppes-conf-enab}
  Let $\mathsf{P} = (E, <, \#)$ be a p\pes\ and $X\subseteq E$ a conflict-free
  set of events. We say $A\subseteq E$ is \emph{enabled} at $X$ if
  \begin{enumerate}
  \item
    $A\cap X = \emptyset$ and $\CF{X\cup A}$, and
  \item
    $\forall e\in A$. if $e' < e$ then $e'\in X$.
  \end{enumerate}

  If $A$ is enabled at $X$, then $X \toconf{A}{p\pes} Y$
  with $Y = X\cup A$.
\end{definition}

Intuitively, a set of events that are not present in $X$ is considered enabled
if there are no conflicts among them or with the events in $X$, and all their
causes are already included in $X$.
It is worth observing that condition $(2)$ in the preceding definition can be
formulated as follows: for every event $e$ in $A$, it holds that
$\histtwo{e}{<}\setminus\setenum{e}\subseteq X$, i.e., all its causes are in
$X$.

\begin{example}
  Consider the p\pes{} $\mathsf{P}$ in \Cref{ex:ppes}.
  The sets $\{a\}$, $\{b\}$, $\{d\}$, $\{a,d\}$, and $\{b, d\}$ are all
  enabled at $\emptyset$ because they are all conflict-free and they contain
  just minimal elements (according to $<$).
  This enables the derivation of transitions such as
  $\emptyset \toconf{\{a\}}{p\pes} \{a\}$ and
  $\emptyset \toconf{\{b,d\}}{p\pes} \{b,d\}$.
  Conversely,  neither
  $\{a,b\}$ nor $\{c\}$ are enabled at $\emptyset$. The former, because $a$
  and $b$ are in conflict; the latter because $\emptyset$ does not contain
  $b$, which is a cause of $c$.
  Moreover, $\{c\}$ is enabled at $\{b\}$, because the unique cause of $c$ is
  $b$; consequently, $\{b\} \toconf{\{c\}}{p\pes} \{b,c\}$ holds.
\end{example}

The notion of configurations of a p\pes{} can be reformulated based on the
transitions of enabled events.

\begin{definition}\label{de:ppes-forwconf}
  Let $\mathsf{P} = (E, <, \#)$ be a p\pes. A set of events $X\subseteq E$ is
  a \emph{reachable configuration} if it is conflict-free, i.e., $\CF X$, and
  there exists a sequence {$A_1,A_2,A_3, \dots$} of nonempty set of events
  such that
  \begin{enumerate}
   \item $X_0 = \emptyset$, $X_i = \bigcup_{j\leq i}A_i$,
   \item $X_i\toconf{A_i}{\pes} X_{i+1}$ for all $i$, and
   \item $X = \bigcup_{i}A_i$.
  \end{enumerate}
\end{definition}

In simpler terms, a reachable configuration is constructed step by step,
adding enabled sets of events in each step.

\begin{proposition}
  Let $\mathsf{P} = (E, <, \#)$ be a p\pes{} and $X\subseteq E$. Then, $X$ is
  a configuration iff it is a reachable configuration.
\end{proposition}

\begin{proof}
  \begin{itemize}
  \item[$\Rightarrow$)] We need to demonstrate that a reachable configuration
    $X$ is indeed a configuration.
    Given that $X$ is a reachable configuration, there exists a sequence
    $A_1, A_2, A_3, \dots$ of sets of events such that $X = \bigcup_{i}A_i$,
    $X_{i-i} = \bigcup_{j<i}A_j$ and $X_{i-1}\toconf{A_i}{p\pes}X_i$.
    Since $X$ is conflict-free by definition, we need to show that
    $\histtwo{e}{<}\subseteq X$ for all $e\in X$.
    Suppose $e\in A_i$ for some $i$. As $X$ is a reachable configuration, we
    have $X_{i-1}\toconf{A_i}{p\pes}X_i$ and $X_{i-1} \subseteq X$. Due to
    $X_{i-1}\toconf{A_i}{p\pes}X_i$, all $e' < e$ are in $X_{i-1}$ and,
    consequently, in $X$. Thus, the thesis holds.

  \item[$\Leftarrow$)] We need to show that a configuration $X$ is indeed a
    reachable configuration.
    As $X$ is a configuration, its elements can be totally ordered with respect
    to $<$, resulting in the sequence $e_0,e_1,e_2, \dots$. By defining
    $A_i=\setenum{e_i}$, we demonstrate the thesis. In fact each subset of $X$
    is conflict-free, and for all $i$,
    $X_{i-1} = \setenum{e_0,e_1,e_2, \dots, e_{i_1}}$ contains all events
    preceding $e_i$. Additionally, $X_{i-1} \cup A_i = X_i$ is conflict-free,
    and therefore, $X_{i-1}\toconf{A_i}{p\pes}X_i$.
  \end{itemize}
\end{proof}

\begin{example}
  Consider once again the p\pes{} $\mathsf{P}$ in \Cref{ex:ppes}.
  It is noteworthy that certain conflict-free sets of events do not correspond
  to states in the computation of $\mathsf{P}$. For example, the set
  $\{c, d\}$---despite being conflict-free---is not a configuration of
  $\mathsf{P}$. This is due to the fact that it cannot be reached from the
  initial state $\emptyset$: the introduction of $c$ to a configuration
  necessitates the presence of $b$.
\end{example}

The following definition and result from~\cite{PU:jlamp15} highlight
that
we can reconstruct a \pes from a p\pes\ by enforcing the heredity of conflicts.

\begin{definition}\label{de:hereditary-closure}
  Let $\mathsf{P} = (E, <, \#)$ be a p\pes. Then
  $\mathsf{hc}(\mathsf{P}) = (E, <, \sharp)$ is the \emph{hereditary closure}
  of $\mathsf{P}$, where $\sharp$ is inductively defined by the following
  rules
  \[
  \begin{array}{ccccc}
    \infer{e\ \sharp\ e'}{e\ \#\ e'}
    & \hspace*{1cm}
    & \infer{e\ \sharp\ e''}{e\ \sharp\ e'\ \ \ \ \ e' < e''}
    & \hspace*{1cm}  & \infer{e\ \sharp\ e'}{e'\ \sharp\ e}\\
  \end{array}
  \]
\end{definition}

\begin{proposition}\label{pr:ppes-prop}
 Let $\mathsf{P} = (E, <, \#)$ be a p\pes, then
 \begin{itemize}
   \item $\mathsf{hc}(\mathsf{P}) = (E, \leq, \sharp)$ is a \pes,
   \item if $\mathsf{P}$ is a \pes, then
     $\mathsf{hc}(\mathsf{P}) = \mathsf{P}$, and
   \item $\Conf{\mathsf{P}}{p\pes} = \Conf{\mathsf{hc}(\mathsf{P})}{\pes}$.
 \end{itemize}
\end{proposition}


\subsection{Reversible prime event structures}
\label{sec:rpes-rpes}
We now revisit the concept of a \emph{reversible prime event structure}
presented in~\cite{PU:jlamp15}.
Reversible event structures extend \pes{}es by by introducing the possibility
for \emph{some} events to be \emph{reversible} or \emph{undoable}. A
reversible event, denoted as $\anr$, implicitly possesses a corresponding
reversing event, symbolized as $\underline{\anr}$, capable of undoing its
effects. In other words, the execution of $\anr$ followed by
$\underline{\anr}$ is indistinguishable from no execution at all.
Due to this characteristic, the configurations of a reversible prime event
structure may not evolve monotonically, as reversible events have the ability
to disappear because of the execution of their corresponding reversing events.

To accommodate various nuances of reversibility, a reversible prime event
structure incorporates two relations known as \emph{prevention} and
\emph{reverse causality}. These relations define the manner in which reversing
events can be executed.

\begin{definition}\label{de:rpes}
  A \emph{reversible prime event structure} (\rpes) is a tuple
  $\mathsf{P} = (E, \anR, <, \#, \prec, \triangleright)$ where $(E, <, \#)$ is
  a p\pes, $\anR\subseteq E$ are the \emph{reversible/undoable} events (with
  corresponding reverse events being denoted by
  $\underline{\anR} = \setcomp{\underline{\anr}}{\anr\in\anR}$ and disjoint
  from $E$, i.e., $\un \anR \cap E = \emptyset$) and
 \begin{enumerate}
 \item
   $\prec\ \subseteq E\times \underline{\anR}$ is the \emph{reverse
   causality} relation; which is defined such that $\anr\prec
   \underline{\anr}$ for all $\anr\in\anR$; and moreover $\setcomp{e\in
     E}{e\prec\underline{\anr}}$ is finite and conflict-free for all
   $\anr\in\anR$,
 \item
   $\triangleright\ \subseteq E\times \underline{\anR}$ is the
   \emph{prevention} relation defined such that ${\prec} \cap
        {\triangleright} = {\emptyset}$,
 \item
   the \emph{sustained causation} $\ll$ is a transitive relation
   defined such that if $e \ll e'$ then
   \begin{itemize}
     \item $e < e'$,
     \item if $e\in\anR$ then $e' \triangleright \underline{e}$,
  \end{itemize}
 \item
   $\#$ is hereditary with respect to $\ll$: if $e\ \#\ e' \ll e''$,
   then $e\ \#\ e''$.
 \end{enumerate}
\end{definition}

The reverse causality relation prescribes the events necessary for the
execution of each reversing event. In other words,
{$e \prec \underline{\anr}$} signifies that {$\underline{\anr}$} can be
executed (or equivalently, $\anr$ can be undone) only when the event $e$ is
present. Therefore, the condition $\anr\prec \underline{\anr}$ asserts that an
event $\anr$ can be undone solely when it is present in a configuration.
Conversely, the prevention captures cases where an event can be reversed only
if some other event is absent in the configuration. Specifically,
{$e\triangleright \underline{\anr}$} denotes that $\anr$ can be reversed only
if $e$ is not present in the configuration.

Despite the fact that the underlying structure $(E, <, \#)$ is a p\pes, where
conflicts may not necessarily be inherited through causality, the definition
mandates that conflicts must be inherited through the newly introduced
\emph{sustained causation} relation $\ll$. This relation, which is coarser
than $<$, accommodates situations where the causes of certain events may
vanish from a configuration, as illustrated in \Cref{ex:rpes-out-of-order}.

\begin{example}\label{ex:rpes-causal}
  Let $\mathsf{P_1} = (E, \anR, <, \#, \prec, \pr)$ be an \rpes\ defined
  such that
  \[
  \begin{array}{l@{\hspace{2cm}}l}
    E = \{a, b, c, d\}
    &
    \anR = \{b, c\}
    \\
    {<} = \{(b, c)\}
    &
    {\#} =\{(a, b), (b, a), (a, c), (c, a)\}
    \\
    {\prec} = \{(b, \un b), (c, \un c))\}
    &
    {\pr} = \{(c, \un b)\}
  \end{array}
  \]

  While $b$ and $c$ are reversible in $\mathsf{P_1}$, $a$ and $d$ are not
  because $a, d\not\in\anR$.
  Moreover, there is a causal dependency between $b$ and $c$ ($b < c$); and
  $a$ is in conflict with both $b$ and $c$ because of the definition of $\#$.
  As a result, $d$ is concurrent w.r.t. $a$, $b$ and $c$.
  In the reverse causality relation, each reversible event is undone only if
  it has been executed (i.e., $b \prec \un b$ and $c \prec \un c$).
  The prevention relation stipulates that $b$ cannot be reversed if $c$ has
  been executed, i.e., $c \pr \un b$, which is typical of causal reversible
  models.
  In this example, sustained causation coincides with causality
  (${\ll} = {<}$): specifically, $b \ll c$ holds because $b < c$ and
  $c \pr \un b $ implies so. Also, it is important to note that conflicts are
  inherited along sustained causation and, consequently, along causality, as
  demonstrated by $a\ \#\ b < c$ and $a\ \#\ c$.
\end{example}

\begin{example}\label{ex:rpes-not-causal}
  Consider $\mathsf{P_2} = (E, \anR, <, \#, \prec, \pr)$, a variant of
  $\mathsf{P_1}$ obtained by extending the definition of $\prec$ with the pair
  $(d, \un c)$. In this scenario, the event $c$ can only be reversed after the
  execution of $d$. This illustrates a case of non-causal reversibility, where
  concurrency interferes with reversibility: the reversal of $c$ is contingent
  on the execution of a concurrent (and hence, unrelated) event.
\end{example}

The following definition extends the notion of transitions between sets of
events to account for reversing events.

\begin{definition}\label{de:rpes-conf-enab}
  Let $\mathsf{P} = (E, \anR, <, \#, \prec, \triangleright)$ be an \rpes\ and
  $X\subseteq E$ a conflict-free set of events, i.e., $\CF{X}$. For
  $A\subseteq E$ and $B\subseteq \anR$, we say that $A\cup \underline{B}$ is
  \emph{enabled} at $X$ if
  \begin{enumerate}
  \item $A\cap X = \emptyset$, $B \subseteq X$ and $\CF{X\cup A}$,
  \item $\forall e\in A, e'\in E$. if $e' < e$ then $e'\in X\setminus B$,
  \item $\forall e\in B, e'\in E$. if $e' \prec \underline{e}$ then
        $e' \in X\setminus (B\setminus\setenum{e})$,
  \item $\forall e\in B, e'\in E$. if $e' \triangleright
    \underline{e}$ then $e'\not\in X\cup A$.
  \end{enumerate}

 If $A\cup \underline{B}$ is enabled at $X$ then
 $X \toconf{A\cup\underline{B}}{\rpes} Y$ with $Y = (X\setminus B)\cup A$.
\end{definition}
\begin{example}\label{ex:rpes-red}
  In the \rpes\ $\mathsf{P_1}$ from \Cref{ex:rpes-causal}, we have, e.g., the
  following transitions:
  \begin{itemize}
  \item
    $\emptyset \toconf{\{a\}}{\rpes} \{a\} \toconf{\{d\}}{\rpes} \{a, d\}$,
  \item
    $\emptyset \toconf{\{b,d\}}{\rpes} \{b,d\} \toconf{\{\un b\}}{\rpes}
    \{d\}$, and
  \item
    $\emptyset \toconf{\{b\}}{\rpes} \{b\} \toconf{\{c\}}{\rpes} \{b, c\}
    \toconf{\{d, \un c\}}{\rpes} \{b,d\} \toconf{\{c\}}{\rpes} \{b,c,d\}$.
  \end{itemize}

  Now, consider the \rpes\ $\mathsf{P_2}$ presented in
  \Cref{ex:rpes-not-causal}. Notice that the set $\{d, \un c\}$ is not enabled
  at $\{b, c\}$. Despite the presence of $c$, the reverse causality
  $d \prec \un c$ indicates that $d$ is also required for the reversal of $c$.
\end{example}

The reachable configurations of an \rpes are the sets of events that can be
reached from the empty set by performing events or undoing previously
performed events, as stated below.

\begin{definition}\label{de:rpes-conf}
  Let $\mathsf{P} = (E, \anR, <, \#, \prec, \triangleright)$ be an \rpes\ and
  $X\subseteq E$ a conflict-free set of events, i.e., $\CF{X}$ holds. We say
  that $X$ is a \emph{(reachable) configuration} if there exist two sequences
  of sets $A_i$ and $B_i$, for $i=1,\ldots,n$, such that
  \begin{enumerate}
  \item $A_i\subseteq E$ and $B_i\subseteq U$ for all $i$, and
  \item $X_i\toconf{A_i\cup\underline{B_i}}{\rpes} X_{i+1}$ for all $i$ with
    $X_1 = \emptyset$ and $X_{n+1}=X$.
 \end{enumerate}

 The set of configurations of $\mathsf{P}$ is denoted by
 $\Conf{\mathsf{P}}{r\pes}$.
\end{definition}

\begin{definition}\label{de:rpesequiv}
  Let $\mathsf{P}_1$ and $\mathsf{P}_2$ be two \rpes{}es. We say that they are
  \emph{equivalent}, written $\mathsf{P}_1 \equiv \mathsf{P}_2$, if they have
  the same set of configurations, i.e.,
  $\Conf{\mathsf{P}_1}{r\pes} = \Conf{\mathsf{P}_2}{r\pes}$.
\end{definition}

\begin{example}
  Consider the r\pes{} $\mathsf{P_1}$ introduced in \Cref{ex:rpes-causal}. The
  set of configurations of $\mathsf{P_1}$ is
  \[ \Conf{\mathsf{P_1}}{r\pes} = \setenum{\emptyset, \setenum{a},
      \setenum{b}, \setenum{d}, \setenum{a,d}, \setenum{b,d}, \setenum{b,c},
      \setenum{b,c,d}}\]

  (Refer to \Cref{ex:rpes-red} for the corresponding sequences of
  transitions.)
\end{example}

The out-of-causal-order reversibility of \rpes{}es poses a significant
challenge when trying to integrate it into an operational model, as
illustrated by the next example.

\begin{example}\label{ex:rpes-out-of-order}
  Consider $\mathsf{P_3} = (E, \anR, <, \#, \prec, \pr)$ as a variant of the
  \rpes $\mathsf{P_2}$ in \Cref{ex:rpes-not-causal}, defined as follows
  \[
    \begin{array}{l@{\hspace{2cm}}l}
      E = \{a,b,c,d\}
      & \anR = \{b,c\} \\
      {<} = \{(b,c)\}
      &
        {\#} =\{(a,b),(b,a)\}
      \\
      {\prec} = \{(b,\un b), (c,\un c), (d,\un c)\}
      &
        {\pr} = \emptyset
    \end{array}
  \]

  In this scenario, the induced sustained causation is empty, meaning
  $b\not\ll c$ holds because $b < c$ does so and $c \pr \un b$ does not.
  Since $c \pr \un b$ is not satisfied in $\mathsf{P_3}$, the reversal of $b$
  can occur even after executing $c$, which, in turn, depends causally on $b$.
  For instance, we can observe the sequence:
  $\emptyset \toconf{\{b\}}{\rpes} \{b\} \toconf{\{c\}}{\rpes}
  \{b,c\}\toconf{\{\un b\}}{\rpes}\{c\} \toconf{\{a\}}{\rpes}\{a,c\}$.
  It is important to note that the configuration $\{a, c\}$ would be
  prohibited in standard \pes{}es because $a\ \#\ b < c$ holds. However,
  \rpes{}es permit such configurations since conflicts are not necessarily
  inherited through causality, crucial for accommodating out-of-causal-order
  reversibility.
\end{example}

The previous examples highlight the primary challenges to address when
formulating a net semantics for \rpes{}es. These challenges can be summarised
as follows:
\begin{itemize}
\item
  configurations may drop events during computation, i.e.,
  $\{b,c\}\toconf{\{\un b\}}{\rpes}\{c\}$ in $\mathsf{P_3}$;
\item
  reachable configurations may not necessarily encompass all the causes;
  for instance, in $\mathsf{P_3}$, the configuration $\{c\}$ does not include
  $b$ despite $b < c$
\item
  causes can be re-enabled; for example, $b$ is enabled at $\{c\}$ in
  $\mathsf{P_3}$, but the re-execution of $a$ disables $b$ again, i.e., $b$ is
  not enabled at $\{a,c\}$;
\item
  conflicts are not inherited through causality, as in $\mathsf{P_3}$;
\item
  reversibility induces dependencies on concurrent events, for example,
  $c$ in $\mathsf{P_2}$ can be reversed only after the concurrent event $d$
  has been executed.
\end{itemize}

\section{Nets} \label{sec:nets}
\subsection{Nets with inhibitor arcs}

We summarize the fundamentals of Petri nets with inhibitor arcs following the
presentation in~\cite{MR:CN,BBCP:rivista}.

\begin{definition}\label{def:ptnet}
  A \emph{Petri net} is a 4-tuple $N = \langle S, T, F, \mathsf{m}\rangle$
  where $S$ is a set of \emph{places}, $T$ is a set of \emph{transitions} with
  $S \cap T = \emptyset$, $F \subseteq (S\times T)\cup (T\times S)$ is the
  \emph{flow} relation, and $\mathsf{m}\in \mu S$ is the \emph{initial
    marking}.
\end{definition}

\begin{definition}\label{de:contextualnet}
  A \emph{Petri net with inhibitor arcs} (\inet for short) is a tuple
  $N = \langle S, T, F, I, \mathsf{m}\rangle$, where
  $\langle S, T, F, \mathsf{m}\rangle$ is a Petri net, and
  $I \subseteq S\times T$ is the \emph{inhibiting} relation.
\end{definition}

Given an \inet $N = \langle S, T, F, I, \mathsf{m}\rangle$ and $x\in S\cup T$,
the {\em pre-} and {\em postset} of $x$ are respectively defined as the
(multi)sets $\pre{x} = \setcomp{y}{(y,x)\in F}$ and
$\post{x} = \setcomp{y}{(x,y)\in F}$.
If $x\in S$, then $\pre{x} \in \mu T$ and $\post{x} \in \mu T$; analogously,
if $x\in T$, then $\pre{x}\in\mu S$ and $\post{x} \in \mu S$.
The {\em inhibitor set} of a transition $t$ is the (multi)set
$\inib{t} = \setcomp{s}{(s,t)\in I}$.
The definitions of $\pre{\cdot}$, $\post{\cdot}$, $\inib{\cdot}$ generalise
straightforwardly to multisets of transitions.

\begin{example}\label{ex:cont-net1}
  Consider the simple \inet $N_1$ depicted in \Cref{ex:introa}, which consists
  of six places depicted as circles and three transitions drawn as boxes. The
  flow relation is represented by black arrows, while the inhibitor relation
  is shown by red lines ended with a small circle.
  The initial marking $\mathsf{m} = \{s_1, s_2, s_3\}$ is represented by the
  bullets drawn within the corresponding places. Let us consider the
  transition $b$, which consumes tokens from $s_2$ and $s_3$, produces a token
  in $s_5$, and is inhibited by $s_1$. Hence, its pre-, post- and inhibiting
  sets are respectively $\pre{b} = \{s_2,s_3\}$, $\post{b} = \{s_5\}$, and
  $\inib{b} = \{s_1\}$.
\end{example}

A (multiset of) transition(s) $A\in \mu T$ is {\em enabled at a marking}
$m\in \mu S$, written $m\trans{A}$, whenever ${\pre{A}} \subseteq m$ and
$\forall s\in\flt{\inib{A}}.\ m(s) = 0\ \land\ \post{A}(s) = 0$.
The last condition requires the absence of tokens in all places connected via
inhibitor arcs to the transitions in $\flt{A}$. Observe that the multiset
$\zero$ is enabled at every marking.
A (multiset of) transition(s) $A$ enabled at a marking $m$ can \emph{fire} and
its firing produces the marking $m' = m - \pre{A} + \post{A}$. The firing of
$A$ at a marking $m$ is denoted by $m\trans{A}m'$.
We assume that each transition $t$ of an \inet $N$ is defined such that
$\pre{t}\neq\emptyset$, i.e., it cannot fire \emph{spontaneously} in an
uncontrolled manner without consuming tokens.

\begin{example}
  Consider the \inet introduced in \Cref{ex:cont-net1} and note that both $a$
  and $c$ are enabled at the initial marking $\mathsf{m}$. On the contrary,
  $b$ is not enabled because its inhibitor place $s_1$ contains a token.
  The firing of $a$ produces the marking $m' = \setenum{s_2, s_3, s_4}$,
  denoted as $\mathsf{m}\trans{a} m'$. In this marking, $b$ becomes enabled
  since its preset $s_2$ and $s_3$ are marked, while its inhibitor place $s_1$
  is not.
\end{example}

We now give a precise definition of firing sequences, reachable markings and
states of a \inet in terms of sequences of firings. Given an \inet
$N = \langle S, T, F, I, \mathsf{m}\rangle$, a \emph{firing sequence}
(shortened as \fs) of $N$ is a sequence $\sigma$ of markings such that for
each $0 \leq i < \len{\sigma}$ there is a multiset of transitions $A_i$
enabled at $\sigma(i)$ and 
$\sigma(i)\trans{A_i}\sigma(i+1)$.
A firing sequence $\sigma$ is written as $m_0\trans{A_0}m_1$ $\cdots$
$m_{n}\trans{A_n} m_{n+1}$ $\cdots$ and we use $\start{\sigma}$ to indicate
its initial marking $\sigma(0) = m_0$; should $\sigma$ be finite (i.e.,
$\len{\sigma} < \infty$), $\lead{\sigma}$ designates its final marking
$\sigma(\len{\sigma})$.
We say that $\sigma$ starts at a marking $m$ if $\start{\sigma} = m$, and let
$\firseq{N}{m}$ denote the set of all firing sequences of the \inet $N$ that
start at $m$.
A marking $m$ is \emph{reachable} in $N$ iff there exists an \fs\
$\sigma \in \firseq{N}{\mathsf{m}}$ such that $m = \sigma(i)$ for some
$i\leq \len{\sigma}$.
The set of all reachable markings of $N$ is
$\reachMark{N} = \bigcup_{\sigma\in\firseq{N}{\mathsf{m}}}
\setcomp{\sigma(i)}{i\leq \len{\sigma}}$.
Given an \fs\ $\sigma$, $\mathcal{X}_{\sigma}$ is the set of all sequences of
multisets of transitions that \emph{agree} with $\sigma$, namely the set
$\setcomp{\theta}{\len{\theta} = \len{\sigma}\ \land\
  (\sigma(i)\trans{\theta(i)}\sigma(i+1)\ \mathit{with}\ i<\len{\sigma})}$,
and
$X_{\sigma} = \setcomp{\sum_{i=1}^{\len{\theta}} \theta(i)}{\theta\in
  \mathcal{X}_{\sigma}}$ for the set of multisets of transitions associated to
an \fs.
Each multiset of $X_{\sigma}$ is a \emph{state} of the net and write
\( \states{N} = \bigcup_{\sigma\in\firseq{N}{\mathsf{m}}} X_{\sigma} \) for
the set of states of the net $N$, and $\theta\in \mathcal{X}_{\sigma}$ is an
\emph{execution} of the net.

\begin{definition}\label{de:net-equiv}
  We say that $N_1$ and $N_2$ are {\em equivalent}, written $N_1 \equiv N_2$,
  if they have the same set of states, i.e., $\states{N_1} = \states{N_2}$.
\end{definition}

\begin{example}
  The set of reachable markings of the \inet in \Cref{ex:cont-net1} is
  \(\reachMark{N_1} = \{ \mathsf{m}, \{s_2, s_3, s_4\}, \{s_1, s_2, s_6\},
  \{s_4, s_5\},$
  $\{s_2, s_4, s_6\} \} \) while its set of states is
  \( \states{N_1} = \{ \emptyset, \{a\}, \{c\}, \{a,b\}, \{a,c\} \} \).
\end{example}

An \inet $N$ is {\em safe} if every reachable marking is a set, i.e.,
$\forall m\in \reachMark{N}.m = \flt{m}$.
Henceforth, our focus will be exclusively safe \inet.

For a given net $N = \langle S, T, F, \mathsf{m}\rangle$, we denote the
transitive closure of $F$ as $<_N$. The net $N$ is \emph{acyclic} if the
relation $\leq_N$ is a partial order.

\section{Causal nets}\label{ssec:causal-nets}

We now restrict our attention to representing causal dependencies through
inhibitor arcs. We introduce a class of contextual Petri nets, dubbed {\em
  causal nets}, where causality is determined by the inhibiting relation
rather than the typical flow relation.
The main result in this section (\Cref{th:pesandcn}) states that causal nets
are an adequate operational counterpart of \pes{}es. As a matter of fact, they
are tightly connected with occurrence nets (as discussed in
\Cref{sec:discussion}).

We define the relation $\lessdot$ between transitions $t, t'$ of an \inet. We
stipulate $t \lessdot t'$ iff $\pre{t}\cap\inib{t'}\neq\emptyset$, i.e., the
firing of $t$ consumes (at least) one of the tokens that inhibit the firing of
$t'$. Similarly, we define $\cnconf$ by $t \ \cnconf\ t'$ iff
$\pre{t}\cap\pre{t'}\neq\emptyset$.

\begin{definition}\label{de:pre-causal-net}
  Let $C = \langle S, T, F, I, \mathsf{m}\rangle$ be an \inet. $C$ is a
  \emph{pre-causal} net (p\ca) if the following conditions are satisfied:
 \begin{enumerate}
 \item ${<_C} \cap {(T\times T)} = \emptyset$;
 \item $\flt{\post T} = \post T$;
 \item
   $\forall t\neq t'\in T.\ {\pre{t}} \cap{\pre{t'}} \cap{\Inib{T}{}} =
   \emptyset$,
 \item $\forall t\in T$. $\inib{t}$ is finite;
 \item $\lessdot$ is an irreflexive partial order;
 \item $\forall t', t''\in \histtwo{t}{\lessdot}.\ t' \ \cnconf\
   t'' 
   \Rightarrow\ t' = t''$; and
 \item $\mathsf{m} = \pre{T}$ and $\Inib{T}{}\subseteq \mathsf{m}$.
 \end{enumerate}

 We say a p\ca\ $C$ is a {\em causal} net (\ca) if it also satisfies the
 following condition
 \[\forall t, t', t''\in T.\ t\ \cnconf\  t'\ \land\
   t'\lessdot t''\ \Rightarrow\ t \ \cnconf\  t''
 \]
\end{definition}

The conditions imposed on \ca{}s share motivations with those posed on
occurrence nets, unravel nets \cite{Pinna:PN11,CaPi:PN14, CP:soap17} and flow
nets~\cite{Bou:FESFN}, which are aimed at explaining computations without
resorting to firing sequences.
The first condition, alternatively expressed as
$\forall t\in T.\ \forall s\in\pre{t}.\ \pre{s} = \emptyset$, signifies that
causal dependencies within \ca{}s do not originate from the flow relation.
This is evident as $\post{t}\cap\pre{t}' = \emptyset$ holds for all
$ t, t'\in T$.
The second condition guarantees the absence of backward conflicts, meaning
that a place is, at most, part of the postset of a single transition.
The third condition asserts that any place present in the preset of at least
two transitions cannot be connected via inhibitor arcs to other places in the
net. This restriction prohibits or-causality, where the firing of a transition
could have multiple distinct causes.
Together with the fourth requirement, these conditions ensure that each
transition has a finite set of causes, specifically, the set
$\setcomp{t'\in T}{t'\lessdot t}$ is finite.
The fifth condition prevents the occurrence of cycles in the dependencies
arising from inhibitor arcs. Given that $\lessdot$ is irreflexive, it ensures
that none of the transitions is blocked, i.e.,
${\pre{t}} \cap{\inib{t}}=\emptyset$ for all $t$.
The sixth condition bears resemblance to the prerequisite of p\pes, ensuring
local inheritance of conflicts. In other words, any pair of different
transitions $t', t''$ in $\histtwo{t}{\lessdot}$ are not in \emph{conflict}.
The last requirement states that the preset of all transitions are initially
marked ($\mathsf{m} = \pre{T}$). Additionally,
$\Inib{T}{}\subseteq \mathsf{m}$, combined with the first requirement, implies
that inhibitor places cannot be part of the postset of any transition.

The supplementary condition for \ca{}s enforces the inheritance of conflicts
along the relation $\lessdot$, where conflicts are represented by shared input
places.

\begin{example}\label{ex:preca1}
  The \inet $C_1$, illustrated in \Cref{ex:a-simple-precab}, is a p\ca net, as
  detailed below. Conditions 1 and 2 of \Cref{de:pre-causal-net} are satisfied
  because each place in the postset of a transition has precisely one incoming
  and no outgoing arcs.
  The only place belonging to the preset of at least two transitions is $s_2$,
  yet there is no inhibitor arc connected to
  $s_2$. Therefore, Condition 3 is also met. Condition 4 is immediately
  satisfied since there is only one inhibitor arc.
  The induced causality relation is the irreflexive partial order
  $\lessdot = \{(b, c)\}$, satisfying Condition 5. If $t\in T$ and
  $t\neq c$, then $\histtwo{t}{\lessdot} = \{t\}$. For $t = c$, we have
  $\histtwo{c}{\lessdot} = \{b, c\}$, and $\pre b$ and $\pre c$ are disjoint,
  fulfilling Condition $6$.
  Condition $7$ is immediate.
  We remark that $C_1$ is not a \ca{} because conflicts are not inherited
  along $\lessdot$. Specifically, $a \ \cnconf\  b$ and $b \lessdot c$, yet
  $a \ \cnconf\  c$. In contrast, the \inet $C_2$ in \Cref{ex:a-simple-ca}, which
  explicitly captures the conflict between $a$ and $c$, is a \ca.

\begin{figure}[t]
    \begin{subfigure}{.45\textwidth}
      \begin{center}
        \scalebox{0.75}{\begin{tikzpicture}[scale=0.8]
\tikzstyle{inhibitorred}=[o-, draw=red,thick]
\tikzstyle{inhibitorblu}=[o-, draw=blue,thick]
\tikzstyle{pre}=[<-,thick]
\tikzstyle{post}=[->,thick]
\tikzstyle{readblue}=[-, draw=blue,thick]
\tikzstyle{transition}=[rectangle, draw=black,thick,minimum size=5mm]
\tikzstyle{place}=[circle, draw=black,thick,minimum size=5mm]
\node[place,tokens=1] (p1) at (0,2.5) [label=left:$s_1$] {};
\node[place,tokens=1] (p2) at (2,2.5) [label=left:$s_2$] {};
\node[place,tokens=1] (p3) at (4,2.5) [label=left:$s_3$] {};
\node[place,tokens=1] (p4) at (6,2.5) [label=left:$s_4$] {};
\node[place,tokens=1] (p5) at (8,2.5) [label=left:$s_5$] {};
\node[](pg) at (0,3.5) {};
\node[place] (p6) at (0,0) [label=left:$s_6$] {};
\node[place] (p7) at (4,0) [label=left:$s_7$] {};
\node[place] (p8) at (6,0) [label=left:$s_8$] {};
\node[place] (p9) at (8,0) [label=left:$s_9$] {};
\node[transition] (t1) at (0,1.25)  {$a$}
edge[pre] (p1)
edge[pre] (p2)
edge[post] (p6);
\node[transition] (t2) at (4,1.25)  {$b$}
edge[pre] (p2)
edge[pre] (p3)
edge[post] (p7);
\node[transition] (t3) at (6,1.25)  {$c$}
edge[pre] (p4)
edge[inhibitorred] (p3)
edge[post] (p8);
\node[transition] (t4) at (8,1.25)  {$d$}
edge[pre] (p5)
edge[post] (p9);
\end{tikzpicture}}
      \end{center}
      \caption{$C_1$\label{ex:a-simple-precab}}
    \end{subfigure}
    \begin{subfigure}{.45\textwidth}
      \begin{center}
        \scalebox{0.75}{\begin{tikzpicture}[scale=0.8]
\tikzstyle{inhibitorred}=[o-, draw=red,thick]
\tikzstyle{inhibitorblu}=[o-, draw=blue,thick]
\tikzstyle{pre}=[<-,thick]
\tikzstyle{post}=[->,thick]
\tikzstyle{readblue}=[-, draw=blue,thick]
\tikzstyle{transition}=[rectangle, draw=black,thick,minimum size=5mm]
\tikzstyle{place}=[circle, draw=black,thick,minimum size=5mm]
\node at (3,4) {~};
\node[place,tokens=1] (p0) at (3,3.5) [label=above:{$s_0$}] {};
\node[place,tokens=1] (p1) at (0,2.5) [label=left:{$s_1$}] {};
\node[place,tokens=1] (p2) at (2,2.5) [label=below:{$s_2$}] {};
\node[place,tokens=1] (p3) at (4,2.5) [label=left:{$s_3$}] {};
\node[place,tokens=1] (p4) at (6,2.5) [label=above:{$s_4$}] {};
\node[place,tokens=1] (p5) at (8,2.5) [label=left:{$s_5$}] {};
\node[place] (p6) at (0,0) [label=left:{$s_6$}] {};
\node[place] (p7) at (4,0) [label=left:{$s_7$}] {};
\node[place] (p8) at (6,0) [label=left:{$s_8$}] {};
\node[place] (p9) at (8,0) [label=left:{$s_9$}] {};
\node[transition] (t1) at (0,1.25)  {$a$}
edge[pre, bend left] (p0)
edge[pre] (p1)
edge[pre] (p2)
edge[post] (p6);
\node[transition] (t2) at (4,1.25)  {$b$}
edge[pre] (p2)
edge[pre] (p3)
edge[post] (p7);
\node[transition] (t3) at (6,1.25)  {$c$}
edge[pre, bend right] (p0)
edge[pre] (p4)
edge[inhibitorred] (p3)
edge[post] (p8);
\node[transition] (t4) at (8,1.25)  {$d$}
edge[pre] (p5)
edge[post] (p9);
\end{tikzpicture}}
      \end{center}
      \caption{$C_2$\label{ex:a-simple-ca}}
    \end{subfigure}
    \caption{Two pre-causal nets}
     \Description{Two pre-causal nets}
    \label{ex:simple-pcas}
  \end{figure}
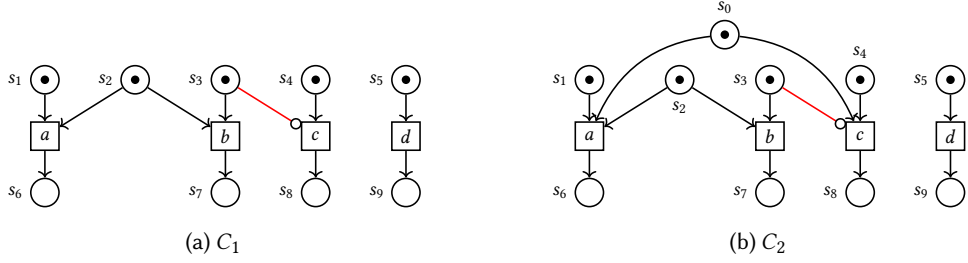
\end{example}

The initial condition outlined in \Cref{de:pre-causal-net} asserts that
$\post{t} \cap \pre{'t} = \emptyset$ for any pair of distinct transitions $t$
and $t'$. As a consequence, each transition is restricted to being executed at
most once. This limitation arises from the fact that its preset cannot be
marked again after firing, as formally expressed below.

\begin{proposition}\label{pr:cn-singleexecution}
 Let $C = \langle S, T, F, I, \mathsf{m}\rangle$ be a \ca.
 For every $t\in T$ and every $X\in \states{C}$, it holds that $X(t) \leq 1$.
\end{proposition}

\begin{proof}
  Since $\pre{T} = \mathsf{m}$ and $\post{t} \cap \pre{'t} = \emptyset$ for
  any pair of transitions $t, t' \in T$, it is evident that each transition
  can be fired at most once.
\end{proof}

\begin{definition}\label{de:pca-conf}
  Let $C = \langle S, T, F, I, \mathsf{m}\rangle$ be a p\ca. A set of
  transitions $X\subseteq T$ is a \emph{configuration} of $C$ if:
  \begin{enumerate}
  \item $\forall t, t'\in X. t\ \cnconf\ t' \Rightarrow\ t = t'$
    (\emph{conflict freeness}), and
  \item $\forall t\in X$. $\histtwo{t}{\lessdot}\subseteq X$ (\emph{left
      closedness} with respect to $\lessdot$).
  \end{enumerate}

  The set of configurations of a p\ca\ $C$ is denoted by $\Conf{C}{p\ca}$. If
  $C$ is a \ca, we use $\Conf{C}{\ca}$ in lieu of $\Conf{C}{p\ca}$.
\end{definition}

\begin{example} \label{ex:conf-ca-c1b} Consider the p\ca\ $C_1$ and the \ca\
  $C_2$ in \Cref{ex:simple-pcas}. Their configurations are identical, as
  illustrated below
  \[
    \Conf{C_1}{p\ca} = \Conf{C_2}{\ca} = \{ \{a\}, \{b\}, \{d\}, \{a, d\},
    \{b, c\}, \{b, d\}, \{b, c, d\}\}
  \]
\end{example}

\begin{example}
  Consider the \ca\ $C_2$ in \Cref{ex:a-simple-ca}. The marking
  $m' = \{s_1,s_7,s_8,s_9\}$ is reachable due to the following firing
  sequence:
  \[
    \mathsf{m}
    \ \trans{\{b,d\}} \
    \{s_0,s_1,s_4,s_7,s_9\}
    \ \trans {\{c\}} \ m'
  \]

  Then, it holds that $\pre{m'} = \{b, c, d\}$ is a configuration of $C_2$
  (where $\Conf{C_2}{\ca}$ is defined in \Cref{ex:conf-ca-c1b}). Conversely,
  take $X = \{b,c,d\}\in\Conf{C_2}{\ca}$, and note that
  $\pre X = \{s_0,s_2, s_3, s_4, s_5\}$ and $\post X = \{s_7, s_8, s_9\}$.
  In this case, ${\sf m} - \pre X + \post X = \{s_1,s_4,s_7,s_9\} = m'$ is is
  a reachable marking of $C_2$.
\end{example}

\section{Causal nets and Event structures}
In this section, we establish the connection between causal nets and event
structures.

\subsection{From p\pes{}es to p\ca{}s}\label{ssec:pestocn}
We demonstrate that every p\pes{} can be linked to a p\ca{} with identical
configurations. To begin, we introduce the mapping $\mathcal{A}$ from
p\pes{}es to p\ca.

\begin{definition}\label{def:ppestocn}
  Let $\mathsf{P} = (E, <, \#)$ be a p\pes. The corresponding p\ca{} is
  $\pestocn{\mathsf{P}} = \langle S, E, F, I, \mathsf{m}\rangle$ where
  \begin{enumerate}
  \item
    $S = \setcomp{(\ast, e)}{e\in E} \ \cup\ \setcomp{(e, \ast)}{e\in E}\cup\
    \setcomp{(\setenum{e, e'},\#)}{e\ \#\ e'}$,
  \item
    $F = \setcomp{(s, e)}{s = (\ast, e) \lor\ (s = (W, \#)\ \land\ e\in W)}\cup\
    \setcomp{(e, s)}{s = (e, \ast)}$,
  \item $I = \setcomp{(s, e)}{s = (\ast, e')\ \land\ e' < e}$, and
  \item
    $\mathsf{m} = \setcomp{(\ast, e)}{e\in E} \cup
    \setcomp{(\setenum{e, e'}, \#)}{e\ \#\ e'}$.
  \end{enumerate}
\end{definition}

The construction associates the p\pes{} $\mathsf{P}$ with a p\ca\ that has as
many transitions as there are events in $\mathsf{P}$. Places are identified
with pairs, taking one of the following forms: (i) $(\ast, e)$ for the
precondition of $e$; (ii) $(e, \ast)$ for the postcondition of $e$; and (iii)
$(\setenum{e, e'},\#)$ for the conflict $e\ \#\ e'$.
The flow relation is defined so that each transition $e$ consumes tokens from
$(\ast, e)$ and every $(W, \#)$ where $e\in W$; and only produces a token in
$(e, \ast)$.
In contrast to the classical construction of occurrence nets from
p\pes{}es~\cite{Win:ES}, places do not convey causal dependencies.
Instead, causal dependencies are modeled through inhibitor arcs: if $e$
causally depends on $e'$ (i.e., $e' < e$), then there is an inhibitor arc
between the transition $e$ and the place $(\ast, e')$, which is a place in the
preset of $e'$.
The initial marking $\sf m$ assigns a token to any place appearing in the
preset of a transition.

\begin{example}\label{ex:ppes-to-pcn}
  Let $\mathsf{P} = (E,<,\#)$ be a p\pes\ where $E = \{a,b,c,d\}$,
  ${<} = \{(b,c)\}$ and ${\#} =\{(a,b),(b,a),(a,c),(c,a)\}$.
  The corresponding p\ca{} is illustrated in \Cref{ex:simple-occ-net-b}. Note
  that the transition $c$ is not initially enabled due to the inhibitor arc
  from $(\ast, b)$; however, all other transitions are enabled. Additionally,
  $a$ and $b$ are in conflict since they both consume tokens from
  $(\{a,b\},\#)$.
  \begin{figure}[t]
    \centerline{\scalebox{0.8}{\begin{tikzpicture}[scale=.8]
\tikzstyle{inhibitorred}=[o-, draw=red,thick]
\tikzstyle{inhibitorblu}=[o-, draw=blue,thick]
\tikzstyle{pre}=[<-,thick]
\tikzstyle{post}=[->,thick]
\tikzstyle{readblue}=[-, draw=blue,thick]
\tikzstyle{transition}=[rectangle, draw=black,thick,minimum size=5mm]
\tikzstyle{place}=[circle, draw=black,thick,minimum size=5mm]
\node[place,tokens=1] (p0) at (2.8,3.5) [label=above:{$(\{a,c\},\#)$}] {};
\node[place,tokens=1] (p1) at (0,2.5) [label=above:{$(*,a)$}] {};
\node[place,tokens=1] (p2) at (1.9,2.5) [label=below:{$(\{a,b\},\#)$}] {};
\node[place,tokens=1] (p3) at (3.8,2.5) [label=left:{$(*,b)$}] {};
\node[place,tokens=1] (p4) at (5.7,2.5) [label=above:{$(*,c)$}] {};
\node[place,tokens=1] (p5) at (7.3,2.5) [label=above:{$(*,d)$}] {};
\node[place] (p6) at (0,0) [label=below:{$(a,*)$}] {};
\node[place] (p7) at (3.8,0) [label=below:{$(b,*)$}] {};
\node[place] (p8) at (5.7,0) [label=below:{$(c,*)$}] {};
\node[place] (p9) at (7.3,0) [label=below:{$(d,*)$}] {};
\node[transition] (t1) at (0,1.25)  {$a$}
edge[pre, bend left] (p0)
edge[pre] (p1)
edge[pre] (p2)
edge[post] (p6);
\node[transition] (t2) at (3.8,1.25)  {$b$}
edge[pre] (p2)
edge[pre] (p3)
edge[post] (p7);
\node[transition] (t3) at (5.7,1.25)  {$c$}
edge[pre, bend right] (p0)
edge[pre] (p4)
edge[inhibitorred] (p3)
edge[post] (p8);
\node[transition] (t4) at (7.3,1.25)  {$d$}
edge[pre] (p5)
edge[post] (p9);
\end{tikzpicture}}}
    \caption{$\pestocn{\mathsf{P}}$\label{ex:simple-occ-net-b}}
     \Description{$\pestocn{\mathsf{P}}$\label{ex:simple-occ-net-b}}
  \end{figure}
\end{example}

The adequacy of the mapping $\pestocn{\cdot}$ is formally stated by
demonstrating the equivalence of the respective sets of configurations.

\begin{proposition}\label{pr:ppestopcn}
 Let $\mathsf{P}$ is a p\pes. Then,
 $\pestocn{\mathsf{P}}$
 is a p\ca{} and
 $\Conf{\mathsf{P}}{p\pes} = \Conf{\pestocn{\mathsf{P}}}{p\ca}$.
\end{proposition}

\begin{proof}
  We begin by examining the conditions to ensure that $\pestocn{\mathsf{P}}$
  is a p\ca.
  \begin{enumerate}
  \item According to the definition of $F$, for every event $e$, we have that
    $\post e = \{(e, \ast)\}$. Additionally, $(e, \ast) \not\in\pre{e'}$ for
    all $e'\in E$. Consequently,
    ${<_{\pestocn{\mathsf{P}}}} \cap (E\times E) = \emptyset$.
  \item By the definition of $F$, $\post{E} = \setcomp{(e, \ast)}{e\in E}$.
    Hence, $\flt{\post E} = \post E$.
  \item By the definition of $F$, for every pair of transitions $e$ and $e'$,
    if a place $s$ belongs to $\pre{e}\cap\pre{e'}$, then it has the form
    $(\setenum{e,e'}, \#)$. According to the definition of $I$, there are no
    inhibitor arcs connected to places of the form $(\setenum{e,e'}, \#)$.
    Therefore,
    $\forall e\neq e'\in E.\ \pre{e}\cap\pre{e'} \cap\Inib{E}{} = \emptyset$
    holds.
  \item According to the definition of $I$, $\forall e\in E$.
    $\inib{e} = \setcomp{(s,e)}{s = (\ast,e')\ \land\ e' < e}$. Since
    $\hist{e}_{<}$ is finite, $\inib{e}$ is so.
  \item According to the definitions of $F$ and $I$, we have
    $\pre e\cap \inib {e'} = \setcomp{(s,e)}{s = (\ast,e')\ \land\ e' < e}$.
    Therefore, $e'\lessdot e$ if and only if $e' < e$. Consequently,
    $\lessdot$ is an irreflexive partial order, given that $<$ is also such.
  \item By analogous reasoning to the previous point,
    $\histtwo{e}{\lessdot} = \histtwo{e}{<}$. We argue by contradiction.
    Suppose there exist $e', e'' \in \histtwo{e}{<}$ such that
    $e'\ \cnconf\ e''$ and $e' \neq e''$. From $e'\ \cnconf\ e''$, we deduce
    that $\pre{e'} \cap \pre{e''} \neq \emptyset$. According to the definition
    of $F$, $\pre{e'} \cap \pre{e''} = {(\setenum{e',e''}, \#)}$ and
    $e'\ \#\ e''$. Since $\histtwo{e}{\lessdot} = \histtwo{e}{<}$, it follows
    that $e' < e$ and $e'' < e$. Therefore, two causes of $e$ are in conflict
    in $\mathsf{P}$, contradicting the hypothesis that $mathsf{P}$ is a p\pes.
  \item $\mathsf{m} = \pre{E}$ and $\Inib{E}{}\subseteq \mathsf{m}$ both hold
    by construction.
  \end{enumerate}

  It remains to show that configurations of $\mathsf{P}$ and
  $\pestocn{\mathsf{P}}$ coincide.
  \begin{itemize}
  \item [$\subseteq$)] Consider a configuration $X$ of $\mathsf P$, i.e.,
    $X\in \Conf{\mathsf{P}}{p\pes}$. Assume a total ordering
    $\setenum{e_1, \dots, e_n}$ of the events in $X$ that is compatible with
    the causality relation ($<$). Let $X_i$ represent the set of events
    $\setenum{e_1, \dots, e_i}$ for $i\in \setenum{1, \dots, n}$.
    Through a straightforward induction on $i$, it can be demonstrated that
    the transition $e_i$ is enabled at
    $\mathsf{m} - \pre{X_{i-1}} + \post{X_{i-1}}$.
    Consequently, $\mathsf{m} - \pre{X} + \post{X}$ is a reachable
    marking, and, as a result, $X \in \Conf{\pestocn{\mathsf{P}}}{p\ca}$.

  \item [$\supseteq$)] Consider a configuration $X$ in
    $\Conf{\pestocn{\mathsf{P}}}{p\ca}$. For all $e, e' \in X$, it holds that
    $\pre{e} \cap \pre{e'} = \emptyset$, implying $\neg(e \ \cnconf\ e')$.
    Consequently, $\neg(e \ \#\ e')$. Additionally,
    $\histtwo{e}{\lessdot} = \histtwo{e}{<}$. Hence, $X$ is downward closed with
    respect to $<$. Therefore, $X \in \Conf{\mathsf{P}}{p\pes}$.
    \qedhere
  \end{itemize}
\end{proof}

\begin{corollary}\label{cor:pestocn}
 Let $\mathsf{P}$ be a \pes{}. Then $\pestocn{\mathsf{P}}$
 is a \ca{} and
 $\Conf{\mathsf{P}}{\pes} = \Conf{\pestocn{\mathsf{P}}}{\ca}$.
\end{corollary}

\subsection{From p\ca{}s to p\pes{}es}\label{ssec:cntopes}

Now, we introduce the mapping $\mathcal{Q}$ from p\ca{}s to p\pes{}es. Its
definition relies on the facts that $\lessdot$ is an irreflexive partial
order, and conflicts are locally preserved in a $p\ca$.

\begin{definition}\label{def:cntoppes}
  Let $C = \langle S, T, F, I, \mathsf{m}\rangle$ be a p\ca. The associated
  p\pes{} is $\cntopes{C} = (T, \lessdot, \#)$ where
  \[\#= \setcomp{ (t, t')}{t\neq t'\in T\ \wedge\ t\ \cnconf\ t'}.\]
\end{definition}

\begin{example}
  Consider the p\ca{} $C_1$ of \Cref{ex:preca1}. Note that, $b \lessdot c$
  holds because $\pre{b}\cap \inib{c}\neq\emptyset$. Additionally, $a$ and $b$
  are in conflict (i.e., both $a \# b$ and $b \# a$ holds) because
  $\pre{a}\cap\pre{b}\neq\emptyset$. Hence,

  \[\cntopes{C_1} = (\setenum{a,b,c,d}, \{(b, c)\}, \{(a, b),(b,a)\})\]

  The \pes{} associated with the \ca{} $C_2$ of \Cref{ex:preca1} is

  \[\cntopes{C_2} = (\setenum{a,b,c,d}, \{(b,c)\},
    \{(a,b),(b,a),(a,c),(b,c))\}\]

  which is actually a \pes{}\ because conflicts are hereditary.
\end{example}

\begin{proposition}\label{pr:pcn-to-ppes}
  Let $C$ be a p\ca. Then, $\cntopes{C}$ is a p\pes{} and
  $\Conf{C}{p\ca} = \Conf{\cntopes{C}}{p\pes}$.
\end{proposition}

\begin{proof}
  Consider the p\ca $C = \langle S, T, F, I, \mathsf{m}\rangle$ and its
  encoding $\mathcal{Q}(C) = (T, \lessdot, \#)$.
  We first show that $\mathcal{Q}(C)$ is a p\pes.
  Note that $\#$ is irreflexive by definition. It is also symmetric because
  $\cnconf$ is so. Since $C$ is a p\ca{}, $\lessdot$ is an irreflexive partial
  order. Moreovoer, $\setcomp{e'\in E}{e' \lessdot e}$ is finite and conflict
  free because $\hist{e}_{\lessdot}$ is finite and conflict free.
  Since $C$ is a p\ca, $e \lessdot e'$ implies $\neg (e\ \cnconf\ e')$.
  Therefore, we conclude that $e < e'$ implies $e\ \#\ e'$.
  Hence, $\mathcal{Q}(C)$ is a p\pes.

  We now show that $\Conf{\cntopes{C}}{p\pes} = \Conf{C}{p\ca}$. The proof is
  analogous to the one in \Cref{pr:ppestopcn} but we spell it in some details.
  \begin{itemize}
  \item [$\subseteq$)] Consider a reachable marking $m'$ of $C$. It is easy to
    see that $\pre{m'}$ is a configuration of $\mathcal{Q}(C)$, as the firing
    sequence leading to $m'$ guaranties conflict freeness of $\pre{m'}$ and
    also that for each $e\in \pre{m'}$ it holds that $\histtwo{e}{\lessdot}$
    is a subset of $\pre{m'}$.
  \item[$\supseteq$)] Consider a configuration $X$ of $\cntopes{C}$.
    Assume a total ordering
    $\setenum{e_1, \dots, e_n}$ of the events in $X$ that is compatible with
    the causality relation ($\lessdot$). Let $X_i$ represent the set of events
    $\setenum{e_1, \dots, e_i}$ for $i\in \setenum{1, \dots, n}$.
    Through a straightforward induction on $i$, it can be demonstrated that
    the transition $e_i$ is enabled at
    $X_i$.
    Consequently, $X$ is a reachable
    marking, and, as a result, $X \in \Conf{\cntopes{C}}{p\pes}$.\qedhere
  \end{itemize}
\end{proof}

\begin{corollary}\label{cor:cn-to-pes}
  Let $C$ be a \ca. Then, $\cntopes{C}$ is a \pes{} and
  $\Conf{\cntopes{C}}{\pes} = \Conf{C}{\ca}$.
\end{corollary}

\begin{proof}
  If $C$ is a \ca, then conflicts are also inherited along $\lessdot$. Hence,
  $\mathcal{Q}(C)$ is a \pes.
\end{proof}

The following result ensures that the notion of causal nets is adequate for
\pes{}es. It is obtained by combining \Cref{pr:ppestopcn} and
\Cref{cor:pestocn} (in \Cref{ssec:pestocn}) with \Cref{pr:pcn-to-ppes} and
\Cref{cor:cn-to-pes} above.

\begin{theorem}\label{th:pesandcn}
  If $C$ is a causal net, then $C \equiv \mathcal{A}(\mathcal{Q}(C))$. If
  $\mathsf{P}$ is a \pes, then
  $\mathsf{P} \equiv \mathcal{Q}(\mathcal{A}(\mathsf{P}))$.
\end{theorem}

\section{Reversible {\ca}{}s and reversible {\pes}{}es}\label{sec:rpesandnets}
In this section we introduce a reversible version of p\ca{}s and show that
they are an operational counterpart for r\pes{}es.

\subsection{Reversible causal nets}\label{ssec:rcn}

The intuition behind the definition of a reversible causal net is that of
extending a (pre) causal net with transitions referred to as {\em backward}.
These backward transitions are designated to reverse or undo the effects of
previously executed ordinary transitions, which are called {\em forward}.
Consider an \inet $N = \langle S, T, F, I, \mathsf{m}\rangle$, and let
$t\in T$ be a transition. We use $S_t$ to denote the set of places
$\setcomp{s\in S}{s\in\pre{t}\ \land\ \post{s} = \setenum{t}}$.
In other words, $S_t$ represents the places in the preset of $t$ that do not
appear in the preset of any other transitions.

\begin{definition}\label{de:reversible-causal-net}
  An \inet $V = \langle S, T, F, I, \mathsf{m}\rangle$ is a \emph{reversible
    causal net} (\rcn) if there exists a partition $\{\fwdset, \bwdset\}$ of
  $T$ satisfying the following conditions (here, $\fwdset$ represents the set of forward transitions, and $\bwdset$ comprises the backward transitions):
 \begin{enumerate}
 \item
   $\langle S, \fwdset, F_{|\fwdset\times\fwdset}, I_{|\fwdset\times\fwdset},
   \mathsf{m}\rangle$ is a p\ca\ net;
 \item $\flt{\pre \bwdset}={\pre \bwdset}$ and
   $\forall \abwd\in\bwdset.\ \exists!\ \afwd\in \fwdset$ such that
   $\post{\afwd} = \pre{\abwd}$, $\pre{\afwd} = \post{\abwd}$, and
   $S_{\afwd}\cap\inib{\abwd}\neq \emptyset$;
 \item $\forall \abwd\in\bwdset.$
   $K = \{\afwd\ |\ \afwd \in \fwdset \wedge \inib{\abwd} \cap \pre \fwdset
   \neq \emptyset\}$ is finite and $\flt{\pre K}={\pre K}$;
 \item $\forall \abwd\in\bwdset.$$\forall \afwd\in\fwdset.$ if $\pre\afwd\cap
   \inib{\abwd}{} \neq \emptyset$ then $\post\afwd\cap \inib{\abwd}{} =
   \emptyset$;
 \item ${\lll} \subseteq
   {\fwdset\times\fwdset}$ is a transitive relation defined such that
   $\afwd\lll \afwd'$ if $\afwd \lessdot \afwd'$ and if there exists $\abwd
   \in \bwdset$ such that $\pre\abwd = \post
   \afwd$ then $\inib{\abwd}\cap\post{\afwd'}\neq\emptyset$; and
 \item $\forall \afwd, \afwd', \afwd''\in \fwdset.\
   \pre{\afwd}\cap\pre{\afwd}'\neq\emptyset\ \land\ \afwd'\lll \afwd''\
   \Rightarrow\ \pre{\afwd}\cap\pre{\afwd''}\neq\emptyset$.
 \end{enumerate}

 We write $\arcn V \bwdset $ for an \rcn\
 $V$ with backward transitions $\bwdset$.
\end{definition}

The first condition states that the subnet consisting of just forward
transitions is a pre-causal net.
The second condition establishes that each backward transition $\abwd$
precisely reverses one forward transition $\afwd$. As a result, $\abwd$
consumes the tokens produced by $\afwd$ (i.e., $\post{\afwd} = \pre{\abwd}$)
and produces the tokens consumed by $\afwd$ (i.e.,
$\post{\abwd} = \pre{\afwd}$). The condition
$S_{\afwd}\cap\inib{\abwd}\neq \emptyset$ ensures that $\abwd$ causally
depends on $\afwd$. Requiring $\flt{\pre \bwdset}={\pre \bwdset}$ ensures that
a forward transition has at most one reversing transition.
The remaining conditions mirror those imposed on \pes{}es, where inhibitor arcs
model the reverse causality relation ($\prec$) and the prevention relation
($\triangleright$). If an inhibitor arc connects a backward transition to a
place in the preset of some forward transition, then the modeled relation is
reverse causality. Conversely, prevention is represented by linking a backward
transition to a place in the postset of a forward transition.
Consequently, the third condition requires each backward transition to
causally depend on a finite number of forward transitions (i.e., those in
$K$), which should also be conflict-free (i.e., not sharing places in their
preset).
The fourth condition stipulates that a backward transition $\abwd$ causally
dependent on a forward transition $\afwd$ (i.e.,
$\pre\afwd\cap \inib{\abwd}{} \neq \emptyset$) cannot be prevented by the same
transition (i.e., $\post\afwd\cap \inib{\abwd}{} = \emptyset$).
The relation $\lll$ defined by the fifth condition is analogous to sustained
causation, coinciding with causality only when prevention enforces causal
reversibility. The last condition asserts that conflicts are inherited along
$\lll$.

\begin{example}
  Consider the \inet $V$ in \Cref{ex:simple-rcn-net}, which is an $\rcn$ with
  reversing transitions $\bwdset = \{\un b, \un c\}$.
  Firstly, observe that
  $\langle S, \fwdset, F_{|\fwdset\times\fwdset}, I_{|\fwdset\times\fwdset},
  \mathsf{m}\rangle$ corresponds to the causal net $C_2$ illustrated in
  \Cref{ex:a-simple-ca}.
  There are two backward transitions: $\un b$, for reversing $b$, and $\un c$
  for reversing $c$. Each of them reverses the markings of its associated
  forward transition, e.g., $\un b$ consumes from $s_7$ and produces on $s_2$
  and $s_3$.
  The inhibitor arcs linking $s_3$ to $\un b$ and $s_4$ to $\un c$ encode
  reverse causality, e.g., establishing a causal dependency of $\un b$ on $b$.
  In this case, the behaviour of the net would remain unaffected if these
  inhibitor arcs were omitted.
  However, their inclusion is mandated for a consistent representation of
  causality through $\lessdot$, as opposed to relying on the flow relation.
  Moreover, the inhibitor arc connecting $s_5$ to $\un c$ encodes also reverse
  causality by stipulating that $\un c$ cannot be fired until $d$ is fired and
  the token in $s_5$ is consumed.
  Finally, the inhibitor arc linking $s_8$ to $\un b$ models prevention.
  Specifically, $\un b$ is not enabled if $s_8$ is marked, signifying that $c$
  has been fired and preventing the execution of $\un b$.

  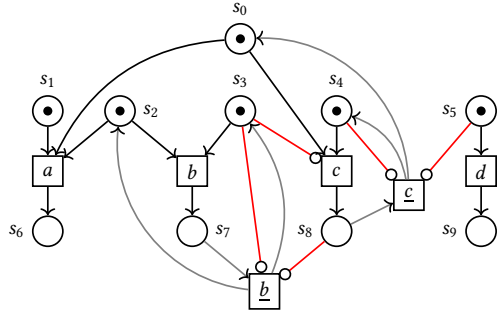
\begin{figure}[t]
    \begin{center}
      \scalebox{0.8}{\begin{tikzpicture}[scale=.8]
\tikzstyle{inhibitorred}=[o-, draw=red,thick]
\tikzstyle{inhibitorblu}=[o-, draw=blue,thick]
\tikzstyle{pre}=[<-,thick]
\tikzstyle{post}=[->,thick]
\tikzstyle{readblue}=[-, draw=blue,thick]
\tikzstyle{transition}=[rectangle, draw=black,thick,minimum size=5mm]
\tikzstyle{place}=[circle, draw=black,thick,minimum size=5mm]
\tikzstyle{prerev}=[<-,thick,draw=gray]
\tikzstyle{postrev}=[->,thick,draw=gray]
\node[place,tokens=1] (p0) at (4,4) [label=above:$s_0$] {};
\node[place,tokens=1] (p10) at (0,2.5) [label=above:$s_1$] {};
\node[place,tokens=1] (p1) at (1.5,2.5) [label=right:$s_2$] {};
\node[place,tokens=1] (p2) at (4,2.5) [label=above:$s_3$] {};
\node[place,tokens=1] (p3) at (6,2.5) [label=above:$s_4$] {};
\node[place,tokens=1] (p4) at (9,2.5) [label=left:$s_5$] {};
\node[place] (p5) at (0,0) [label=left:$s_6$] {};
\node[place] (p6) at (3,0) [label=right:$s_7$] {};
\node[place] (p7) at (6,0) [label=left:$s_8$] {};
\node[place] (p8) at (9,0) [label=left:$s_9$] {};

\node[transition] (t1) at (0,1.25)  {$a$}
edge[pre, bend left] (p0)
edge[pre] (p10)
edge[pre] (p1)
edge[post](p5)
;
\node[transition] (t2) at (3,1.25)  {$b$}
edge[pre] (p1)
edge[pre] (p2)
edge[post](p6)
;

\node[transition] (t5) at (4.5,-1.25)  {$\underline b$}

edge[prerev](p6)
edge[postrev,bend right](p2)
edge[inhibitorred] (p2)
edge[inhibitorred](p7)
edge[postrev, bend left=45](p1)
;

\node[transition] (t3) at (6,1.25)  {$c$}
edge[pre] (p0)

edge[inhibitorred](p2)
edge[pre](p3)
edge[post](p7)

;

\node[transition] (t5) at (7.5,0.75)  {$\underline c$}
edge[prerev](p7)
edge[inhibitorred](p3)
edge[inhibitorred](p4)
edge[postrev, bend right](p3)
edge[postrev, bend right=45](p0)

;

\node[transition] (t4) at (9,1.25)  {$d$}
edge[pre](p4)
edge[post](p8)

;

\end{tikzpicture}}
      \caption{A simple r\ca}
       \Description{A simple r\ca}
      \label{ex:simple-rcn-net}
    \end{center}
  \end{figure}

\end{example}


\subsection{From \rpes{}es to \rcn{}es}\label{ssec:rpesandrcn}
This section delves into the relation between \rpes{}es and \rcn{}es. We
establish the adequacy of the identified connection in \Cref{ssec:rcnandrpes}.

\begin{definition}\label{def:rcntorpes-a}
  Let $\mathsf{P} = (E, \anR, <, \#, \prec, \triangleright)$ be an \rpes and
  $\pestocn{E,<,\#} = \langle S, T', F', I', \mathsf{m}\rangle$ the p\ca\
  associated with the forward flow. Then, the corresponding \rcn\ is
  $\rpestorcn{\mathsf{P}} = \langle S, T, F, I, \mathsf{m}\rangle$ where
  \begin{enumerate}
  \item $T = T'\cup \setcomp{(\anr, \re)}{\anr\in\anR}$;
  \item
    $F = F' \cup \setcomp{(s, (\anr, \re))}{s\in\post{\anr}} \cup
    \setcomp{((\anr, \re), s)}{s\in\pre{\anr}}$;
  \item
    $I = I' \cup \setcomp{((*, e), (\anr, \re))}{e \prec \underline{\anr}}
    \cup\ \setcomp{((e, *), (\anr, \re))}{e \triangleright \underline{\anr}}$.
  \end{enumerate}
\end{definition}

The construction employs the mapping $\pestocn{\_}$ to derive a p\ca{} from
the underlying p\pes{} that consists solely of forward events. This resulting
structure is then extended with an equal number of backward transitions as
there are reversible events in $\mathsf{P}$. A backward transition
$(\anr,\re)$ reversing $\anr$, is defined with its preset being the postset of
$\anr$ and its poset being the preset of $\anr$. Inhibitor arcs are
implemented as anticipated: the reverse causality relation $e \prec \un \anr$
translates into an arc connecting the backward transition $(\anr ,\re)$ to the
enabling condition of $e$, i.e., $(*, e)$; the prevention relation
$e \triangleright \un \anr$ is represented by an arc connecting $(\anr ,\re)$
to the post condition of $e$, i.e., $(e, *)$.

\begin{example}
  Consider the \rpes $\mathsf{P_1} = (E,\anR,<,\#,\prec,\pr)$ presented in
  \Cref{ex:rpes-causal}. Notably, the structure $(E,<,\#)$ and its associated
  \ca{} were discussed in \Cref{ex:ppes-to-pcn}. Then, the \rcn\
  $\rpestorcn{\mathsf{P_1}}$ illustrated in \Cref{ex:simple-rcn-net} is
  obtained by extending the \ca{} in \Cref{ex:simple-occ-net-b} with the
  transitions $(b,\re)$ and $(c,\re)$ corresponding to the reversing events
  $\un b$ and $\un c$. The inhibitor arcs from $(b,*)$ to $(b,\re)$ and
  $(c,*)$ to $(c,\re)$ indicate that the backward transitions are only enabled
  after the respective forward transitions have been executed. Furthermore,
  the inhibitor arc from $(*,c)$ to $(b,\re)$ indicates that the firing of $b$
  is prohibited if $c$ has already been executed.

  \begin{figure}[t]
    \begin{center}
      \scalebox{0.8}{\begin{tikzpicture}[scale=.8]
\tikzstyle{inhibitorred}=[o-, draw=red,thick]
\tikzstyle{inhibitorblu}=[o-, draw=blue,thick]
\tikzstyle{pre}=[<-,thick]
\tikzstyle{post}=[->,thick]
\tikzstyle{readblue}=[-, draw=blue,thick]
\tikzstyle{rev}=[-, draw=gray,thick]
\tikzstyle{prerev}=[<-,thick,draw=gray]
\tikzstyle{postrev}=[->,thick,draw=gray]

\tikzstyle{transition}=[rectangle, draw=black,thick,minimum size=5mm]
\tikzstyle{place}=[circle, draw=black,thick,minimum size=5mm]
\node[place,tokens=1] (p0) at (4,4) [label=above:${(\{a,c\},\#)}$] {};
\node[place,tokens=1] (p10) at (-1,2.5) [label=above:${(*,a)}$] {};
\node[place,tokens=1] (p1) at (1.5,2.5) [label=above:${(\{a,b\},\#)}$] {};
\node[place,tokens=1] (p2) at (3.5,2.5) [label=above:${(*,b)}$] {};
\node[place,tokens=1] (p3) at (6,2.5) [label=above:${(*,c)}$] {};
\node[place,tokens=1] (p4) at (8.5,2.5) [label=above:${(*,d)}$] {};
\node[place] (p5) at (-1,0) [label=below:${(a,*)}$] {};
\node[place] (p6) at (3,0) [label=right:${(b,*)}$] {};
\node[place] (p7) at (6,0) [label=below:${(c,*)}$] {};
\node[place] (p8) at (8.5,0) [label=below:${(d,*)}$] {};

\node[transition] (t1) at (-1,1.25)  {$a$}
edge[pre, bend left=45] (p0)
edge[pre] (p10)
edge[pre] (p1)
edge[post](p5)
;
\node[transition] (t2) at (3,1.25)  {$b$}
edge[pre] (p1)
edge[pre] (p2)
edge[post](p6)
;

\node[rev] (t5) at (5,-1.25)  {$(b,\re)$}

edge[prerev](p6)
edge[postrev,bend right](p2)
edge[inhibitorred] (p2)
edge[inhibitorred](p7)
edge[postrev, bend left=45](p1)
;

\node[transition] (t3) at (6,1.25)  {$c$}
edge[pre] (p0)

edge[inhibitorred](p2)
edge[pre](p3)
edge[post](p7)

;

\node[rev] (t5) at (7.3,0.75)  {$(c,\re)$}
edge[prerev](p7)
edge[inhibitorred](p3)
edge[postrev, bend right](p3)
edge[postrev, bend right = 60](p0)

;

\node[transition] (t4) at (8.5,1.25)  {$d$}
edge[pre](p4)
edge[post](p8)

;

\end{tikzpicture}}
    \end{center}
    \caption{$\rpestorcn{\mathsf{P_1}}$\label{ex:rpes2ca1}}
     \Description{$\rpestorcn{\mathsf{P_1}}$\label{ex:rpes2ca1}}
  \end{figure}

  The \rcn{} $\pestocn{\mathsf{P_3}}$ corresponding to the \rpes{}
  $\mathsf{P_3}$ in \Cref{ex:rpes-out-of-order} is depicted in
  \Cref{ex:rpes2ca2}. Note the absence of the place $(\{a,c\},\#)$, reflecting
  that $a$ and $c$ do not conflict in $\mathsf{P_3}$.
  Furthermore, there is not an inhibitor arc from $(*,c)$ to $(b,\re)$,
  indicating that $b$ can be reversed even when $(*,c)$ is marked. Moreover,
  the inhibitor arc from $(d,*)$ to $(c,\re)$ serves to prevent the reversal
  of $c$ until $d$ has been fired.

  \begin{figure}[t]
    \begin{center}
      \scalebox{0.8}{\begin{tikzpicture}[scale=.8]
\tikzstyle{inhibitorred}=[o-, draw=red,thick]
\tikzstyle{inhibitorblu}=[o-, draw=blue,thick]
\tikzstyle{pre}=[<-,thick]
\tikzstyle{post}=[->,thick]
\tikzstyle{readblue}=[-, draw=blue,thick]
\tikzstyle{rev}=[-, draw=gray,thick]
\tikzstyle{prerev}=[<-,thick,draw=gray]
\tikzstyle{postrev}=[->,thick,draw=gray]

\tikzstyle{transition}=[rectangle, draw=black,thick,minimum size=5mm]
\tikzstyle{place}=[circle, draw=black,thick,minimum size=5mm]
\node[place,tokens=1] (p10) at (0,2.5) [label=above:${(*,a)}$] {};
\node[place,tokens=1] (p1) at (1.5,2.5) [label=above:${(\{a,b\},\#)}$] {};
\node[place,tokens=1] (p2) at (3.5,2.5) [label=above:${(*,b)}$] {};
\node[place,tokens=1] (p3) at (6,2.5) [label=above:${(*,c)}$] {};
\node[place,tokens=1] (p4) at (8.5,2.5) [label=above:${(*,d)}$] {};
\node[place] (p5) at (0,0) [label=below:${(a,*)}$] {};
\node[place] (p6) at (3,0) [label=right:${(b,*)}$] {};
\node[place] (p7) at (6,0) [label=below:${(c,*)}$] {};
\node[place] (p8) at (8.5,0) [label=below:${(d,*)}$] {};

\node[transition] (t1) at (0,1.25)  {$a$}
edge[pre] (p10)
edge[pre] (p1)
edge[post](p5)
;
\node[transition] (t2) at (3,1.25)  {$b$}
edge[pre] (p1)
edge[pre] (p2)
edge[post](p6)
;

\node[rev] (t5) at (5,-1.25)  {$(b,\re)$}

edge[prerev](p6)
edge[postrev,bend right](p2)
edge[inhibitorred] (p2)
edge[postrev, bend left=45](p1)
;

\node[transition] (t3) at (6,1.25)  {$c$}

edge[inhibitorred](p2)
edge[pre](p3)
edge[post](p7)

;

\node[rev] (t5) at (7.3,0.75)  {$(c,\re)$}
edge[prerev](p7)
edge[inhibitorred](p3)
edge[inhibitorred](p4)
edge[postrev, bend right](p3)

;

\node[transition] (t4) at (8.5,1.25)  {$d$}
edge[pre](p4)
edge[post](p8)

;

\end{tikzpicture}}
    \end{center}
    \caption{$\rpestorcn{\mathsf{P_3}}$\label{ex:rpes2ca2}}
     \Description{$\rpestorcn{\mathsf{P_3}}$\label{ex:rpes2ca2}}
  \end{figure}

  The encoding of the \rpes{} $\mathsf{P_2}$ in \Cref{ex:rpes-not-causal}
  results in a net that is isomorphic to the one shown in
  \Cref{ex:simple-rcn-net}, with only the names of places and transitions
  being different.
\end{example}

\begin{proposition}\label{pr:rpes-to-rca}
  If $\mathsf{P}$ is an \rpes{}, then $\rpestorcn{\mathsf{P}}$ is an $\rcn$.
\end{proposition}

\begin{proof}

  Assume $\mathsf{P} = (E, \anR, <, \#, \prec, \triangleright)$. We show that
  $\rpestorcn{\mathsf{P}} = \langle S, T, F, I, \mathsf{m}\rangle$ with
  $\bwdset = \setcomp{(\anr,\re)}{\anr\in\anR}$ satisfies the conditions in
  \Cref{de:reversible-causal-net}.

  \begin{enumerate}
  \item $\pestocn{E,<,\#} = \langle S, T', F', I', \mathsf{m}\rangle$ is a
    p\ca\ by \Cref{pr:ppestopcn}. Moreover, $T' = E$ by \Cref{def:ppestocn}.
    Hence, $\fwdset = E$.
  \item It is immediate that, for any $(\anr,\re)$ there exists a unique
    $u\in T$ such that $\post{\anr} = \pre{(\anr,\re)}$ and
    $\pre{\anr} = \post{(\anr,\re)}$ based on the definition of $F$. Moreover,
    the condition $S_{\anr}\cap\inib{(\anr,\re)}\neq \emptyset$ is satisfied
    by the definition of $F$ and $I$.
  \item By construction,
    $\setcomp{e}{e \in E\ \wedge\ \inib{(\anr,\re)} \cap \pre e \neq
      \emptyset} = \setcomp{e}{ ((*,e), (\anr,\re))\in I \wedge\ e\prec
      \underline{\anr}}$ is finite. This is due to the fact that $\mathsf{P}$
    is an \rpes{}, and thus $\setcomp{e\in E}{e\prec\underline{\anr}}$ is
    finite.
  \item If $\pre e\cap \inib{(\anr,\re)}{} \neq \emptyset$ then
    $((e,*), (\anr,\re)) \in I$. Hence, $e \prec \underline{\anr}$ by the
    definition of $\rpestorcn{\_}$. Since $\mathsf{P}$ is an \rpes{}, it
    follows that $\neg(e \triangleright \un\anr)$. Therefore,
    $((*,e), (\anr,\re)) \not\in I$ and
    $\post e\cap \inib{(\anr,\re)}{} = \emptyset$ by the definition of
    $\rpestorcn{\_}$.
  \item Take $\lll = \ll$, i.e., $e\lll e'$ iff $e\ll e'$. Given that
    $\mathsf{P}$\ is an \rpes, $e\ll e'$ holds if $e < e'$ and if $e\in\anR$,
    then $e' \triangleright \underline{e}$. Therefore, $e < e'$ implies
    $\pre{e}\cap\inib{e'}\neq\emptyset$, and hence $e \lessdot e'$ by
    \Cref{def:ppestocn}. Moreover, $e\in\anR$ and
    $e' \triangleright \underline{e}$, implies $(e,\re)\in \bwdset$ and
    $((e',*), (e,\re))\in I$ (by \Cref{def:rcntorpes-a}). Hence,
    $\inib{(e,\re)} \cap \post{e'}\neq \emptyset$.
  \item If $\pre{e}\cap\pre{e}'\neq\emptyset$, then $e\# e'$ by
    \Cref{def:ppestocn}. If $e'\lll e''$, then $e'\ll e''$ because $\lll$ and
    $\ll$ coincide. Since $\mathsf{P}$ is an \rpes, $e\# e'\ll e''$ implies
    $e\# e''$. Hence, $\pre{e}\cap\pre{e''}\neq\emptyset$ by
    \Cref{def:ppestocn}. \qedhere
  \end{enumerate}

\end{proof}

\subsection{From \rcn{}es to \rpes{}es}\label{ssec:rcntorpes}

We now address the transformation of \rcn{}s into \rpes{}es, presenting the
encoding function as $\rcntorpes{\_}$.

\begin{definition}\label{def:rcntorpes}
  Let $\anRCN = \langle S, T, F, I, \mathsf{m}\rangle$ be an \rcn\ with
  backward transitions $\bwdset$. Then, the associated r\pes\ is
  $\rcntorpes \anRCN= (E, \anR, <, \#, \prec, \triangleright)$ where:
  \begin{enumerate}
  \item $E = T\setminus \bwdset$;
  \item
    $\anR = \{ t\ |\ \abwd\in \bwdset \wedge \afwd \in T \wedge \pre\afwd =
    \post\abwd\}$;
  \item ${<} = {\lessdot_{| E\times E}}$;
  \item ${\#} = \setcomp{(e, e')}{e\neq e'\in E\ \wedge e\ \cnconf\ e'}$;
  \item
    $\prec = \{ (\afwd, \abwd') \ |\ \afwd\in E \wedge \abwd' \in \bwdset
    \wedge {\pre \afwd\cap\inib{\abwd'}} \neq {\emptyset}\}$;
  \item
    $\triangleright = \{ (\afwd, \abwd') \ |\ \afwd\in E \wedge \abwd' \in
    \bwdset \wedge {\post \afwd\cap\inib{\abwd'}} \neq {\emptyset}\}$.
  \end{enumerate}
\end{definition}

The construction establishes a mapping from \rcn{} $\anRCN$ to an \rpes{}
where events correspond to the forward transitions $T\setminus \bwdset$ of
$\anRCN$ (Condition (1)). By Condition (2), only forward transitions with
corresponding reversing transitions in $\bwdset$ are designated as undoable
events (belonging to $U$). Conditions (3) and (4) specify that causality ($<$)
and conflict ($\#$) relations are obtained as proper restrictions of those
induced by the structure of $\anRCN$. Conditions (5) and (6) describe the
reconstruction of reverse causation ($\prec$) and prevention
($\triangleright$) from inhibitor arcs.

\begin{example}
  Consider the \rcn{} depicted in \Cref{ex:simple-rcn-net}. The events of the
  associated \rpes{} are $\setenum{a, b, c, d}$, with $\setenum{b, c}$
  identified as reversible events. The sole causal dependency is
  $b \lessdot c$. Conflicts arise from shared places in the presets of
  transitions, specifically $s_0$ and $s_2$, resulting in $a\ \#\ c$ and
  $b\ \#\ d$. The inhibitor arc connecting $s_3$ to $\un b$ is translated into
  the reverse causality $b \prec \un b$ because $s_3$ belongs to the preset of
  $b$. Similarly, the arcs from $s_4$ to $\un c$ and from $s_5$ to $\un c$ are
  mapped to $c \prec \un c$ and $d \prec \un c$, respectively. In contrast,
  the arc from $s_8$ to $\un b$ results in $c \triangleright \un b$ since
  $s_8$ is part of the postset of $c$. Consequently, the derived \rpes{}
  corresponds to the one defined in \Cref{ex:rpes-not-causal}.
\end{example}

\begin{proposition}\label{pr:rca-to-rpes}
  If $\anRCN$ is an \rcn{}, then $\rcntorpes \anRCN$ is an $\rpes$.
\end{proposition}

\begin{proof}
  Consider $\anRCN = \langle S, T, F, I, \mathsf{m}\rangle$ with backward
  transitions $\bwdset$, and let
  $\rcntorpes \anRCN= (\fwdset, \anR, <, \#, \prec, \triangleright)$ be
  defined as in \Cref{def:rcntorpes}. According to
  \Cref{de:reversible-causal-net}(1),
  $\langle S, \fwdset, F_{|\fwdset\times\fwdset}, I_{|\fwdset\times\fwdset},
  \mathsf{m}\rangle$ is a \ca. By \Cref{pr:pcn-to-ppes},
  $(\fwdset ,\lessdot_{| \fwdset \times\fwdset}, \#)$ is a p\pes. It remains
  to show the conditions in \Cref{de:rpes}.

  \begin{enumerate}
  \item According to the definition of $\rcntorpes{\_}$,
    $\prec = \{ (\afwd, \abwd) \ |\ \afwd\in \fwdset \wedge \abwd \in \bwdset
    \wedge {\pre \afwd\cap\inib{\abwd}} \neq {\emptyset}\}$. From
    \Cref{de:reversible-causal-net}(2), it is established that
    $S_{\afwd}\cap\inib{\abwd}\neq \emptyset$. Hence, $\afwd\prec\abwd$ holds
    for all $\abwd\in\bwdset$. Analogously, by using
    \Cref{de:reversible-causal-net}(3), it can be deduced that
    $\setcomp{t \in \fwdset}{t\prec\underline{\anr}}$ is finite and
    conflict-free for every $\anr\in\anR$.
  \item Immediate.
  \item If $t \prec \underline{\anr}$, then
    ${\pre t \ \cap\ \inib{\un \anr}} \neq {\emptyset}$ according to the
    definition of $\rcntorpes{\_}$. Based on
    \Cref{de:reversible-causal-net}(4), it is established that
    ${\post t\ \cap\ \inib{\un\anr}} = {\emptyset}$. Hence,
    $\neg(t \triangleright \underline{\anr})$ holds by the definition of
    $\rcntorpes{\_}$.
  \item It follows by setting $\lll = \ll$ and reasoning as it is done in the proof of \Cref{pr:rpes-to-rca}.
  \item It follows from \Cref{de:reversible-causal-net}(6).
  \end{enumerate}
\end{proof}

\subsection{Correspondence between \rcn{}es and \rpes{}es}\label{ssec:rcnandrpes}

The mappings introduced in the preceding sections establish a close
correspondence between \rpes{}es and \rcn{}s concerning configurations, as
will be evident in what follows.

\begin{definition}\label{de:rcn-pre-conf}
  Let $\anRCN = \langle S, T, F, I, \mathsf{m}\rangle$ be an \rcn. A {\em
    pre-configuration} $X$ of $\anRCN$ is a conflict-free set of forward
  transitions, i.e., $X \subseteq T \setminus\bwdset$ and
  $\forall t, t'\in X.\ \pre{t}\cap\pre{t'}\neq \emptyset\ \Rightarrow\ t =
  t'$. The \emph{associated marking} is
  $m_X = \mathsf{m} - \pre{X} + \post{X}$.
\end{definition}

If $X$ is a pre-configuration of $\anRCN$, it is noteworthy that $m_X$ may not
necessarily be a reachable marking. For example, consider the
pre-configuration $\{c,d\}$ of the \rcn{} in \Cref{ex:simple-rcn-net}. The
associated reachable marking $m_X = \{s_1, s_2, s_3, s_8, s_9\}$ is not
reachable itself, as firing $c$ is inhibited when $s_3$ is marked.

\begin{definition}\label{de:rcn-conf}
  A pre-configuration $X$ of an \rcn{} $\anRCN$ is a \emph{configuration} if
  $m_X$ is a reachable marking of $\anRCN$. The set of all configurations of
  $\anRCN$ is denoted by $\Conf{\anRCN}{\rcn}$.
\end{definition}


\begin{proposition}\label{pr:rcn-marktoconf}
  Let $\anRCN = \langle S, T, F, I, \mathsf{m}\rangle$ be an \rcn, and
  $m\in \reachMark{\anRCN}$ a reachable marking. Then,
  $X = \flt{\pre{m}}\cap (T\setminus\bwdset)$ is a configuration.
\end{proposition}
\begin{proof}
 Trivial.
\end{proof}

The subsequent auxiliary results are pivotal for establishing the proof of the
main results in this section.

\begin{proposition}\label{pr:rpes-rcn-preconf}
  Given an \rpes\ $\mathsf{P} = (E, \anR, <, \#, \prec, \triangleright)$,
  every conflict-free set of events $X \subseteq E$ is a pre-configuration of
  $\rpestorcn{\mathsf{P}}$.
\end{proposition}

\begin{proof}
  It follows straightforwardly by observing that the conflict-free condition
  for $X$ implies that in the corresponding \rpes{} $\rpestorcn{\mathsf{P}}$,
  for any pair of events $e$ and $e'$ in $X$,
  $\pre{e}\cap\pre{e'} = \emptyset$ holds.
\end{proof}

\begin{lemma}\label{lm:rpesc-rcnc-1}
  Let $\mathsf{P} = (E, \anR, <, \#, \prec, \triangleright)$ be an \rpes\ and
  $X\subseteq E$ a conflict-free set of events. If $A\cup\underline{B}$ is
  enabled at $X$, then $m_X\trans{A\cup\setcomp{(\anr,\re)}{\anr\in B}}$ in
  $\rpestorcn{\mathsf{P}}$.
\end{lemma}

\begin{proof}
  Assuming that $A\cup\underline{B}$ is enabled at $X$, we demonstrate that
  $m_X\trans{A\cup\setcomp{(\anr,\re)}{\anr\in B}}$.
  Since $X$ is conflict-free, it is a pre-configuration, and $m_X$ is the
  associated marking.
  The enabling of $A\cup\underline{B}$ at $X$ implies:

 \begin{enumerate}
 \item $A\cap X = \emptyset$, $B \subseteq X$ and $\CF{X\cup A}$,
 \item $\forall e\in A, e'\in E$. if $e' < e$ then $e'\in X\setminus B$,
 \item $\forall e\in B, e'\in E$. if $e' \prec \underline{e}$ then
   $e' \in X\setminus (B\setminus\setenum{e})$,
 \item $\forall e\in B, e'\in E$. if $e' \triangleright \underline{e}$ then
   $e'\not\in X\cup A$.
 \end{enumerate}
 We examine the different conditions:

 \begin{enumerate}
 \item The condition $A\cap X = \emptyset$ implies that
   $\pre{A}\subseteq m_X$, while $B \subseteq X$ means that
   $\post{B} \subseteq m_X$. Consequently,
   $B\times\setenum{\re}\subseteq m_X$. Finally, the conflict-freeness
   condition $\CF{X\cup A}$ is trivially verified as $\pre{A}\subseteq m_X$.
 \item For every $e\in A$ and $e'\in E$, if $e' < e$, then $e'\in X\setminus B$. Now,
   $e' < e$ implies that $(*,e') \in \inib{e}$, and since $e'\in X\setminus B$, we
   have  $m_X (*,e') = 0$.
 \item  For every $e\in B$ and $e'\in E$, if $e' \prec \underline{e}$, then
   $e' \in X\setminus (B\setminus\setenum{e})$.  Once more, as
   $e' \prec \underline{e}$, the inhibitor set  $\inib{(e,\re)}$ includes $(e',*)$.
   Since $e' \in X\setminus (B\setminus\setenum{e})$, it follows that
   $m_X(e',*) = 0$.
 \item For every $e\in B$ and $e'\in E$, if  $e' \triangleright \underline{e}$, then
   $e'\not\in X\cup A$. Once again, $e' \triangleright \underline{e}$ implies that
   $(e',*) \in \inib{(e,\re)}$, and since $e'\not\in X\cup A$, we have
   $m_X(e',*) = 1$, indicating that $e'$ must not be in $X$, as
   $m_X(e',*) = 1$ shows.
 \end{enumerate}

 Combining these arguments, it follows that
 $m_X\trans{A\cup\setcomp{(\anr,\re)}{\anr\in B}}$: in fact the first condition implies
 that $\pre{A} + \pre{(B\times\setenum{\re})} \subseteq m_X$, whereas the
 other three imply that all inhibiting places are empty.
\end{proof}

\begin{lemma}\label{lm:rpesc-rcnc-2}
  Let $\mathsf{P} = (E, \anR, <, \#, \prec, \triangleright)$ be an \rpes,
  $X \subseteq E$ a conflict-free set of events, and
  $Y \subseteq E\cup (\anR\times\setenum{\re})$ a set of transitions of
  $\rpestorcn{\mathsf{P}}$. If $m_X\trans{Y}$ then
  $(Y\cap E)\cup \{ \un u\ |\ (u, \re) \in Y\} $ is enabled at $X$.
\end{lemma}

\begin{proof}
  First, we observe that $X$ is a pre-configuration and, according to
  \Cref{pr:rpes-rcn-preconf}, $m_X = \mathsf{m} - \pre{X} + \post{X}$ is its
  associated marking. Assume that $m_X\trans{Y}$. We will demonstrate that
  $(Y\cap E)\cup (Y\cap \anR\times\setenum{\re})$ is enabled at $X$. We need
  to show that, denoting $A = (Y\cap E)$ and $B = (Y\cap \anR)$, the
  conditions of \Cref{de:rpes-conf-enab} hold:

 \begin{enumerate}
 \item For $A\cap X = \emptyset$, $B \subseteq X$ and $\CF{X\cup A}$, we
   proceed as follows.
   \begin{itemize}
   \item Observe that $m_X\trans{Y}$ implies
     $m_X\trans{A}$.  Then, $\pre{A}\subseteq m_X$, which implies
     $A\cap X = \emptyset$.
   \item From $B = (Y\cap \anR)$ and $m_X\trans{Y}$, we conclude
     $\pre{(Y\cap \anR\times\setenum{\re})}\subseteq m_X$. Consequently,
     $B = Y\cap\anR\subseteq X$,
   \item $\CF{X\cup A}$ holds trivially because $m_X\trans{A}$ and conflicts
     cannot be introduced in the net.
   \end{itemize}
 \item Condition $\forall e\in A, e'\in E$. if $e' < e$ then
   $e'\in X\setminus B$ is obtained as follows.

   We have that $((*,e'),e)\in I$ and $m_X\trans{Y}$, which means $e' \in X$.
   Additionally, $e'\not\in B$ because if $e'\in B$, it would imply that
   $m_X((e',)) = 1$ and $\post{Y}((e',)) = 1$, contradicting the fact that
   $m_X\trans{Y}$.
 \item Condition $\forall e\in B, e'\in E$. if $e' \prec \underline{e}$ then
   $e' \in X\setminus (B\setminus\setenum{e})$ is shown below:

   $e' \prec \underline{e}$ implies that $(*,e')\in \inib{(e,\re)}$. As
   $m_X\trans{Y}$ we have that $(e,\re)$ is in $Y$ and then $m_X((*,e')) = 0$,
   implying that $e' \in X$. Furthermore, $e'\not\in B$ because otherwise
   $m_X\trans{Y}$ would not hold. Thus, we can conclude that
   $e' \in X\setminus (B\setminus\setenum{e})$.

 \item Condition $\forall e\in B, e'\in E$. if $e' \triangleright \underline{e}$ then
   $e'\not\in X\cup A$ is shown below.

   The condition $e' \triangleright \underline{e}$ implies that
   $(e',*)\in \inib{(e,\re)}$. Given that $m_X\trans{Y}$ and $(e',*)$ is not
   marked, it follows that $e'\not\in X\cap A$.

 \end{enumerate}

 As all the conditions of \Cref{de:rpes-conf-enab} are satisfied, we conclude
 that $(Y\cap E)\cup (Y\cap \anR\times\setenum{\re})$ is enabled at $X$.
\end{proof}

\begin{lemma}\label{lm:rcnc-rpesc-1}
  Let $\anRCN = \langle S, T, F, I, \mathsf{m}\rangle$ be an \rcn,
  $X \subseteq T\setminus\bwdset$ a pre-configuration of $\anRCN$,
  $A\subseteq T\setminus\bwdset$ a set of forward transitions,
  $B\subseteq\bwdset$ a set of backward transitions, and
  $\un B = \{ \un u \ |\ u\in T \wedge \abwd \in \bwdset \wedge \post u = \pre
  \bwdset\}$. If $A\cup\underline{B}$ is enabled at $X$ in
  $\rcntorpes{\anRCN}$ then $m_X\trans{A\cup B}$.
\end{lemma}

\begin{proof}
 Along the same line of \Cref{lm:rpesc-rcnc-1}
\end{proof}

\begin{lemma}\label{lm:rcnc-rpesc-2}
  Let $\anRCN = \langle S, T, F, I, \mathsf{m}\rangle$, $X$ a
  pre-configuration of $\anRCN$, $A\subseteq T\setminus\bwdset$ a set of
  forward transitions, $B\subseteq\bwdset$ a set of backward transitions, and
  $\un B = \{ \un u \ | u\in T \wedge \abwd \in \bwdset \wedge \post u = \pre
  \bwdset\}$. If $m_X\trans{A\cup B}$ then $A \cup \un B$ is enabled at $X$ in
  $\rcntorpes{\anRCN}$.
\end{lemma}

\begin{proof}
 Following a similar line of reasoning as in \Cref{lm:rpesc-rcnc-2}.
\end{proof}


As a consequence of the above result, we have the operational correspondence
for $\rpestorcn{\_}$ and $\rcntorpes{\_}$ as stated below.

\begin{theorem}
  \label{th:conf-rpes-rcn}
  Let $\mathsf{P}$ be an \rpes. Then, $X \in \Conf{\mathsf{P}}{\rpes}$ iff
  $X \in \Conf{\rpestorcn{\mathsf{P}}}{\rcn}$.
\end{theorem}

\begin{proof}
  Follows from \Cref{lm:rpesc-rcnc-1} and \Cref{lm:rpesc-rcnc-2}
\end{proof}

\begin{theorem}
  \label{th:conf-rcn-rpes}
  Let $\anRCN$ be an \rcn. $X \in \Conf{\anRCN}{\rcn}$ iff
  $X \in \Conf{\rcntorpes{\anRCN}}{\rpes}$.
\end{theorem}

\begin{proof}
  Follows from \Cref{lm:rcnc-rpesc-1} and \Cref{lm:rcnc-rpesc-2}
\end{proof}

Combining the previous results, we conclude that \rcn{}s are a proper counterpart of \rpes{}es.

\begin{theorem}\label{th:corr-rpes-rcn}
  If $\anRCN$ is an \rcn{} then
  $\anRCN \equiv \rpestorcn{\rcntorpes{\anRCN}}$. If $\mathsf{P}$ is an
  \rpes{} then $\mathsf{P} \equiv \rcntorpes{\rpestorcn{\mathsf{P}}}$.
\end{theorem}

\section{Causal nets and Occurrence nets}\label{sec:discussion}

The developments in \Cref{ssec:pestocn,ssec:cntopes} have shown that \ca{s}
serve as counterparts to prime event structures, as highlighted in
\Cref{th:pesandcn}.
Now, we delve into the connections between \ca{s} and occurrence nets,
typically associated with prime event structures.
Finally, we present additional evidence that indicates the difficulty of
adapting occurrence nets to model out-of-causal order reversibility.

Let us start by introducing some key concepts about nets that will be
essential throughout this section. An \inet $N$ is deemed
\textit{single-execution} if its states are sets, meaning that for all
$X \in \states{N}$, $X = \flt{X}$. In simpler terms, each transition in a
single-execution net can be fired at most once in a firing sequence.
Single-execution nets (albeit without inhibitor arcs) are also known as
\emph{1-occurrence} nets~\cite{GP:CS,GP:CSESPN}.
The net $N_1$ in \Cref{ex:introa} is single-execution.

An \inet is said {\em conflict-saturated} if any pair of transitions, not
present in any firing sequence of the net, share a common place in their
presets.

\begin{definition}\label{de:sat-confl}
  Let $N = \langle S, T, F, I, \mathsf{m}\rangle$ be an \inet.
  Two transitions $t, t' \in T$ are deemed to be in \textit{conflict} if they
  do not simultaneously appear in any state of $N$, expressed as
  $\setenum{t, t'}\not\subseteq \flt{X}$ for all $X \in \states{N}$. The \inet
  $N$ is said \textit{conflict-saturated} when, for all $t, t' \in T$, if $t$
  and $t'$ are in conflict, then $\pre{t}\cap\pre{t'} \neq \emptyset$.
\end{definition}

A conflict-saturated net is one in which conflicts are characterised
structurally.

\begin{example}
  The \inet shown in \Cref{fig:conf-sat} is conflict-saturated since the
  absence of pairs $a$ and $c$, as well as $b$ and $c$, in any configuration
  of the net implies that each pair of transitions shares a place in their
  preset ($s_7$ for $a$ and $c$, and $s_3$ for $b$ and $c$).
  \begin{figure}[ht]
    \centerline{\scalebox{0.9}{\scalebox{0.9}{\begin{tikzpicture}
\tikzstyle{inhibitorred}=[o-, draw=red,thick]
\tikzstyle{inhibitorblu}=[o-, draw=blue,thick]
\tikzstyle{pre}=[<-,thick]
\tikzstyle{post}=[->,thick]
\tikzstyle{readblue}=[-, draw=blue,thick]
\tikzstyle{transition}=[rectangle, draw=black,thick,minimum size=5mm]
\tikzstyle{place}=[circle, draw=black,thick,minimum size=5mm]
\node[place,tokens=1] (p1) at (0,2.5) [label=left:$s_1$] {};
\node[place,tokens=1] (p3) at (2,2.5) [label=right:$s_2$] {};
\node[place,tokens=1] (p5) at (4,2.5) [label=right:$s_3$] {};
\node[place,tokens=1] (p7) at (2,0) [label=right:$s_7$] {};
\node[place] (p2) at (0,0) [label=left:$s_4$] {};
\node[place] (p4) at (1,1.25) [label=above:$s_5$] {};
\node[place] (p6) at (4,0) [label=right:$s_6$] {};
\node[transition] (a) at (0,1.25)  {$a$}
edge[pre] (p1)
edge[pre] (p7)
edge[post](p2);

\node[transition] (b) at (2,1.25) {$b$}
edge[pre] (p3)
edge[pre] (p5)
edge[post] (p4)
edge[inhibitorred, bend right] (p1)
;
\node[transition] (c) at (4,1.25)  {$c$}
edge[pre] (p5)
edge[pre] (p7)
edge[post] (p6);
\end{tikzpicture}}}}
    \caption{A conflict-saturated net}\label{fig:conf-sat}
    \Description{A conflict-saturated net}
  \end{figure}
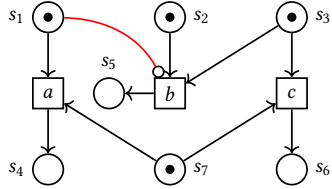
\end{example}

Every single-execution \inet can be transformed into a conflict-saturated,
equivalent one, i.e., without changing the executions of the net. This can be
achieved by adding a new place for any pair of conflicting transitions and
then extending the presets of the transitions with the corresponding new
places.
The formalisation of this procedure is provided below.

Consider a single-execution \inet $N = \langle S, T, F, I, \mathsf{m}\rangle$.
Define the set $\mathsf{confl}(N)$ of pairs of transitions in conflict as
\[
  \mathsf{confl}(N) = \setcomp{T'\subset T}{|T'| = 2\ \land\ \forall X\in
    \states{N}.\ T'\not\subseteq X}.
\]
Then, the conflict-saturated version of $N$ is given by
$N^{\mathsf{confl}(N)} = \langle S', T, F', I, \mathsf{m}'\rangle$ where
\begin{itemize}
\item $S' = S \cup\setcomp{s_T}{T'\in \mathsf{confl}(N)}$,
\item $F' = F\cup\setcomp{(s_T',t)}{t\in T'}$;
\item $\mathsf{m} = \mathsf{m} \cup\setcomp{s_T}{T'\in \mathsf{confl}(N)}$.
\end{itemize}
It should be noted that for every transition $t\in T$ and each reachable
marking $m\in \reachMark{N}$ such that $m\trans{t}$, there exists a reachable
marking $m'\in \reachMark{N^{\mathsf{confl}(N)}}$ such that $m'\trans{t}$ and
$\forall s\in S.\ m'(s) = m(s)$. In other words, the transformation preserves
the behaviour despite $m$ and $m'$ may differ in the places added by the
transformation. The correspondence also holds in the opposite direction: for
each transition $t\in T$ and each reachable marking
$m'\in \reachMark{N^{\mathsf{confl}(N)}}$ such that $m'\trans{t}$, there
exists a reachable marking $m\in \reachMark{N}$ such that $m\trans{t}$ and
$\forall s\in S.\ m'(s) = m(s)$.

The construction above provides a procedure for obtaining an equivalent,
conflict-saturated version of an \inet, as stressed by the following
proposition.

\begin{proposition}\label{pr:sat-confl-tot}
  Let $N$ be a single-execution \inet, then $N^{\mathsf{confl}(N)} \equiv N$.
\end{proposition}
\begin{proof}
  Straightforward, as $N$ is a single-execution \inet, the added places do not
  have any incoming arc and are initially marked, and the transitions in their
  postsets are in conflict.
\end{proof}

It is worth stressing that p\ca{}s and \ca{}s are single execution nets;
moreover, \ca{}s are also conflict-saturated.


\subsection{Occurrence nets}
We recall the notion of \emph{occurrence} nets\footnote{Occurrence nets are
  often the result of the \emph{unfolding} of a (safe) net.
  In this perspective an occurrence net is meant to describe precisely the
  non-sequential semantics of a net, and each reachable marking of the
  occurrence net corresponds to a reachable marking of the unfolded net. Here
  we focus purely on occurrence nets and not on the nets they are the
  unfoldings of, nor on the relation between the net and its unfolding.}.
We adopt the usual convention by which places and transitions are called
\emph{conditions} and \emph{events}; and write $B$ and $E$ for their
respective sets (instead of $S$ and $T$) and $\mathsf{c}$ for the initial
marking.
We may confuse conditions with places and events with transitions.
Moreover, we will omit the inhibiting relation $I$ since occurrence nets do
not have inhibitor arcs (i.e., $I = \emptyset$).

\begin{definition}\label{de:occnet}
  An \emph{occurrence net} (\cn) $O = \langle B, E, F, \mathsf{c}\rangle$ is
  an acyclic, safe net satisfying the following restrictions:
  \begin{enumerate}
  \item $\forall b\in B$. $\pre{b}$ is either empty or a singleton, and
    $\forall b\in \mathsf{c}$. $\pre{b} = \emptyset$,
  \item $\forall b\in B$. $\exists b'\in \mathsf{c}$ such that $b' \leq_O b$,
  \item for all $e\in E$ the set $\setcomp{e' \in E}{e'\leq_O e}$ is finite,  and
  \item $\#$ is an irreflexive and symmetric relation defined as follows:
    \begin{itemize}
    \item $e\ \#_0\ e'$ iff $e, e' \in E$, $e\neq e'$ and
      $\pre{e}\cap\pre{e'}\neq \emptyset$,
    \item $x\ \#\ x'$ iff $\exists y, y'\in E$ such that $y\ \#_0\ y'$ and
      $y \leq_O x$ and $y' \leq_O x'$.
    \end{itemize}
  \end{enumerate}
\end{definition}

Each condition $b$ in an \cn\ represents the occurrence of a token. Hence, $b$
either belongs to the initial marking $\mathsf{c}$ or is produced by the
\emph{unique} event in $\pre{b}$.
The flow relation is interpreted as the \emph{causality} relation among the
elements of the net; for this reason we say that $x$ \emph{causally depends}
on $y$ iff $y \leq_O x$.
Observe that ${\leq_O} \cap {(E\times E)}$ is a partial order and $\#$ is
inherited along $\leq_O$; hence, $e\ \#\ e' \leq_O e''$ implies $e\ \#\ e''$.

\begin{proposition}
  Let $O = \langle B, E, F, \mathsf{c}\rangle$ be an occurrence net. Then, $O$
  is a single execution net.
\end{proposition}

As for \Cref{pr:sat-confl-tot}, every single execution \inet\ can be converted
into an equivalent conflict-saturated net. Since the construction does not
modify the inhibitor relation, the same holds for occurrence nets.

\begin{definition}\label{de:cn-conf}
  Let $O = \langle B, E, F, \mathsf{c}\rangle$ be an \cn. A set of events
  $X\subseteq E$ is a \emph{configuration} of $O$ if:
  \begin{enumerate}
  \item $\forall e, e'\in X. e \neq e'\ \Rightarrow \neg (e\ \#\ e')$, i.e.,
    it is {\em conflict-free}; and
  \item $\forall e\in X$. $\hist{e}\subseteq X$, i.e., it is {\em
      left-closed}.
  \end{enumerate}

  The set of configurations of $O$ is denoted by $\Conf{O}{\cn}$.
\end{definition}

If a \cn{} is conflict-saturated then the first condition can be rewritten as
$\forall e, e'\in X. \pre{e}\cap\pre{e'}\neq\emptyset \Rightarrow\ e = e'$.
Moreover, observe that each configuration of an \cn{} is also a state.


\subsection{From \cn{} to \ca}\label{ssec:cntoca}
We demonstrate that it is possible to transform every occurrence net into a
causal one, establishing that the latter notion constitutes a conservative
extension of the former.

\begin{definition}\label{def:occtocausal}
  Given an \cn\ $O = \langle B, E, F, \mathsf{c}\rangle$ we can associate with
  it a net $\ontocn{O}$ defined as $\langle S, E, F', I, \mathsf{m}\rangle$
  where
  \begin{enumerate}
  \item
    $S = \setcomp{(\ast,e)}{e\in E}\cup\setcomp{(e,\ast)}{e\in E}\cup\
    \setcomp{(\setenum{e,e'},\#)}{e\ \#\ e'}$;
  \item
    $F' = \setcomp{(s,e)}{s = (\ast,e) \lor\ (s = (W,\#)\ \land\ e\in W)}\cup\
    \setcomp{(e,s)}{s = (e,\ast)}$;
  \item $I = \setcomp{(s,e)}{s = (\ast,e')\ \land\ e' <_C e}$; and
  \item $\mathsf{m} : S \to \nat$ is such that $\mathsf{m}(s) = 0$ if
    $s = (e,\ast)$ and $\mathsf{m}(s) = 1$ otherwise,
  \end{enumerate}
\end{definition}

The construction resembles the one from \pes{}es to \ca{}s: each event $e$ of
an occurrence net is associated with a homonymous transition and two places
$(\ast,e)$ and $(e,\ast)$ in the corresponding causal net. When marked,
$(\ast,e)$ represents the fact that $e$ has not been fired yet, while
$(e,\ast)$ describes the fact that $e$ has been executed. Moreover, there is
one additional place $(\setenum{e,e'},\#)$ for any pair of conflicting events
$e\#e'$ in $O$.
As expected, the preset of an event $e$ consists of the place ${(\ast,e)}$ and
all the places of the form $(\{e,e'\},\#)$, which represent the conflicts of
$e$ with some other event $e'$.
Causal dependencies are mapped into inhibitor arcs so that $(e', *)$ belongs
to the inhibitor set of $e$ only if $e$ causally depends on $e'$ in $O$. The
initial marking assigns a token to every place belonging to the preset of some
transition.

\begin{example}
  \Cref{ex:occtoca} shows the encoding of the \cn\ $O$ in
  \Cref{ex:simple-occ-net-a} as the \ca\ $\ontocn{O}$ in
  \Cref{ex:simple-occ-net}.

  \begin{figure}[th]
    \begin{subfigure}{.4\textwidth}
      \begin{center}
	\scalebox{0.8}{\begin{tikzpicture}[scale = 0.8]
\tikzstyle{inhibitor}=[o-,draw=red,thick]
\tikzstyle{pre}=[<-,thick]
\tikzstyle{post}=[->,thick]
\tikzstyle{read}=[-,thick]
\tikzstyle{transition}=[rectangle, draw=black,thick,minimum size=5mm]
\tikzstyle{place}=[circle, draw=black,thick,minimum size=5mm]
\node[place,tokens=1] (q1) at (2,3) {};
\node[place] (q2) at (1,.5) {};
\node[place] (q3) at (3,.5) {};
\node[place] (q4) at (3,-2) {};
\node[place,tokens=1] (q5) at (4.5,3) {};
\node[place] (q6) at (4.5,.5) {};
\node[place,tokens=1] (q7) at (2,.5) {};
\node[transition] (tt1) at (1,1.75)  {$\pmv{a}$}
edge[pre] (q1)
edge[pre] (q7)
edge[post] (q2);
\node[transition] (tt2) at (3,1.75)  {$\pmv{b}$}
edge[pre] (q1)
edge[post] (q3);
\node[transition] (tt3) at (3,-.75)  {$\pmv{c}$}
edge[pre] (q3)
edge[pre] (q7)
edge[post] (q4);
\node[transition] (tt4) at (4.5,1.75)  {$\pmv{d}$}
edge[pre] (q5)
edge[post] (q6);
\end{tikzpicture}}
      \end{center}
      \caption{$O$\label{ex:simple-occ-net-a}}
    \end{subfigure}
    \begin{subfigure}{.4\textwidth}
      \begin{center}
	\scalebox{0.8}{\begin{tikzpicture}[scale=.8]
\tikzstyle{inhibitorred}=[o-, draw=red,thick]
\tikzstyle{inhibitorblu}=[o-, draw=blue,thick]
\tikzstyle{pre}=[<-,thick]
\tikzstyle{post}=[->,thick]
\tikzstyle{readblue}=[-, draw=blue,thick]
\tikzstyle{transition}=[rectangle, draw=black,thick,minimum size=5mm]
\tikzstyle{place}=[circle, draw=black,thick,minimum size=5mm]
\node[place,tokens=1] (p0) at (2.8,3.5) [label=above:{$(\{a,c\},\#)$}] {};
\node[place,tokens=1] (p1) at (0,2.5) [label=above:{$(*,a)$}] {};
\node[place,tokens=1] (p2) at (1.9,2.5) [label=below:{$(\{a,b\},\#)$}] {};
\node[place,tokens=1] (p3) at (3.8,2.5) [label=left:{$(*,b)$}] {};
\node[place,tokens=1] (p4) at (5.7,2.5) [label=above:{$(*,c)$}] {};
\node[place,tokens=1] (p5) at (7.3,2.5) [label=above:{$(*,d)$}] {};
\node[place] (p6) at (0,0) [label=below:{$(a,*)$}] {};
\node[place] (p7) at (3.8,0) [label=below:{$(b,*)$}] {};
\node[place] (p8) at (5.7,0) [label=below:{$(c,*)$}] {};
\node[place] (p9) at (7.3,0) [label=below:{$(d,*)$}] {};
\node[transition] (t1) at (0,1.25)  {$a$}
edge[pre, bend left] (p0)
edge[pre] (p1)
edge[pre] (p2)
edge[post] (p6);
\node[transition] (t2) at (3.8,1.25)  {$b$}
edge[pre] (p2)
edge[pre] (p3)
edge[post] (p7);
\node[transition] (t3) at (5.7,1.25)  {$c$}
edge[pre, bend right] (p0)
edge[pre] (p4)
edge[inhibitorred] (p3)
edge[post] (p8);
\node[transition] (t4) at (7.3,1.25)  {$d$}
edge[pre] (p5)
edge[post] (p9);
\end{tikzpicture}}
      \end{center}
      \caption{$\ontocn{O}$\label{ex:simple-occ-net}}
    \end{subfigure}
    \caption{An occurrence net as a causal net}
     \Description{An occurrence net as a causal net}
    \label{ex:occtoca}
  \end{figure}
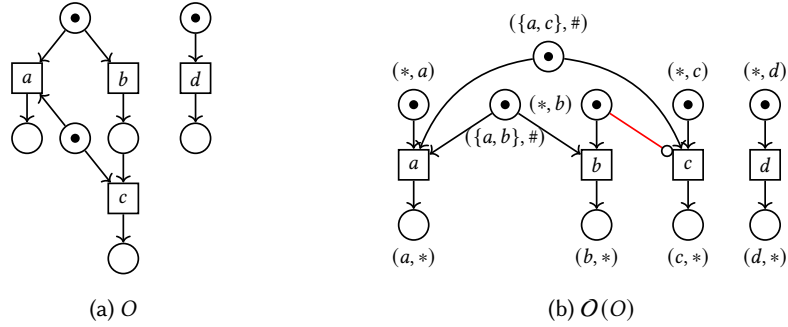

\end{example}

\begin{proposition}\label{pr:occtocausal}
  Let $O$ be an occurrence net. Then, $\ontocn{O}$ is a \ca\ and
  $O \equiv \ontocn{O}$.
\end{proposition}

\begin{proof}
  Consider the \cn\ $O = \langle B, E, F, \mathsf{c}\rangle$ and let
  $\ontocn{O} = \langle S, E, F', I, \mathsf{m}\rangle$ be its corresponding
  net. We first show that $\ontocn{O}$ is a \ca.

  The requirements (1), (2), and (7) in \Cref{de:pre-causal-net} are trivially satisfied by the construction of $\ontocn{O}$. For the third one, we proceed as follows: observe that $\pre{e}\cap\pre{e'}\neq\emptyset$ implies $e\#e'$
  and, by construction, $\pre{e}\cap\pre{e'} = \setenum{(\setenum{e,e'},\#)}$
  but $(\setenum{e,e'},\#)$ does not belong to the inhibitor set of any
  transition in $\ontocn{O}$. Condition (4) reduces to showing that
  $\hist{e}_\lessdot$ is finite, which holds because $\hist{e}$ is so in $O$.
  Condition (5) follows because $O$ is an \cn, and hence $F^{+}\cap E\times E$
  is acyclic. Finally observe that $\hist{e}_\lessdot = \hist{e}$ and this
  implies that for all $e', e''\in \hist{e}_\lessdot$, if $e'\cnconf e''$ then
  $e' = e''$.  Therefore, we can conclude that $\ontocn{O}$ is indeed a \ca.

  We now show that $\states{O} = \states{\ontocn{O}}$.
  \begin{itemize}
  \item $\states{O} \subseteq \states{\ontocn{O}}$. Assume $X\in \states{O}$
    and a firing sequence $\sigma$ such that $X = X_{\sigma}$. We prove by
    induction on the length of a firing sequence
    $\sigma\in\firseq{O}{\mathsf{c}}$ that there exists a firing sequence
    $\sigma'\in\firseq{\ontocn{O}}{\mathsf{m}}$ such that
    $X_\sigma = X_{\sigma'}$. The base case ($\len\sigma = 0$) is immediate
    since $\zero\in \states{O}$ and $\zero\in\states{\ontocn{O}}$ holds.
    For the inductive case, we assume that $\len\sigma = n$ and
    $\sigma\in\firseq{O}{\mathsf{c}}$ implies there exists
    $\sigma'\in\firseq{\ontocn{O}}{\mathsf{m}}$ such that
    $X_{\sigma} = X_{\sigma'}$. Assume now that $\sigma\trans{e}c'$. Then,
    $X_{\sigma} \cup\setenum{e}$ is conflict-free, and
    $\hist{e}\subseteq X_{\sigma}$. Hence, $X_{\sigma'} \cup\setenum{e}$ is
    conflict-free, and $\hist{e}_\lessdot\subseteq X_{\sigma'}$. Therefore,
    $\sigma'\trans{e}$.
    %

  \item $\states{\ontocn{O}} \subseteq \states{O}$ follows along the lines of
    the case above.
  \end{itemize}

Hence $O \equiv \ontocn{O}$.
\end{proof}

We stress that $\ontocn{O}$ is conflict-saturated for any \cn\ $O$.


\subsection{From p\ca{} to \cn}
The encoding from p\ca{}s to \cn{}s basically requires to embed causal
dependencies expressed in terms of inhibitor arcs into the the flow relation.
For this reason, the encoding incorporates new places of the form $(t,t')$ as
a way for representing the dependency $t\lessdot t'$.

\begin{definition}\label{def:pcatoocc}
  Let $C = \langle S, T, F, I, \mathsf{m}\rangle$ be a p\ca{}. The associated
  \cn{} is a net $\pcntoocc{C} = \langle B, T, F', \mathsf{c}\rangle$ where
  \begin{itemize}
  \item $B = S \cup \setcomp{(t,t')}{t\lessdot t'}$;
  \item
    $F' = F\cup \setcomp{(s,t)}{s = (t',t)} \cup\ \setcomp{(t,s)}{s =
      (t,t')}$; and
  \item $\mathsf{c} : B \to \nat$ is such that $\mathsf{c}(b) = \mathsf{m}(b)$
    if $b\in S$ and $\mathsf{c}(b) = 0$ if $b\not\in S$.
  \end{itemize}
\end{definition}

Note that the flow relation is extended to account for causal dependencies:
each transition $t$ produces tokens in the places $(t,t')$ (i.e.,
$(t, (t,t')) \in F$) and consumes tokens from $(t'',t)$ (i.e., $(t,(t'',t))$).
In this way, a transition $t$ is enabled when all its causes have been already
fired (i.e., the places $(t'',t)$ are marked).

The following results highlights that the transformation preserves the
behaviour of the net.

\begin{proposition}\label{pr:pcatoocc}
  Let $C$ be a p\ca, then $\pcntoocc{C}$ is an \cn{} and
  $C \equiv \pcntoocc{C}$.
\end{proposition}

\begin{proof}
  Take the \ca{} $C = \langle S, T, F, \mathsf{m}\rangle$. We first check that
  $\pcntoocc{C} = \langle B, T, F', \mathsf{c}\rangle$ is an \cn.
  Observe that each place in a \ca{} has at most one input arc. The new
  conditions introduced in the transformation have just one input transition,
  so for all $b\in B$, it holds that $|\pre{b}|\leq 1$. Furthermore, the added
  conditions are not marked, and since $\mathsf{c}$ is equal to $\mathsf{m}$
  when the condition in $S$ are considered, we have that for all
  $b\in \mathsf{c}$, $\pre{b} = \emptyset$ holds.
  The transitive closure of $F'$ restricted to $T\times T$ coincides with
  $\lessdot$, hence $\histtwo{t}{<_{F'}}$ is finite as $\histtwo{t}{\lessdot}$
  is finite, and furthermore $<_{F'}$ is an irreflexive partial order, and the
  net is acyclic.
  We prove that the conflict relation in $\pcntoocc{C}$ is inherited along the
  causality one. Assume that $t\ \#_0\ t'$ as
  $\pre{t}\cap\pre{t'}\neq \emptyset$ and assume that $t' < t''$, which means
  that there is a condition $b = (t',t'')$ such that $\pre{b} = t'$ and
  $\post{b} = t''$. As the token in the place in $\pre{t}\cap\pre{t'}$ is used
  by $t$, the condition $b$ cannot be marked and this implies that
  $t\ \#\ t''$, and $\pcntoocc{C}$ is indeed a \ca{}.

  We now show that $\states{C} = \states{\pcntoocc{C}}$.
  \begin{itemize}
  \item $\states{C} \subseteq \states{\pcntoocc{C}}$. We prove by induction on
    the length of $\sigma$ that $\sigma\in\firseq{C}{\mathsf{m}}$ implies that
    there exists $\sigma'\in\firseq{\pcntoocc{C}}{\mathsf{c}}$ such that
    $X_\sigma = X_{\sigma'}$. The base case ($\len\sigma = 0$) is immediate
    since $\zero\in \states{C}$ and $\zero\in\states{\pcntoocc{C}}$ as well.
    For the inductive case, we assume that $\len\sigma = n$ and
    $\sigma\in\firseq{C}{\mathsf{m}}$ implies there exists
    $\sigma'\in\firseq{\pcntoocc{C}}{\mathsf{c}}$ such that
    $X_{\sigma} = X_{\sigma'}$. Assume now that $\sigma\trans{t}m'$. Now as
    $\lead{\sigma}\trans{t}$ it means that $\forall s\in\inib{t}$.
    $\lead{\sigma}(s) = 0$ thus all the conditions $(t',t)$ are marked for all
    $t'\in\post{\inib{t}}$ and this implies that $\lead{\sigma'}\trans{t}$.
  \item $\states{\pcntoocc{C}} \subseteq \states{C}$ is proved similarly.
  \end{itemize}

Hence $C \equiv \pcntoocc{C}$.
\end{proof}

The net $\pcntoocc{C}$ in not necessarily conflict-saturated, as $C$ could be
not conflict-saturated.

\begin{example}
  In \Cref{ex:pcatoocc} we depict a p\ca{} $C$ (on the left) and the
  associated \cn{} (on the right), where the causal dependency of $c$ on $b$
  is represented by the place $(b, c)$.

  \begin{figure}[h]
    \begin{subfigure}{.2\textwidth}
      \begin{center}
	\scalebox{0.8}{\begin{tikzpicture}[scale=0.8]
\tikzstyle{inhibitorred}=[o-, draw=red,thick]
\tikzstyle{inhibitorblu}=[o-, draw=blue,thick]
\tikzstyle{pre}=[<-,thick]
\tikzstyle{post}=[->,thick]
\tikzstyle{readblue}=[-, draw=blue,thick]
\tikzstyle{transition}=[rectangle, draw=black,thick,minimum size=5mm]
\tikzstyle{place}=[circle, draw=black,thick,minimum size=5mm]
\node[place,tokens=1] (p1) at (0,2.5)  [label=above:{$s_0$}] {};
\node[place,tokens=1] (p2) at (1.5,2.5) [label=above:{$s_2$}] {};
\node[place,tokens=1] (p3) at (3,2.5) [label=above:{$s_3$}] {};
\node[place,tokens=1] (p4) at (5,2.5) [label=above:{$s_5$}] {};
\node[place] (p6) at (0,0) [label=below:{$s_1$}] {};
\node[place] (p7) at (3,0) [label=below:{$s_4$}] {};
\node[place] (p8) at (5,0) [label=below:{$s_6$}] {};
\node[transition] (t1) at (0,1.25)  {$a$}
edge[pre] (p1)
edge[pre] (p2)
edge[post] (p6);
\node[transition] (t2) at (3,1.25)  {$b$}
edge[pre] (p2)
edge[pre] (p3)
edge[post] (p7);
\node[transition] (t3) at (5,1.25)  {$c$}
edge[pre] (p4)
edge[inhibitorred] (p3)
edge[post] (p8);
\end{tikzpicture}}
      \end{center}
      \caption{$C$\label{ex:simple-pca-net-a}}
    \end{subfigure}\hspace*{2cm}
    \begin{subfigure}{.35\textwidth}
      \begin{center}
	\scalebox{0.8}{\begin{tikzpicture}[scale=0.8]
\tikzstyle{inhibitorred}=[o-, draw=red,thick]
\tikzstyle{inhibitorblu}=[o-, draw=blue,thick]
\tikzstyle{pre}=[<-,thick]
\tikzstyle{post}=[->,thick]
\tikzstyle{readblue}=[-, draw=blue,thick]
\tikzstyle{transition}=[rectangle, draw=black,thick,minimum size=5mm]
\tikzstyle{place}=[circle, draw=black,thick,minimum size=5mm]
\node[place,tokens=1] (p1) at (0,2.5)  [label=above:{$s_0$}] {};
\node[place,tokens=1] (p2) at (1.5,2.5) [label=above:{$s_2$}] {};
\node[place,tokens=1] (p3) at (3,2.5) [label=above:{$s_3$}] {};
\node[place,tokens=1] (p4) at (5.5,2.5) [label=above:{$s_5$}] {};
\node[place] (p6) at (0,0) [label=below:{$s_1$}] {};
\node[place] (p7) at (3,0) [label=below:{$s_4$}] {};
\node[place] (p8) at (5.5,0) [label=below:{$s_6$}] {};
\node[place] (p9) at (4.25,0) [label=below:{$(b,c)$}]  {};
\node[transition] (t1) at (0,1.25)  {$a$}
edge[pre] (p1)
edge[pre] (p2)
edge[post] (p6);
\node[transition] (t2) at (3,1.25)  {$b$}
edge[pre] (p2)
edge[pre] (p3)
edge[post] (p7)
edge[post] (p9);
\node[transition] (t3) at (5.5,1.25)  {$c$}
edge[pre] (p4)
edge[pre] (p9)
edge[post] (p8);
\end{tikzpicture}}
      \end{center}
      \caption{$\pcntoocc{C}$\label{ex:pcatocn-net}}
    \end{subfigure}
    \caption{A causal net as an occurrence net}
     \Description{A causal net as an occurrence net}
    \label{ex:pcatoocc}
  \end{figure}
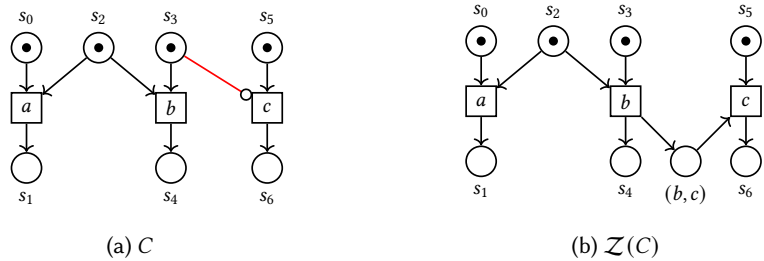
\end{example}

By combining \Cref{pr:occtocausal} and \Cref{pr:pcatoocc} we obtain the
following result.

\begin{theorem}\label{th:onandcn}
  Let $C$ be a causal net. Then $C = \ontocn{\pcntoocc{C}}$. Let $O$ be an
  occurrence net. Then $O \equiv \pcntoocc{\ontocn{O}}$.
\end{theorem}

Despite being able to directly obtain a \pes{} from a \cn{} and vice
versa~\cite{Win:ES}, the mappings introduced in
\Cref{ssec:pestocn,ssec:cntopes} provide an alternative procedure. This
procedure relies on the intermediate representation of \pes{}s in terms of
\ca{}s, as stated below.

\begin{theorem}\label{th:pesandcn2}
  Let $\mathsf{P}$ be a \pes. Then $\pcntoocc{\pestocn{\mathsf{P}}}$ is an
  \cn{} and
  $\Conf{\mathsf{P}}{\pes} = \Conf{\pcntoocc{\pestocn{\mathsf{P}}}}{\cn}$. Let
  $O$ be an \cn. Then $\cntopes{\ontocn{O}}$ is a \pes{} and
  $\Conf{O}{\cn} = \Conf{\cntopes{\ontocn{O}}}{\pes}$.
\end{theorem}


\subsection{Reversible occurrence net}
A causal-consistent reversible version of \cn{}s has be proposed
in~\cite{revPT,MMPPU:RC20}. This model associates each reversible event of an
occurrence net with a \emph{reversing} transition, similar to our concept of
\rcn{}s.
As already noticed in~\cite{MMPPU:RC20}, this approach encounters limitations
when dealing with the full generality of \rpes. The issue arises from the
tight correspondence between the configurations of an occurrence net and their
associated reachable markings. For an \cn{}
$O = \langle B, E, F, \mathsf{c}\rangle$ and a configuration $X$, the
associated reachable marking $m_X$ is computed as
$(\mathsf{c}\cup\flt{\post{X}})\setminus\flt{\pre{X}}$.
This equation makes clear that certain \emph{effects}---i.e., the produced
tokens--- resulting from the execution of events may remain {\em observable}
in the reachable marking. Specifically, if $e\in X$ and
$\post e \cap \pre X \neq \emptyset$, then $\post e \not\subseteq m_X$. This
scenario is exemplified in the configuration $\{b,c\}$ of the \cn{} depicted
in \Cref{ex:pcatocn-net}, where the condition $(b,c)$ is \emph{used} by the
firing of $c$ and ceases to hold after $c$ is executed.
Assuming that $b$ is an out-of-causal-order reversible event, we introduce the
reversing transition $\un b$ that consumes from $s_4$ and $(b,c)$.
Consequently, it becomes evident that $\un b$ is not enabled at $m_X$.
Consequently, the reversal of $b$ would require the reversal of $c$, thereby
enforcing a causally-ordered reversibility model.

\section{Applications}\label{sec:app}

\paragraph{Out-of-causal order behaviours}
Out-of-causal order reversibility was initially developed to model biochemical
reactions. However, in this section, we broaden its scope to explore
applications beyond biochemistry. We showcase its application in standard
programming languages, demonstrating its utility in scenarios such as
implementing the \emph{sagas pattern} \cite{sagas} for managing long-running
transactions.
The saga pattern has been employed in various contexts since its introduction
as a mechanism for managing database transactions. It has found applications
in workflow composition languages and service-oriented composition languages.
More recently, with the rise of microservices, the saga pattern has proven to
be a practical solution for implementing distributed long-running
transactions.
The concept of long-running transactions typically relies on a weaker notion
of atomicity based on compensations \cite{sagas}.
A saga breaks down the steps of a long-running transaction into a series of
short, local transactions. In a distributed setting, a local transaction
serves as the unit of work performed by a saga participant. Each step may be
accompanied by a compensating transaction---a transaction designed to reverse,
to the extent possible, the effects of its corresponding local transaction.
The core principle guiding saga execution is that the steps are carried out in
the order specified by the saga. However, if any step encounters a failure,
the compensations for the already executed steps are triggered in an order
consistent with the reversal of those executed steps.
Moreover, the saga pattern ensures that either all operations complete
successfully, or the corresponding compensation transactions are executed to
undo the work previously completed.
It is essential to emphasize that while perfect reversal is attainable in
certain situations, in several others, reversals are partial. Consequently,
the execution of a saga is not atomic in the conventional sense, as partial
effects of its execution may become observable when a saga is rolled back.

Consider a business process for handling orders, composed of the following
three services:

\begin{itemize}
\item $o$: Placement order service that registers information about the order.
\item $p$: Payment service responsible for charging the customer's credit
  card.
\item $f$: Fulfillment of the order.
\end{itemize}

Services $o$ and $p$ are designed to execute in parallel, while the activation
of $f$ depends on the completion of the first two actions. The fulfillment
service can either succeed or fail. In the former case, the process concludes
successfully; in the latter case, the process is aborted. When fulfillment
fails, compensation is required for the preceding steps.

While $o$ can be perfectly undone, i.e., there exists a reversing transaction
$\un o$ that completely rolls back the effects of performing $o$ (e.g.,
resetting the information in the database), the service $p$ is not reversible
but compensatable. In case of need, a transaction $r$ for refunding the charge on
the credit card should be invoked. For simplicity, we assume that neither
$o$ nor $p$ fails.

The above saga, can be represented as an $r\pes$
$\mathsf{P}_{\sf S} = (E, \anR, <, \#, \prec, \pr)$ defined as follows:
\[
  \begin{array}{l@{\hspace{2cm}}l@{\hspace{2cm}}l}
    E = \{o,p,f,a, r\}
    & \anR = \{o\}
    & {<} = \{(o, f), (p, f), (o, a), (p, a), (a, r) \} \\
    {\#} =\{(f, a), (a, f)\}
    & {\prec} = \{(o, \un{o}), (a, \un{o})\}
    & {\pr} = \{(f,\un{o})\}
  \end{array}
\]

The observable events in $E$ can be interpreted as follows:
\begin{itemize}
\item $o$: Successful completion of the placement order service.
\item $p$: Successful completion of the payment service.
\item $f$: Successful completion of the order fulfillment service.
\item $a$: Abortion of the process due to the failure of the fulfillment
  service.
\item $r$: Completion of the credit card refunding.
\end{itemize}

As described earlier, the sole reversible action is the placement of an order
($o$). The causal dependencies in $<$ dictate that the fulfillment event ($f$)
and the abortion ($a$) can only occur after $o$ and $p$ ocurr. Moreover, the
refunding $r$ can only happen after abortion $a$ occurs. The conflict relation
states that the process either completes successfully ($f$) or aborts ($a$).
The prevention relation $\pr$ states that the placement of an order cannot be
undone if the order has been fulfilled. The reverse causality dictates that
the placement of an order can be reversed only if the order has been placed
($o$) and the process has been aborted ($a$).

The \rpes described above can be associated with the \rcn{} depicted in
\Cref{fig:sagarcn} using the constructions provided in the previous sections.
We now comment on the structure of the net. The successful completion of $o$
and $p$ enabling $f$ is represented by the inhibitor arcs connecting $f$ to
$s_1$ and $s_2$. Similarly, the arcs connecting the abortion event $a$ to
$s_1$ and $s_2$ signify that the process can abort only after the execution of
$o$ and $p$. The fact that the process completes either with the successful
fulfillment $f$ or the abortion is captured by the events $f$ and $a$ being in
conflict due to the place in $s_4$. The refunding event $r$ is only executed
in case of abortion, as indicated by the arc connecting $s_5$ with $r$,
ensuring that the compensating event $r$ is enabled only when fulfillment
aborts (consuming the token in $s_5$). Analogously, the occurrence of $a$
enables the reversing of $o$. Conversely, when $f$ completes successfully, the
undoing of $o$ is prevented by the presence of the token in $s_9$ generated by
$f$.

\begin{figure}[!t]
  \begin{center}
    \scalebox{0.75}{\scalebox{0.9}{\begin{tikzpicture}
\tikzstyle{inhibitorred}=[o-, draw=red,thick]
\tikzstyle{pre}=[<-,thick]
\tikzstyle{post}=[->,thick]
\tikzstyle{transition}=[rectangle, draw=black,thick,minimum size=5mm]
\tikzstyle{place}=[circle, draw=black,thick,minimum size=5mm]
\tikzstyle{rev}=[rectangle, draw=gray,thick,minimum size=5mm]
\tikzstyle{prerev}=[<-,thick,draw=gray]
\tikzstyle{postrev}=[->,thick,draw=gray]

\node[place,tokens=1] (p1) at (0,2.5) [label=above:$s_1$] {};
\node[place,tokens=1] (p2) at (3,2.5) [label=above:$s_2$] {};
\node[place,tokens=1] (p5) at (6,2.5) [label=left:$s_3$] {};
\node[place,tokens=1] (p11) at (8,2.6) [label=above:$s_{4}$] {};
\node[place,tokens=1] (p7) at (9,2.5) [label=above:$s_5$] {};
\node[place,tokens=1] (p9) at (11,2.5) [label=above:$s_6$] {};

\node[place] (p3) at (0,0) [label=left:$s_7$] {};
\node[place] (p4) at (3,0) [label=left:$s_8$] {};
\node[place] (p6) at (6,0) [label=right:$s_9$] {};
\node[place] (p8) at (9,0) [label=left:$s_{10}$] {};
\node[place] (p10) at (11,0) [label=left:$s_{11}$] {};

\node[transition] (a1) at (0,1.25)  {$p$}
edge[pre] (p1)
edge[post](p3);

\node[transition] (a2) at (3,1.25) {$o$}
edge[pre] (p2)
edge[post] (p4)
;

\node[rev] (not2) at (4.5, 1) {$\underline {o}$}
edge[inhibitorred] (p2)
edge[prerev, bend left] (p4)
edge[postrev, bend right] (p2)
edge[inhibitorred] (p6)
edge[inhibitorred] (p7)
;

\node[transition] (a3) at (6, 1.10) {$f$}
edge[inhibitorred, bend right=5] (p1)
edge[inhibitorred, bend right=20] (p2)
edge[pre] (p11)
edge[pre] (p5)
edge[post] (p6)
;

\node[transition] (err) at (9, 1.25) {$a$}
edge[pre] (p7)
edge[post] (p8)
edge[pre] (p11)
edge[inhibitorred,bend right=30] (p1)
edge[inhibitorred,bend right=5] (p2)
;

\node[transition] (comp) at (11, 1.25) {$r$}
edge[inhibitorred] (p7)
edge[pre] (p9)
edge[post] (p10)
;

\end{tikzpicture}}}
  \end{center}
  \caption{The \rcn{} associated to the r\pes{} that models a saga
    $\mathsf{P}_{\sf S}$.}
    \Description{The \rcn{} associated to the r\pes{} that models a saga
    $\mathsf{P}_{\sf S}$.}
  \label{fig:sagarcn}
\end{figure}

The implementation of the sagas pattern often requires
ingenuity~\cite{richardson2018microservices,ford2021software}. A saga's
implementation involves coordinating the steps of the saga. The coordination
logic selects and instructs a participant to execute a local transaction. Upon
completion, the sequencing coordination selects and invokes the next saga
participant, continuing until all steps of the saga are executed. In the
context of microservices, it can be seen as a sequence of local transactions
where each update publishes an event, triggering the next update in the
sequence~\cite{richardson2018microservices}. As illustrated by the previous
example, the mappings from \rpes{}es to \rcn{} in the previous section make
explicit the messages that should be generated to activate the execution of
transitions. These constructions provide a helpful blueprint for implementing
the sagas pattern, ensuring that the necessary steps are coordinated, and
compensations are performed when needed.

\paragraph{Reversible debugging in shared memory models}

One successful application of reversibility is causal-consistent reversible
debugging\cite{GiachinoLM14,LaneseSUS22}. Reversible debuggers extend
classical ones by allowing the user to go back in the execution, thereby
making it easier to find the source of a bug. Causal-consistent reversible
debuggers further improve on reversible ones by leveraging causal information
while undoing computations. While this technique has been applied to
message-passing concurrent systems (e.g., actor-like languages), it has not
been applied to shared-memory-based concurrency so far.
One of the main challenges is accommodating asymmetric conflicts. Asymmetric
conflicts arise when two independent actions, both of which can occur in a
computation, cannot happen in any order. In other words, it may happen that
one action prevents the other, but not vice versa. Therefore, both actions can
coexist in a computation only if executed in a particular order.

\begin{figure}[th]
  \begin{center}
    \begin{lstlisting}[language=customc, xleftmargin=3.3cm,numbers=left]
      int x = 0; //global shared variable
      void *tOne(void *vargp)
      {
        int a = x; //thread local variable
      }

      int main()
      {
        pthread_t thread_id1;
        pthread_create(&thread_id1, NULL, tOne, NULL);
        x++;
        return 1;
      }
    \end{lstlisting}
    \caption{Snippet of C code}
    \Description{Snippet of C code}
    \label{fig:c-code}
  \end{center}
\end{figure}

\begin{figure}[t]
  \begin{subfigure}{.25\textwidth}
    \begin{lstlisting}[language=customc]
      a = 0, x = 1

      a = 1, x = 1
        \end{lstlisting}

    \caption{Configurations\label{fig:wmc}}
  \end{subfigure}\qquad\qquad
  \begin{subfigure}{.4\textwidth}
    \scalebox{0.75}{\scalebox{0.9}{\begin{tikzpicture}
\tikzstyle{inhibitorred}=[o-, draw=red,thick]
\tikzstyle{pre}=[<-,thick]
\tikzstyle{post}=[->,thick]
\tikzstyle{transition}=[rectangle, draw=black,thick,minimum size=5mm]
\tikzstyle{place}=[circle, draw=black,thick,minimum size=5mm]
\tikzstyle{rev}=[rectangle, draw=gray,thick,minimum size=5mm]
\tikzstyle{prerev}=[<-,thick,draw=gray]
\tikzstyle{postrev}=[->,thick,draw=gray]

\node[place,tokens=1] (p2) at (4,2.5) [label=above:$s_1$] {};
\node[place,tokens=1] (p5) at (6,2.5) [label=above:$s_2$] {};
\node[place,tokens=1] (p4b) at (7,3.4) [label=below:$s_{3}$] {};
\node[place,tokens=1] (p7) at (8,2.5) [label=left:$s_4$] {};
\node[place,tokens=1] (p9) at (10,2.5) [label=above:$s_5$] {};

\node[place] (p4) at (4,0) [label=left:$s_{6}$] {};
\node[place] (p6) at (6,0) [label=right:$s_{7}$] {};
\node[place] (p8) at (8,0) [label=left:$s_{8}$] {};
\node[place] (p10) at (10,0) [label=left:$s_{9}$] {};


\node[transition] (a2) at (4,1.25) {$a_0$}
edge[inhibitorred] (p5)
edge[pre, bend left=30] (p4b)
edge[pre] (p2)
edge[post] (p4)
;

\node[transition] (a3) at (6, 1.25) {$x_0$}
edge[pre] (p5)
edge[post] (p6)
;

\node[transition] (err) at (8, 1.25) {$x_1$}
edge[pre] (p7)
edge[post] (p8)
edge[inhibitorred] (p5)
;

\node[transition] (comp) at (10, 1.25) {$a_1$}
edge[inhibitorred] (p7)
edge[pre, bend right] (p4b)
edge[pre] (p9)
edge[post] (p10)
;

\end{tikzpicture}}}
    \caption{$N_V$\label{fig:wmf}}
  \end{subfigure}
  \caption{An \ca{} model of the C Code}\label{fig:wm}
  \Description{An \ca{} model of the C Code}
\end{figure}
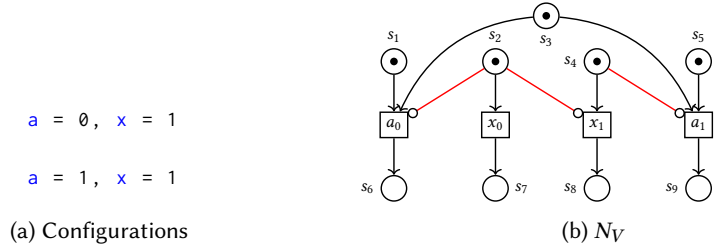

Asymmetric conflicts are typical in shared memory scenarios. Let us consider
the snippet of C code listed in \Cref{fig:c-code}.
This fragment of code illustrates a scenario where a thread read the
current value of a shared global variable,  storing it in a local
variable. A global variable \CI{x} is declared and initialised to \CI{0} (line
1). Then, a thread function (\CI{tOne}) is defined.
The thread function declares a local variable (\CI{a}), initialized with the
current value of the global variable \CI{x}.
The main function creates a thread, assigning it to execute the \CI{tOne}
function. Then, the global variable \CI{x} is
incremented.
The order of execution of the thread and the main program is not guaranteed,
leading to non-deterministic behaviour and potential race conditions. That is,
depending on how the scheduler executes the thread, its local variable
can be initialised with either $0$ or $1$ (assuming sequential consistency).
Therefore, any valid and complete execution of the program lead the system to
one of the configurations (i.e., state variables) in \Cref{fig:wmc}.

The program's behavior can be effectively described using a \pes.
Specifically, one approach is to describe computations in terms of the values
assigned to variables. Accordingly, events may be selected to represent these
value assignments. For clarity, we denote events such as $x_0$ and $x_1$ to
signify instances where variable \CI{x} is assigned the values \CI{0} and
\CI{1}, respectively. Analogously, events like $a_0$ and $a_1$ are employed
for the variable \CI{a}.

Consequently, the program's behavior can be defined using the \pes\
$\mathsf{P_V} = (E, <, \#)$, where
\[
  \begin{array}{l@{\hspace{1.5cm}}l@{\hspace{1.5cm}}l}
    E = \{a_0,a_1, x_0,x_1\}
    & {<} = \{(x_0,x_1), (x_0,a_0),  (x_1,a_1)  \}  &
    {\#} =\{(a_0, a_1), (a_1, a_0)\}
  \end{array}
\]

The causality relation establishes that $x_0$ is the minimal event since the
assignment of \CI{0} to \CI{x} is the first one to occur in the program. The
relation $x_0 < x_1$ signifies that the assignment of \CI{1} due to the
increment in the program follows the initial assignment. The remaining pairs
in the definition of the causality relation capture the dependency of the
value assigned to the local variable on the value of the global variable. For
example, $(x_0,a_0)$ indicates that the assignment to \CI{a} occurs after
reading the value \CI{0} from \CI{x}, while $(x_1, a_1)$ denotes the case
where the value read from \CI{x} is \CI{1}.

Furthermore, it is noteworthy that both $x_0$ and $x_1$ coexist in every
complete execution, as \CI{x} is initially assigned \CI{0} and subsequently
\CI{1}. This is not the case for the local variable \CI{a}, if \CI{a} is
assigned \CI{0} in one execution, it cannot be assigned \CI{1} in the same
execution (and vice versa). This fact is captured by the conflict relation,
which states that the different assignments to the local variable are in
conflict, i.e., $a_0 \# a_1$.

The network depicted in \Cref{fig:wmf} is obtained by applying the encoding
outlined in the previous sections. While the proposed model accurately
captures maximal configurations, it is noteworthy that it exhibits a weakness
in constraining the order in which certain assignments can take place. For
instance, the state $\{ x_0, x_1 \}$ allows the firing of ${a_0}$. This
suggests that the model does not explicitly account for fact that the value
assigned to \CI{a} has been read before the assignment of \CI{1} to \CI{x}.
While this behavior may be acceptable in certain contexts, its justification
becomes less evident when utilising the model for debugging.

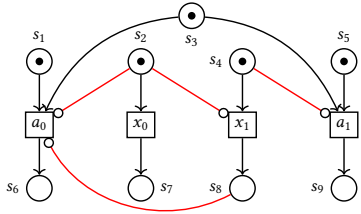
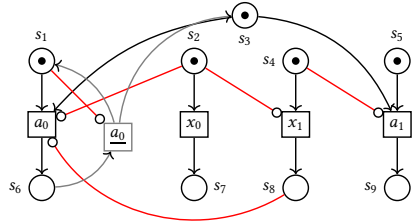
\begin{figure}[t]
  \begin{subfigure}{.55\textwidth}
  \begin{center}
    \scalebox{0.75}{\scalebox{0.9}{\begin{tikzpicture}
\tikzstyle{inhibitorred}=[o-, draw=red,thick]
\tikzstyle{pre}=[<-,thick]
\tikzstyle{post}=[->,thick]
\tikzstyle{transition}=[rectangle, draw=black,thick,minimum size=5mm]
\tikzstyle{place}=[circle, draw=black,thick,minimum size=5mm]
\tikzstyle{rev}=[rectangle, draw=gray,thick,minimum size=5mm]
\tikzstyle{prerev}=[<-,thick,draw=gray]
\tikzstyle{postrev}=[->,thick,draw=gray]

\node[place,tokens=1] (p2) at (4,2.5) [label=above:$s_1$] {};
\node[place,tokens=1] (p5) at (6,2.5) [label=above:$s_2$] {};
\node[place,tokens=1] (p4b) at (7,3.4) [label=below:$s_{3}$] {};
\node[place,tokens=1] (p7) at (8,2.5) [label=left:$s_4$] {};
\node[place,tokens=1] (p9) at (10,2.5) [label=above:$s_5$] {};

\node[place] (p4) at (4,0) [label=left:$s_{6}$] {};
\node[place] (p6) at (6,0) [label=right:$s_{7}$] {};
\node[place] (p8) at (8,0) [label=left:$s_{8}$] {};
\node[place] (p10) at (10,0) [label=left:$s_{9}$] {};

\node[transition] (a2) at (4,1.25) {$a_0$}
edge[inhibitorred] (p5)
edge[inhibitorred, bend right=45] (p8)
edge[pre, bend left=30] (p4b)
edge[pre] (p2)
edge[post] (p4)
;

\node[transition] (a3) at (6, 1.25) {$x_0$}
edge[pre] (p5)
edge[post] (p6)
;

\node[transition] (err) at (8, 1.25) {$x_1$}
edge[pre] (p7)
edge[post] (p8)
edge[inhibitorred] (p5)
;

\node[transition] (comp) at (10, 1.25) {$a_1$}
edge[inhibitorred] (p7)
edge[pre, bend right] (p4b)
edge[pre] (p9)
edge[post] (p10)
;

\end{tikzpicture}}}
    \end{center}
    \caption{$N_V$ extended with the prevention relation between forward
      transitions\label{fig:wmp}}
  \end{subfigure}
  \begin{subfigure}{.55\textwidth}
  	\begin{center}
    \scalebox{0.75}{\scalebox{0.9}{\begin{tikzpicture}
\tikzstyle{inhibitorred}=[o-, draw=red,thick]
\tikzstyle{pre}=[<-,thick]
\tikzstyle{post}=[->,thick]
\tikzstyle{transition}=[rectangle, draw=black,thick,minimum size=5mm]
\tikzstyle{place}=[circle, draw=black,thick,minimum size=5mm]
\tikzstyle{rev}=[rectangle, draw=gray,thick,minimum size=5mm]
\tikzstyle{prerev}=[<-,thick,draw=gray]
\tikzstyle{postrev}=[->,thick,draw=gray]

\node[place,tokens=1] (p2) at (3,2.5) [label=above:$s_1$] {};
\node[place,tokens=1] (p5) at (6,2.5) [label=above:$s_2$] {};
\node[place,tokens=1] (p4b) at (7,3.4) [label=below:$s_{3}$, label=above:$~~$] {};
\node[place,tokens=1] (p7) at (8,2.5) [label=left:$s_4$] {};
\node[place,tokens=1] (p9) at (10,2.5) [label=above:$s_5$] {};

\node[place] (p4) at (3,0) [label=left:$s_{6}$] {};
\node[place] (p6) at (6,0) [label=right:$s_{7}$] {};
\node[place] (p8) at (8,0) [label=left:$s_{8}$] {};
\node[place] (p10) at (10,0) [label=left:$s_{9}$] {};

\node[transition] (a2) at (3,1.25) {$a_0$}
edge[inhibitorred] (p5)
edge[inhibitorred, bend right=45] (p8)
edge[pre, bend left=20] (p4b)
edge[pre] (p2)
edge[post] (p4)
;

\node[rev] (not2) at (4.5, 1) {$\underline {a_0}$}
edge[prerev, bend left] (p4)
edge[postrev, bend right] (p2)
edge[inhibitorred] (p2)
edge[postrev, bend left=40] (p4b)
;

\node[transition] (a3) at (6, 1.25) {$x_0$}
edge[pre] (p5)
edge[post] (p6)
;

\node[transition] (err) at (8, 1.25) {$x_1$}
edge[pre] (p7)
edge[post] (p8)
edge[inhibitorred] (p5)
;

\node[transition] (comp) at (10, 1.25) {$a_1$}
edge[inhibitorred] (p7)
edge[pre, bend right] (p4b)
edge[pre] (p9)
edge[post] (p10)
;

\end{tikzpicture}}}
    \end{center}
    \caption{Reversible version of the extended model\label{fig:wmr}}
  \end{subfigure}
  \caption{A reversible model of the C code}\label{fig:wm-ff}
  \Description{A reversible model of the C code}
\end{figure}

Consider a scenario where the program has reached its maximal configuration
$\{x_0, x_1, a_0\}$ with reverse debugging activated. Given that $a_0$ is a
maximal event, a reverse of $a_0$ would generate tokens in $s_1$ and $s_3$,
consequently enabling both $a_1$ and $a_0$. Notice that, if computing forward,
the execution of $a_0$ should be precluded, as $x_1$ is already present in the
configuration, and thus, the variable \CI{x} holds the value \CI{1}.
The model must explicitly state that the occurrence of event $a_0$ should be
precluded when the variable $x$ has value $1$. We can enhance the model by
introducing inhibitor arcs to account for these dependencies.
In our $\rcn{}$, we use inhibitor arcs between the postset of a transition and
a reversing transition, signifying the prevention relation. This mechanism can
be extended to handle asymmetric conflicts, such as those between $a_0$ and
$x_1$. Specifically, we can broaden the prevention relation to include forward
events, e.g., $a_0 \pr x_1,$ indicating that $a_0$ is prevented by the
presence of $x_1$. This relationship can be effectively modelled with an
inhibitor arc linking the postset of $x_1$ to the transition $a_0$, as
illustrated in \Cref{fig:wmp}. Note the inhibitor arc connecting $a_0$ with
$s_{8}$.

Then, an accurrate reversible model of the code is straightforwardly obtained
by adding the reversing transitions in a standard fashion, as depicted in
\Cref{fig:wmr}.


\section{Conclusions} \label{sec:conc}

The main contribution of this paper is the characterisation of a class of
Petri nets with inhibitor arcs, dubbed {\em reversible causal nets} (\rcn),
which provides an operationally counterpart for the reversibility model behind
reversible prime event structures (\rpes)~\cite{PU:jlamp15}.
The key observation is that three out of the four (primitive) relations that
\rpes{}es define over events can be captured by inhibitor arcs: both {\em
  forward and reverse causality} can be modelled as inhibitor arcs that
connect an event (either forward or reversing) to (some of the conditions in)
the presets of its immediate causes, while {\em prevention} is translated to
arcs that connect an event to (some of the conditions in) the postsets of the
preventing events. The remaining {\em conflict relation} corresponds, as
usual, to overlapped presets.

\begin{figure}[t]
  \centering{ \scalebox{0.9}{\begin{tikzpicture}
\usetikzlibrary{decorations.pathmorphing}

\tikzstyle{ext}=[->,thick]
\tikzstyle{extnew}=[left hook->,thick,dashed]

\tikzstyle{old}=[<->,thick]
\tikzstyle{new}=[<->,thick,red]
\tikzstyle{transition}=[rectangle, draw=none,thick,minimum size=5mm]
\node[transition] (rcn) at (0,0) {$r\ca$}
;
\node[transition] (rpes) at (2,0) {$r\pes$}
;

\node[transition] (cn) at (0,2)  {$\ca$}
edge[extnew] (rcn)
;

\node[transition] (pes) at (2,2)  {$\pes$}
;
\node[transition](ron) at (6,2){$r\cn$}
;

\node[transition] (o) at (4,2)  {$\cn$}
;

\node[transition] (crpes) at (6,0)  {$cr\pes$};

 \draw[<->, thick] (3.7,2) -- (2.4,2) node [pos=.50, label=above:{\footnotesize\cite{Win:ES}}] {};
 \draw[<->, thick,dashed,bend left=50] (3.8,1.9) to  node[below] {\footnotesize Th.~\ref{th:pesandcn2}} (2.3,1.9) [out=150, in=30] ;

 \draw[<-left hook,thick](2,0.2) -- (2,1.8)node [pos=.50, label=left:{\footnotesize\cite{PU:jlamp15}}] {};
 \draw[right hook->,thick](4.4,2) -- (5.6,2)node [pos=.50, label=above:{\footnotesize\cite{MMPPU:RC20}}] {};
 \draw[<->,thick](6,1.8) -- (6,0.2)node [pos=.50, label=right:{\footnotesize\cite{MMPPU:RC20}}] {};
 \draw[<-left hook, thick] (2.4,0) -- (5.5,0) node [pos=.50, label=below:{\footnotesize\cite{PU:jlamp15}}] {};

  \draw[<->, thick,dashed] (0.3,2) -- (1.7,2) node [pos=.50, label=above:{\footnotesize Th.~\ref{th:pesandcn}}] {};
  \draw[<->, thick,dashed] (0.35,0) -- (1.6,0) node [pos=.50, label=below:{\footnotesize Th.~\ref{th:corr-rpes-rcn}}] {};
  \draw[<->, thick,dashed,bend left=50] (0,2.2) to  node[above] {\footnotesize Th.~\ref{th:onandcn}} (4,2.2) [out=150,in=30] ;

\end{tikzpicture}} }

  \caption{Recap of results. Solid arrows represent known results/works in the
    literature while dashed arrows stand for results in this paper.
    Double-headed arrows are used for correspondences while curly tailed
    arrows are for inclusions and reversible extensions.}

  \Description{Recap of results. Solid arrows represent known results/works in
    the literature while dashed arrows stand for results in this paper.
    Double-headed arrows are used for correspondences while curly tailed
    arrows are for inclusions and reversible extensions.}

  \label{fig:res}
\end{figure}
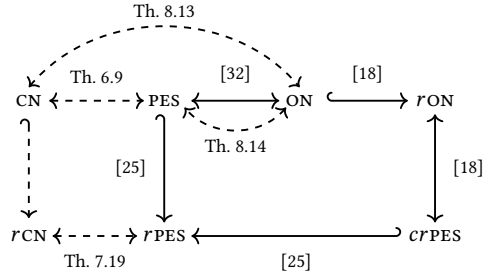

\Cref{fig:res} briefly recaps our results with respect to the state of the
art. The correspondence between occurrence nets and prime event structures is
due to Winskel's seminal work~\cite{Win:ES}. We have introduced a new class of
Petri nets, called {\em causal nets}, and showed their correspondence with
prime event structures (\Cref{th:pesandcn}). We have then presented the
reversible variant of causal nets (\rcn{}s), and showed that they are an
operational counterpart of the reversible prime event structures of
\cite{PU:jlamp15} (\Cref{th:corr-rpes-rcn}). Finally, we showed that causal
nets are tightly connected to occurrence nets (\Cref{th:onandcn}), and give a
revisitation of Winskel's correspondence via \Cref{th:pesandcn2}. To give a
complete picture of known results, we depict the extension of occurrence nets
into reversible occurrence nets proposed in~\cite{MMPPU:RC20}; where the
latter ones correspond to a proper subclass of $r\pes$ that just accounts for
causally-consistent reversibility (cr\pes).

Besides contributing to the well-established line of work initiated
in~\cite{NPW:PNES,Win:ES} that links classes of nets with classes of event
structures, our work explains the reversing mechanisms of \rpes{}es by relying
on a well-studied concurrent model. In this respect, our contribution
constitutes a step forward with respect to \cite{MMPPU:RC20}, which just deals
with the subclass of {\em causal-reversible} \rpes{}es which, e.g., do not
provide out-of-causal-order reversibility.
Remarkably, our reversible causal nets do not appeal to original elements to
achieve reversibility as, e.g., the {\em bonds} of \cite{PhilippouP18}. As a
matter of fact, causal nets do not rely either on the information (i.e.,
colours) carried on by tokens (as done in \cite{BarylskaGMPPP18,revPT}).

The characterisation of \rpes{}es as restrictions over the structure of the
nets enables the study of less constrained models. For instance, an undoable
event in an \rpes can be undone just by performing one reversing event. This
excludes the possibility of defining alternative conditions for reversing an
event. We can overcome this limitation in the operational model just by
dropping the requirement of non-overlapping backward transitions (i.e.,
$\flt{\pre \bwdset}={\pre \bwdset}$).
This kind of analysis seems an interesting line to explore in future work.

Causal nets and their reversible versions play a role analogous to occurrence
nets (in their connection with \pes{}es). Consequently, we may ask ourselves
about the class of (reversible) nets whose unfoldings are (reversible) causal
nets, or, alternatively, which are the suitable categories that would allow
us, if possible, to extend Winskel's chain of coreflection to reversible event
structures.

\section*{Acknowledgment}
This work has been partially supported by the BehAPI project funded by the EU H2020 RISE under the
Marie Sklodowska-Curie action (No: 778233), by the Italian PRIN 2020 project NiRvAna -- Noninterference
and Reversibility Analysis in Private Blockchains, the Italian PRIN 2022 project
DeKLA -- Developing Kleene Logics and their Applications,
 the INdAM-GNCS E53C22001930001 project RISICO -- Reversibilit\`a
in Sistemi Concorrenti: Analisi Quantitative e Funzionali, and the European Union - NextGenerationEU
SEcurity and RIghts in the CyberSpace (SERICS) Research and Innovation Program PE00000014, projects
STRIDE and SWOP.
\bibliographystyle{abbrv}
\bibliography{biblio-note}

\begin{thebibliography}{10}

\bibitem{wg1}
B.~Aman, G.~Ciobanu, R.~Gl{\"{u}}ck, R.~Kaarsgaard, J.~Kari, M.~Kutrib,
  I.~Lanese, C.~A. Mezzina, L.~Mikulski, R.~Nagarajan, I.~C.~C. Phillips, G.~M.
  Pinna, L.~Prigioniero, I.~Ulidowski, and G.~Vidal.
\newblock Foundations of reversible computation.
\newblock In I.~Ulidowski, I.~Lanese, U.~P. Schultz, and C.~Ferreira, editors,
  {\em Reversible Computation: Extending Horizons of Computing - Selected
  Results of the {COST} Action {IC1405}}, volume 12070 of {\em Lecture Notes in
  Computer Science}, pages 1--40. Springer, 2020.

\bibitem{BBCP:rivista}
P.~Baldan, N.~Busi, A.~Corradini, and G.~M. Pinna.
\newblock Domain and event structure semantics for {P}etri nets with read and
  inhibitor arcs.
\newblock {\em Theoretical Computer Science}, 323(1-3):129--189, 2004.

\bibitem{BarylskaGMPPP18}
K.~Barylska, A.~Gogolinska, L.~Mikulski, A.~Philippou, M.~Piatkowski, and
  K.~Psara.
\newblock Reversing computations modelled by coloured petri nets.
\newblock In W.~M.~P. van~der Aalst, R.~Bergenthum, and J.~Carmona, editors,
  {\em Proceedings of the International Workshop on Algorithms {\&} Theories
  for the Analysis of Event Data 2018 Satellite event of the conferences: 39th
  International Conference on Application and Theory of Petri Nets and
  Concurrency Petri Nets 2018 and 18th International Conference on Application
  of Concurrency to System Design {ACSD}}, volume 2115 of {\em {CEUR} Workshop
  Proceedings}, pages 91--111. CEUR-WS.org, 2018.

\bibitem{Bou:FESFN}
G.~Boudol.
\newblock Flow event structures and flow nets.
\newblock In I.~Guessarian, editor, {\em Semantics of Systems of Concurrent
  Processes}, volume 469 of {\em Lecture Notes in Computer Science}, pages
  62--95. Springer, 1990.

\bibitem{CaPi:PN14}
G.~Casu and G.~M. Pinna.
\newblock Flow unfolding of multi-clock nets.
\newblock In G.~Ciardo and E.~Kindler, editors, {\em {Petri} {Nets} 2014},
  volume 8489 of {\em Lecture Notes in Computer Science}, pages 170--189.
  Springer, 2014.

\bibitem{CP:soap17}
G.~Casu and G.~M. Pinna.
\newblock {P}etri nets and dynamic causality for service-oriented computations.
\newblock In A.~Seffah, B.~Penzenstadler, C.~Alves, and X.~Peng, editors, {\em
  Proceedings of {SAC} 2017}, pages 1326--1333. {ACM}, 2017.

\bibitem{DanosK05}
V.~Danos and J.~Krivine.
\newblock Transactions in {RCCS}.
\newblock In M.~Abadi and L.~de~Alfaro, editors, {\em Concurrency Theory, 16th
  International Conference, {CONCUR} 2005}, volume 3653 of {\em Lecture Notes
  in Computer Science}, pages 398--412. Springer, 2005.

\bibitem{ford2021software}
N.~Ford, M.~Richards, P.~Sadalage, and Z.~Dehghani.
\newblock {\em Software Architecture: The Hard Parts}.
\newblock " O'Reilly Media, Inc.", 2021.

\bibitem{sagas}
H.~Garcia{-}Molina and K.~Salem.
\newblock Sagas.
\newblock In U.~Dayal and I.~L. Traiger, editors, {\em Proceedings of the
  Association for Computing Machinery Special Interest Group on Management of
  Data 1987 Annual Conference}, pages 249--259. {ACM} Press, 1987.

\bibitem{GiachinoLM14}
E.~Giachino, I.~Lanese, and C.~A. Mezzina.
\newblock Causal-consistent reversible debugging.
\newblock In S.~Gnesi and A.~Rensink, editors, {\em Fundamental Approaches to
  Software Engineering - 17th International Conference, {FASE} 2014}, volume
  8411 of {\em Lecture Notes in Computer Science}, pages 370--384. Springer,
  2014.

\bibitem{biorev}
S.~Kuhn, B.~Aman, G.~Ciobanu, A.~Philippou, K.~Psara, and I.~Ulidowski.
\newblock Reversibility in chemical reactions.
\newblock In I.~Ulidowski, I.~Lanese, U.~P. Schultz, and C.~Ferreira, editors,
  {\em Reversible Computation: Extending Horizons of Computing - Selected
  Results of the {COST} Action {IC1405}}, volume 12070 of {\em Lecture Notes in
  Computer Science}, pages 151--176. Springer, 2020.

\bibitem{LaneseLMSS13}
I.~Lanese, M.~Lienhardt, C.~A. Mezzina, A.~Schmitt, and J.~Stefani.
\newblock Concurrent flexible reversibility.
\newblock In M.~Felleisen and P.~Gardner, editors, {\em Programming Languages
  and Systems - 22nd European Symposium on Programming, {ESOP} 2013}, volume
  7792 of {\em Lecture Notes in Computer Science}, pages 370--390. Springer,
  2013.

\bibitem{LanesePV19}
I.~Lanese, A.~Palacios, and G.~Vidal.
\newblock Causal-consistent replay debugging for message passing programs.
\newblock In J.~A. P{\'{e}}rez and N.~Yoshida, editors, {\em Formal Techniques
  for Distributed Objects, Components, and Systems - 39th {IFIP} {WG} 6.1
  International Conference, {FORTE} 2019}, volume 11535 of {\em Lecture Notes
  in Computer Science}, pages 167--184. Springer, 2019.

\bibitem{LaneseSUS22}
I.~Lanese, U.~P. Schultz, and I.~Ulidowski.
\newblock Reversible computing in debugging of erlang programs.
\newblock {\em {IT} Prof.}, 24(1):74--80, 2022.

\bibitem{Langerak:1992:BES}
R.~Langerak.
\newblock Bundle event structures: A non-interleaving semantics for lotos.
\newblock In M.~Diaz and R.~Groz, editors, {\em FORTE '92 Conference
  Proceedings}, volume {C-10} of {\em {IFIP} Transactions}, pages 331--346.
  North-Holland, 1993.

\bibitem{DMedicM0Y20}
D.~Medic, C.~A. Mezzina, I.~Phillips, and N.~Yoshida.
\newblock Towards a formal account for software transactional memory.
\newblock In I.~Lanese and M.~Rawski, editors, {\em Reversible Computation -
  12th International Conference, {RC} 2020}, volume 12227 of {\em Lecture Notes
  in Computer Science}, pages 255--263. Springer, 2020.

\bibitem{revPT}
H.~Melgratti, C.~A. Mezzina, and I.~Ulidowski.
\newblock {Reversing Place Transition Nets}.
\newblock {\em {Logical Methods in Computer Science}}, {Volume 16, Issue 4},
  Oct. 2020.

\bibitem{MMPPU:RC20}
H.~C. Melgratti, C.~A. Mezzina, I.~Phillips, G.~M. Pinna, and I.~Ulidowski.
\newblock Reversible occurrence nets and causal reversible prime event
  structures.
\newblock In I.~Lanese and M.~Rawski, editors, {\em Reversible Computation -
  12th International Conference, {RC} 2020}, volume 12227 of {\em Lecture Notes
  in Computer Science}, pages 35--53. Springer, 2020.

\bibitem{lics}
H.~C. Melgratti, C.~A. Mezzina, and G.~M. Pinna.
\newblock A distributed operational view of reversible prime event structures.
\newblock In {\em 36th Annual {ACM/IEEE} Symposium on Logic in Computer
  Science, {LICS}}, pages 1--13. {IEEE}, 2021.

\bibitem{wg2}
C.~A. Mezzina, R.~Schlatte, R.~Gl{\"{u}}ck, T.~Haulund, J.~Hoey, M.~H.
  Cservenka, I.~Lanese, T.~{\AE}. Mogensen, H.~Siljak, U.~P. Schultz, and
  I.~Ulidowski.
\newblock Software and reversible systems: {A} survey of recent activities.
\newblock In I.~Ulidowski, I.~Lanese, U.~P. Schultz, and C.~Ferreira, editors,
  {\em Reversible Computation: Extending Horizons of Computing - Selected
  Results of the {COST} Action {IC1405}}, volume 12070 of {\em Lecture Notes in
  Computer Science}, pages 41--59. Springer, 2020.

\bibitem{MR:CN}
U.~Montanari and F.~Rossi.
\newblock Contextual nets.
\newblock {\em Acta Informatica}, 32(6), 1995.

\bibitem{NPW:PNES}
M.~Nielsen, G.~Plotkin, and G.~Winskel.
\newblock {P}etri {N}ets, {E}vent {S}tructures and {D}omains, {P}art 1.
\newblock {\em Theoretical Computer Science}, 13:85--108, 1981.

\bibitem{quantumcomp}
M.~A. Nielsen and I.~L. Chuang.
\newblock {\em Quantum Computation and Quantum Information (10th Anniversary
  edition)}.
\newblock Cambridge University Press, 2016.

\bibitem{PhilippouP18}
A.~Philippou and K.~Psara.
\newblock Reversible computation in petri nets.
\newblock In J.~Kari and I.~Ulidowski, editors, {\em Reversible Computation -
  10th International Conference, {RC} 2018}, volume 11106 of {\em Lecture Notes
  in Computer Science}, pages 84--101. Springer, 2018.

\bibitem{PU:jlamp15}
I.~Phillips and I.~Ulidowski.
\newblock Reversibility and asymmetric conflict in event structures.
\newblock {\em Journal of Logic and Algebraic Methods in Programming},
  84(6):781--805, 2015.

\bibitem{PhillipsUY12}
I.~Phillips, I.~Ulidowski, and S.~Yuen.
\newblock A reversible process calculus and the modelling of the {ERK}
  signalling pathway.
\newblock In R.~Gl{\"{u}}ck and T.~Yokoyama, editors, {\em Reversible
  Computation, 4th International Workshop, {RC} 2012. Revised Papers}, volume
  7581 of {\em Lecture Notes in Computer Science}, pages 218--232. Springer,
  2013.

\bibitem{Pinna:PN11}
G.~M. Pinna.
\newblock How much is worth to remember? a taxonomy based on {P}etri {N}ets
  {U}nfoldings.
\newblock In L.~M. Kristensen and L.~Petrucci, editors, {\em {PETRI} {NETS}
  2011}, volume 6709 of {\em Lecture Notes in Computer Science}, pages
  109--128, 2011.

\bibitem{richardson2018microservices}
C.~Richardson.
\newblock {\em Microservices patterns: with examples in Java}.
\newblock Simon and Schuster, 2018.

\bibitem{GP:CS}
R.~J. {van G}labbeek and G.~D. Plotkin.
\newblock Configuration structures.
\newblock In {\em 10th Annual {ACM/IEEE} Symposium on Logic in Computer
  Science, {LICS}}, pages 199--209. IEEE Computer Society Press, June 1995.

\bibitem{GP:ESRC}
R.~J. {van G}labbeek and G.~D. Plotkin.
\newblock Event structures for resolvable conflict.
\newblock In J.~Fiala, V.~Koubek, and J.~Kratochv{\'{\i}}l, editors, {\em
  MFCS'04 Conference Proceedings}, volume 3153 of {\em Lecture Notes in
  Computer Science}, pages 550--561. Springer, 2004.

\bibitem{GP:CSESPN}
R.~J. {van G}labbeek and G.~D. Plotkin.
\newblock Configuration structures, event structures and {P}etri nets.
\newblock {\em Theoretical Computer Science}, 410(41):4111--4159, 2009.

\bibitem{Win:ES}
G.~Winskel.
\newblock Event {S}tructures.
\newblock In W.~Brauer, W.~Reisig, and G.~Rozenberg, editors, {\em Petri Nets:
  Applications and Relationships to Other Models of Concurrency}, volume 255 of
  {\em Lecture Notes in Computer Science}, pages 325--392. Springer, 1986.

\end{thebibliography}

\end{document}